\documentclass[11pt]{article}

\PassOptionsToPackage{table}{xcolor}
\usepackage[preprint]{acl}   % 10-06 user: preprint, so the title card shows the authors

\usepackage{times}
\usepackage{latexsym}
\usepackage[T1]{fontenc}
\usepackage[utf8]{inputenc}
\usepackage{microtype}
\usepackage{inconsolata}

\usepackage{amsmath,amsfonts,bm}

\def\eqref#1{equation~\ref{#1}}
\def\1{\bm{1}}

\DeclareMathAlphabet{\mathsfit}{\encodingdefault}{\sfdefault}{m}{sl}
\SetMathAlphabet{\mathsfit}{bold}{\encodingdefault}{\sfdefault}{bx}{n}

\usepackage{url}
\usepackage{graphicx}
\usepackage{subcaption}
\usepackage{adjustbox}
\usepackage{fontawesome5}   % \faCog
\usepackage{booktabs}
\usepackage{float}      % [H]: appendix B full-results tables stay under their subsection
\usepackage{afterpage}  % 10-12 user: Table 4 deferred to the page after Sec 6 starts
\makeatletter
\newcommand{\afterleftcol}[1]{\afterpage{\if@firstcolumn #1\else\afterleftcol{#1}\fi}}
\makeatother
\usepackage{stfloats}   % allows [b] for two-column floats (table*/figure* at the page bottom)
\usepackage{multirow}
\usepackage{pifont}

\usepackage{setspace}
\definecolor{LinkGreen}{RGB}{50,150,100}
\definecolor{LinkBlue}{RGB}{0,140,210}   % sky blue for citations / refs / URLs
\definecolor{LinkOrange}{RGB}{190,85,20}   % 10-11 user: burnt orange for citations / refs / URLs (first try 214,110,30)
\hypersetup{colorlinks=true, citecolor=LinkBlue, linkcolor=LinkBlue, urlcolor=LinkBlue}   % 10-11 user: back to sky blue (ACL default = acl.sty darkblue; orange tried: LinkOrange)

\usepackage[most]{tcolorbox}
\tcbuselibrary{skins,breakable}
\definecolor{rqBlueBg}{HTML}{EAF4FF}
\definecolor{rqBlueBorder}{HTML}{6AADE4}

\definecolor{MyGray}{HTML}{F2F2F2}   % table group-header rows
\definecolor{MyBlue}{HTML}{C0E0F8}   % table: best score
\definecolor{MyBlueLight}{HTML}{E6F2FC}  % table: second-best score
\definecolor{SimMemOrange}{HTML}{DD8452}   % 10-12 user: = memory-unit orange of Fig 5 (Fig 5 caption)
\definecolor{DegOrange}{RGB}{226,140,80}   % table: below-baseline shading (mixed with white); 09-29 orange again, a bit richer (was 214,150,105)
\definecolor{ImpGreen}{RGB}{95,160,210}    % table: above-baseline shading (mixed with white); 09-29 blue, a bit richer than bench_satuate #7DAFCB

\definecolor{Green}{RGB}{110,140,120}
\definecolor{GreenLight}{RGB}{242,247,243}

\definecolor{Blue}{RGB}{100,125,150}
\definecolor{BlueLight}{RGB}{242,245,249}

\definecolor{Yellow}{RGB}{170,150,95}
\definecolor{YellowLight}{RGB}{249,247,240}

\definecolor{Red}{RGB}{145,104,104}
\definecolor{FaithColor}{RGB}{0,0,0}        % black (was green 0,125,65)
\definecolor{AccessColor}{RGB}{0,0,0}       % black (was orange 215,110,0)
\definecolor{RobustColor}{RGB}{0,0,0}       % black (was purple 125,55,170)

\newcommand{\faith}{\texorpdfstring{\textcolor{FaithColor}{faithfulness}}{faithfulness}}

\newcommand{\access}{\texorpdfstring{\textcolor{AccessColor}{accessibility}}{accessibility}}

\definecolor{FaithHead}{HTML}{4F7FA3}    % = darkest blue of the faithfulness bar figure (Full / major loss)
\definecolor{BenefitGreen}{HTML}{4E8B5F} % memory benefit in text: muted green (09-30 user)
\definecolor{TaxRed}{HTML}{B5534B}       % memory tax in text: muted red (09-30 user)
\definecolor{AccessHead}{RGB}{195,110,30}   % amber
\definecolor{RobustHead}{HTML}{4E7D6B}   % dark sage (09-29; was plum 125,70,145)
\newcommand{\FaithHd}{Faithfulness}   % 09-30 user: headings in black (colour FaithHead kept for other uses)
\newcommand{\AccessHd}{Accessibility}   % 09-30 user: headings in black (colour AccessHead kept for other uses)
\newcommand{\RobustHd}{Robustness}   % 09-30 user: headings in black (colour RobustHead kept for other uses)
\definecolor{RedLight}{RGB}{248,243,243}

\definecolor{Purple}{RGB}{118,108,140}
\definecolor{PurpleLight}{RGB}{245,243,248}
\definecolor{ObsBg}{HTML}{F4F9FF}   % lighter than rqBlueBg (EAF4FF)
\tcbset{obsbreakable/.style={breakable, lines before break=1, toprule at break=0pt, bottomrule at break=0pt}}   % lines before break: default 2, so a box near a column end with room for one line moved whole
\makeatletter
\ifacl@linenumbers
  \tcbset{obsbreak/.style={}}
  \newcommand{\obsboxwrap}[1]{\begingroup\begin{nolinenumbers}#1\end{nolinenumbers}\endgroup}
\else
  \tcbset{obsbreak/.style={obsbreakable}}
  \newcommand{\obsboxwrap}[1]{#1}
\fi
\makeatother
\newcommand{\obsbox}[2][]{%
  \obsboxwrap{%
    \begin{tcolorbox}[colframe=black, colback=ObsBg, boxrule=0.8pt, arc=2mm,   % 10-11 user: thinner frame (was 1.2pt)
  top=4pt, bottom=4pt, left=3pt, right=3pt, boxsep=1pt,
  before skip=3.5pt, after skip=3.5pt,   % 10-12 user: slightly tighter (was 5pt)
  obsbreak, enhanced, #1]
        #2
    \end{tcolorbox}}%
}

\tcbset{
  aibox/.style={
    width=\linewidth,
    top=7pt,
    bottom=2pt,
    colback=rqBlueBg,
    colframe=black,
    colbacktitle=black,
    enhanced,
    center,
    attach boxed title to top left={yshift=-0.1in,xshift=0.15in},
    boxed title style={boxrule=0pt,colframe=white,},
  }
}

\definecolor{TakeawayBg}{HTML}{F4F9FF}   % lighter than rqBlueBg (EAF4FF)
\newtcolorbox{takeaway}[1][]{
  aibox,
  colback=TakeawayBg,
  top=2pt, bottom=2pt,                     % no "Takeaway N" title tab (user 2026-09-27)
  before skip=4pt, after skip=4pt,         % tighter vertical footprint
  boxrule=0.4pt,                           % thin frame
  #1
}
\definecolor{TodoBlue}{RGB}{0,90,200}

\newcommand{\catsq}{\rule[0.02ex]{0.6em}{0.6em}}

\newcommand{\boxednum}[1]{\tikz[baseline=(bn.base)]{%
  \node[draw, rectangle, inner sep=1.6pt, line width=0.6pt] (bn) {\footnotesize\textbf{#1}};}}   % 10-05 user: TikZ version

\definecolor{IconBlue}{RGB}{100,155,210}   % industry-system gear icon (light blue)
\definecolor{IconRed}{RGB}{179,27,27}      % arXiv icon (arXiv brand red)
\definecolor{CheckBlue}{RGB}{40,95,175}   % table check marks
\newcommand{\cmark}{\textcolor{CheckBlue}{\faCheck}}
\definecolor{ClsLong}{HTML}{6B7280}    % long context: slate
\definecolor{ClsRAG}{HTML}{A07A32}     % retrieval-only: ochre
\definecolor{ClsFlat}{HTML}{3F6E8F}    % flat store: blue (10-05 user, = Figure 3; was sage 4E7D6B)
\definecolor{ClsHier}{HTML}{D4A72C}    % hierarchical: yellow (10-05 user, = Figure 3)
\definecolor{ClsGraph}{HTML}{665495}   % graph-based: purple (10-05 user, = Figure 3)
\definecolor{ClsTrain}{HTML}{A9564C}   % trained policy: brick (10-05 user, = Figure 3; was flame orange D2643A)
\newcommand{\clsicon}[2]{\makebox[1.35em][l]{\textcolor{#1}{#2}}}

\newcommand{\clsFlat}{\clsicon{ClsFlat}{\faTh}}
\newcommand{\clsHier}{\clsicon{ClsHier}{\faLayerGroup}}
\newcommand{\clsGraph}{\clsicon{ClsGraph}{\faProjectDiagram}}
\newcommand{\clsTrain}{\clsicon{ClsTrain}{\faIcon{fire-alt}}}

\newcommand{\figsep}{\par\nointerlineskip\vspace{3pt}\hbox to\linewidth{\color{black!45}\leaders\hbox to 5pt{\hss\rule{3pt}{0.6pt}\hss}\hfill}\nointerlineskip\vspace{3pt}}

  {\end{minipage}\end{lrbox}\fbox{\usebox{0}}}

\newcommand{\dlt}[1]{\,{\tiny\textcolor{black!60}{(#1)}}}

\newcommand{\phline}{\noindent\textcolor{black!20}{\rule[0.25ex]{\linewidth}{0.55em}}\par}
\newcount\phcount
\newcommand{\phlines}[1]{\par\begingroup\setlength{\parskip}{0pt}\phcount=#1\relax
  \loop\ifnum\phcount>0 \phline\advance\phcount by -1 \repeat\endgroup}
\usepackage{titletoc}

\makeatletter
\tcbset{contbreak/.style={%
  title after break={\kvtcb@title\ \mbox{\normalfont\itshape(continued)}}}}
\makeatother
\DeclareFontFamily{T1}{optimistic}{}
\DeclareFontShape{T1}{optimistic}{m}{n}{<-> s * [0.88] optimistic}{}
\DeclareFontShape{T1}{optimistic}{b}{n}{<-> s * [0.88] optimistic}{}
\DeclareFontShape{T1}{optimistic}{bx}{n}{<-> ssub * optimistic/b/n}{}
\DeclareFontShape{T1}{optimistic}{m}{sc}{<-> ssub * optimistic/m/n}{}
\DeclareFontShape{T1}{optimistic}{b}{sc}{<-> ssub * optimistic/b/n}{}
\DeclareFontShape{T1}{optimistic}{bx}{sc}{<-> ssub * optimistic/b/n}{}
\pdfmapline{+optimistic < Optimistic.ttf <T1-WGL4.enc}
\newcommand{\optimistic}{\fontfamily{optimistic}\selectfont}
\newcommand{\fauxboldwidth}{0.35}
\newcommand{\fauxbold}[1]{%
  \pdfliteral{2 Tr \fauxboldwidth\space w}#1\pdfliteral{0 Tr}}

\colorlet{projectlink}{LinkBlue}   % = urlcolor set by \hypersetup above
\newcommand{\globeicon}{%
  \raisebox{0.03em}{\scalebox{0.81}{\faGlobe}}}
\newcommand{\githubicon}{%
  \raisebox{-0.05em}{\includegraphics[height=0.79em]{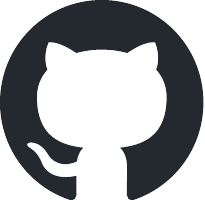}}}
\newcommand{\hficon}{%
  \raisebox{-0.08em}{\includegraphics[height=0.85em]{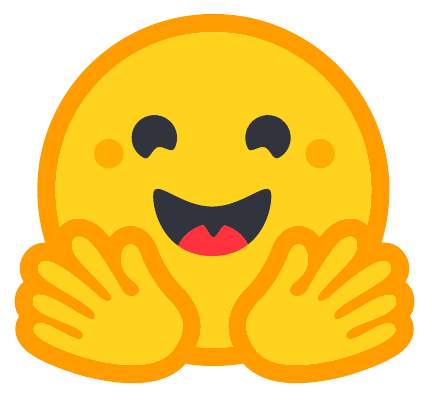}}}
\newcommand{\afflfont}{\fontsize{10pt}{12pt}\selectfont}

\definecolor{metafg}{HTML}{1C2B33}
\definecolor{metabg}{HTML}{F2E4E6}   % 10-23 user: dusty pink (earlier oatmeal F3ECE2, muted light rose F2E3E1)
\definecolor{memtaxaccent}{HTML}{7A2E3A}   % 10-23 user: "Memory Tax" in the title, wine red (earlier caramel 8A5A2B, muted dark red A84B45)

\newtcolorbox{titlecard}{
  enhanced, frame hidden, colback=metabg,
  arc=10pt, boxsep=0pt,
  left=0.55cm, right=0.55cm, top=0.4cm, bottom=0.4cm,   % 10-23 user: tighter, Conclusion one line over (was 0.5cm)
  width=\textwidth,
  grow to left by=0.55cm, grow to right by=0.55cm,
}

\makeatletter
\newcommand{\@abstracttext}{}
\newcommand{\abstracttext}[1]{\gdef\@abstracttext{#1}}

\AtBeginDocument{%
  \def\@maketitle{%
    \begin{titlecard}
      \color{metafg}
      \centering
      {\optimistic\fontsize{16pt}{19pt}\selectfont \fauxbold{\@title} \par}
      \vskip 0.12in   % 10-23: was 0.16in
      {\fontsize{10.8pt}{13pt}\selectfont
       \ifacl@anonymize
         \begin{tabular}[t]{c}{\optimistic Anonymous ACL submission}\end{tabular}%
       \else
         \renewcommand{\textbf}[1]{{\optimistic ##1}}%
         \begin{tabular}[t]{c}\@author\end{tabular}%
       \fi\par}
      \vskip 0.16in   % 10-23: was 0.20in
      \begin{minipage}{\dimexpr\linewidth-0.5cm\relax}
        \fontsize{11pt}{13.6pt}\selectfont   % 10-06 user: one size up (was 10pt / 12pt)
        \setlength{\parindent}{0pt}
        \@abstracttext
      \end{minipage}
      \vskip 0.10in   % 10-23: was 0.14in
      \begin{minipage}{\dimexpr\linewidth-0.5cm\relax}
        \fontsize{10.5pt}{16pt}\selectfont
        \makebox[\linewidth]{{\color{projectlink}%
          \href{https://hexuan21.github.io/MemoryTax/}{\globeicon\ Project Page}\quad
          \href{https://github.com/hexuan21/MemoryTax}{\githubicon\ Code}\quad
          \href{https://huggingface.co/datasets/hexuan21/MemoryTax-Res}{\hficon\ Results}}}%
        \llap{\raisebox{-0.38\height}[0.62\height][5pt]{\includegraphics[height=1.9\baselineskip]{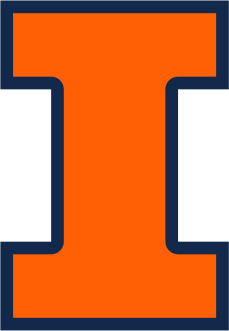}}}   % 10-18 user: 3.2 -> 2.2 -> 2.6 -> 2.2; 10-23 -> 1.9
      \end{minipage}
      \vskip 0.02in
    \end{titlecard}%
  }%
}
\makeatother
\title{\textcolor{memtaxaccent}{Memory Tax}: Dissecting the Net Value of Agent Memory}

\author{
 \textbf{Xuan He\textsuperscript{1}},
 \textbf{Sining Zhoubian\textsuperscript{2}},   \textbf{Ruozhen Yang\textsuperscript{1}},
 \textbf{Yuhao Wen\textsuperscript{2}},
 \textbf{Zhineng Wang\textsuperscript{1}},
\\
 \textbf{Yuyang Hu\textsuperscript{3}},
 \textbf{Ke Yang\textsuperscript{1}},
 \textbf{Yuyu Zhang\textsuperscript{4}},
 \textbf{Ping Nie\textsuperscript{5}},
 \textbf{Jiawei Han\textsuperscript{1}},
 \textbf{ChengXiang Zhai\textsuperscript{1}}
\\
\\
 \afflfont
 \textsuperscript{1}University of Illinois Urbana Champaign,
 \textsuperscript{2}Tsinghua University,
 \\
 \afflfont
 \textsuperscript{3}Renmin University of China,
 \textsuperscript{4}Verdent.ai,
 \textsuperscript{5}University of Waterloo
}

\abstracttext{%
Memory is increasingly used in long-horizon agents, yet the net value of building memory over direct retrieval from raw history remains largely unclear. 
Through a large-scale evaluation across multiple benchmarks and model backbones, we find many dedicated memory systems underperform naive RAG in many cases, indicating a negative net value. 
Taking RAG as a reference, we dissect the net value of memory into the gains and losses over direct retrieval, termed memory benefit and \textbf{memory tax}.
These gains and losses arise because memory process changes how the raw information is organized and accessed. We view such changes as the double-edged deviations from the raw history. They are originally designed and introduced to benefit the task, for example by reorganizing facts or distilling preferences, but some may also remove or obscure evidence, giving rise to memory tax.
Since memory benefit reflects the expected gains of its design, we then focus more on the overlooked memory tax, study how it arises from the system and evaluation side, and generalize the concept to more application scenarios of memory. 
We hope the notion of memory tax can provide a new lens and shift more attention from standalone performance to the net value of memory, informing more reliable memory designs and evaluations. 
We will open-source the code and results to the community. 
}

\begin{document}
\maketitle

\raggedbottom   % main body too: when a box (e.g. an obsbox) does not fit at a column end, leave the space at the bottom
               % instead of stretching it into the glue above the next heading (09-30: big gap before 4.4)
\section{Introduction}
\label{sec:intro}

Memory has become an important component of long-horizon LLM agents, supporting recall across interaction histories and reuse of prior experience~\citep{wang2023voyager,xie2024osworld,gur2024real,hu-etal-2025-hiagent}.
Recent work has introduced increasingly diverse memory representations and management mechanisms~\citep{chhikara2025mem0,kang2025memory,fang2026lightmem,hu2026memoryageaiagents,jiang2026magma}, aiming to make relevant information easier to preserve and utilize, but often incur additional computation  over the original context. 
As memory systems grow more complex, it remains unclear how much these operations improve over a straightforward alternative, such as directly retrieve from the raw history.

Surveying prior works, it shows many memory designs primarily compare against other memories, while naive RAG is often overlooked, and when included, its setup is also not always fully specified (Appendix~\ref{apdx:rag_prior_works}). 
Without a retrieval baseline, the value of building memory remains \textbf{entangled} with the value of direct evidence retrieval, leaving its marginal or \textbf{net value} underexamined. 
% Meanwhile, recent studies also reported some cases where the naive retrieval is competitive with elaborate memory. 
So this suggests a gap between the growing complexity of memory design and our understanding of whether and where that complexity actually helps, motivating us to revisit a fundamental question:

\obsbox{\textit{What is the net value of memory over a straightforward baseline like naive RAG?}}

To study this question, we start from a common setting, personal-agent conversational memory. 
We evaluate representative memory designs across several benchmarks and model backbones, using naive RAG as a reference to assess the net value of memory. The results show that naive RAG remains a strong baseline and many memory systems underperform it across diverse settings, indicating a \textbf{negative net value} in these cases. 
But a closer query-level comparison shows memory is \textbf{not uniformly worse}. It does solve some queries RAG misses, where organization or restructuring may help, while losing others that RAG already solves. We refer to the former as memory benefit and the latter as \textit{\textbf{memory tax}}. When the tax outweighs the benefit, memory falls below RAG overall.

% \textbf{inevitable deviations}\footnote{We use \textbf{deviation} \textit{in a neutral sense} to denote both beneficial and harmful changes from source introduced by memory.}

These gains and losses suggest a different view of memory.
Memory goes far beyond a purely beneficial augmentation as often treated in prior works, and can be better understood as a double-edged process that changes how information from raw history is structured and accessed. We view such changes as \textbf{inevitable deviations} from the raw history. 
These deviations give memory its \textbf{dual nature}. 
They are originally designed and intended to better serve the task, e.g., by organizing scattered facts, distilling preferences, etc., yet some may also have unintended harmful effects due to particular design choices, such as losing or obscuring evidence. 
Memory benefit and tax thus capture the two sides, and their balance defines the net value of memory, reflected in the external performance.

Memory benefit reflects the intended gains of memory design, such as reorganization or abstraction, which is expected. So we focus more on how the memory tax emerges. We first examine \textbf{faithfulness} for how well the information survives memory process, and then \textbf{accessibility} for how easily the preserved information can be recalled or accessed. We further study \textbf{robustness} for reliability of memory across backbone changes. 
These dimensions characterize memory tax from system side, and its balance with memory benefit also highly depends on the task setting. 
Across more personal-agent memory benchmarks, we find naive RAG already achieves high absolute scores, suggesting a retrieval-friendly task setting with limited \textbf{benchmark headroom} for memory to deliver more benefit and thus making the tax more visible.

Moreover, the benefit-tax view is generalizable and not tied to particular memory forms, as nontrivial memory operations can introduce both beneficial and harmful deviations from the source. We further examine this in \textbf{experiential memory} and \textbf{world-model memory}. Finally, we derive some design implications from the perspective of memory deviations and benefit-tax trade-offs. We focus on simple changes that can be applied to existing memories without introducing new architectures.

We hope that the notion of memory tax can \textbf{shift more attention from standalone memory performance to the net value of memory and its dissection into benefit and tax}, thus informing more reliable memory design and evaluation.
Our contributions are as follows:

\boxednum{1} Reassess the net value of long-term memory over direct retrieval holistically.
\boxednum{2} Dissect the net value into memory benefit and memory tax, and systematically analyze the source of memory tax.
\boxednum{3} Generalize the benefit-tax view to broader settings and derive general design implications.
\vspace{-0.1em}
\section{Preliminary}
\label{sec:preliminary}
\vspace{-0.1em}

Long-term memory for agents maintains information from past interactions and selectively exposes it to the model at query time. 
Let the interaction history be $\mathcal{H}=\{h_1,\ldots,h_T\}$, where each $h_t$ may contain user/agent messages, observations, or other types of content. 
A memory system processes this history into a persistent memory $\mathcal{M}=I(\mathcal{H})$, where $I$ denotes the \emph{memory ingestion} process. Given a task query $q$, it retrieves relevant memory units $\mathcal{C}=R(q,\mathcal{M})$ and generates an answer $\hat{y}=G(q,\mathcal{C})$. Different systems instantiate $\mathcal{M}$ as summaries, extracted facts, structured records, episodic entries, or other representations.

Naive chunk-based RAG provides a natural reference under this formulation. 
It keeps raw history with minimal processing and retrieves relevant chunks directly.
We use it throughout this work as the comparison reference for assessing the net value of memory, and measured memory benefit and memory tax are thus relative to this reference.
\vspace{-0.3em}
\section{Experiments}
\label{sec:experiments}
\vspace{-0.1em}

\subsection{Setup}
\label{sec:exp_setup}

% Main results table: rows = baselines (grouped by memory-organization type),
% columns = benchmarks; each benchmark has three columns, one per backbone.
% Scores are LLM-judge accuracy (%). PersonaMem is MCQ exact-match accuracy.
% Empty cells = not yet run / not yet double-checked.
% Cell shading: orange = below RAG (2rd) in that column, green = above; opacity
% grows linearly with |delta| (8% + 2%/point, capped at 75%). Regenerate after
% editing scores.
% Scores synced from update_4B.txt / update_27B.txt / update_5_6_luna.txt (double-checked cells only);
% regenerate with gen_table1.py. 5.6-luna LME: memory systems run on the 100-question stratified subset
% (sub100); the two RAG rows use their sub100 re-scoring (82.00 / 90.00 after the 10-05 v3-pref re-run).
% Removed (2026-09-30): EverOS, DCI-Agent and ReFind rows; left empty: LightMem except PersonaMem; Memory-R2 and MAGMA only 5.6-luna so far (Mem-T / MAGMA filled 10-02, Memory-R2 10-03);
% (PlugMem PersonaMem filled again 2026-09-28 after the e15_s15 re-run.)
% 10-13 user: Full Context is ranked too (best / second best per column over all rows except No / Gold Context).
% Vanilla rows (No Context / Full Context / Gold Context) and RAG (2rd) are unshaded; No / Gold Context are not ranked (RAG (1rd) is unshaded and has no delta, 10-11 user); Full Context MINT truncates stores >250k tokens.
\begin{table*}[ht]
\centering
\caption{Main results. Scores are judge accuracy (\%). We highlight \textbf{the best} and \underline{the second best} per column. Cells are shaded in proportion to how far they \colorbox{DegOrange!30}{fall below} or \colorbox{ImpGreen!30}{rise above} RAG (2rd). \textbf{Q3.5-4B}, \textbf{Q3.6-27B} and \textbf{5.6-luna} denote the \textbf{Qwen3.5-4B}, \textbf{Qwen3.6-27B} and \textbf{GPT-5.6 Luna (low)} backbones, respectively.}
\vspace{-0.8em}
\label{tab:main_results}
\small
\setlength{\tabcolsep}{5pt}
\renewcommand{\arraystretch}{1.03}   % 10-03 user: row spacing 1.1x
\begin{adjustbox}{max width=0.98\textwidth}
\begin{tabular}{l l| c c c| c c c| c c c}
\toprule
\multirow{2}{*}{\textbf{Category}} & \multicolumn{1}{l}{\multirow{2}{*}{\textbf{Method}}} & \multicolumn{3}{c}{\textbf{LongMemEval-s}$^\S$} & \multicolumn{3}{c}{\textbf{PersonaMem-v2}$^\S$} & \multicolumn{3}{c}{\textbf{MINT-Eval}$^\S$} \\
\cmidrule(lr){3-5} \cmidrule(lr){6-8} \cmidrule(lr){9-11}
 & \multicolumn{1}{l}{} & \textbf{Q3.5-4B} & \textbf{Q3.6-27B} & \textbf{5.6-luna} & \textbf{Q3.5-4B} & \textbf{Q3.6-27B} & \textbf{5.6-luna}$^\dagger$ & \textbf{Q3.5-4B} & \textbf{Q3.6-27B} & \textbf{5.6-luna} \\
\midrule
\multirow{2}{*}{\textbf{Vanilla}} & No Context & 9.0 & 9.0 & 17.6 & 20.9 & 25.6 & 28.5 & 2.3 & 3.8 & 8.5 \\
 & Full Context & 58.6 & 68.4 & \underline{84.0} & 30.1 & 45.4 & 43.6 & \underline{26.9} & 28.7 & \textbf{39.7} \\
 & Gold Context & 73.4 & 84.6 & 92.8 & 58.7 & 72.9 & 69.7 & -- & -- & -- \\
\midrule
\multirow{2}{*}{\textbf{Retrieval-only}} & RAG (1rd) & 70.2 & 77.6 & 82.0 & \underline{37.2} & 46.8 & 45.1 & 25.8 & 28.6 & 29.9 \\
 & RAG (2rd) & \textbf{72.2} & \textbf{81.4} & \textbf{90.0} & 36.2 & \underline{48.7} & \textbf{48.4} & 26.1 & \underline{33.1} & 34.5 \\
\midrule
\multirow{3}{*}{\clsFlat{}\textbf{Flat store}} & A-Mem & \cellcolor{DegOrange!12}70.4\dlt{$-$1.8} & \cellcolor{DegOrange!12}\underline{79.4}\dlt{$-$2.0} & \cellcolor{DegOrange!20}\underline{84.0}\dlt{$-$6.0} & \cellcolor{DegOrange!13}33.7\dlt{$-$2.5} & \cellcolor{DegOrange!9}48.4\dlt{$-$0.3} & \cellcolor{DegOrange!21}41.8\dlt{$-$6.6} & \cellcolor{DegOrange!11}24.4\dlt{$-$1.7} & \cellcolor{DegOrange!25}24.4\dlt{$-$8.7} & \cellcolor{DegOrange!23}27.1\dlt{$-$7.4} \\
 & MemOS & \cellcolor{DegOrange!28}62.0\dlt{$-$10.2} & \cellcolor{DegOrange!18}76.6\dlt{$-$4.8} & \cellcolor{DegOrange!32}78.0\dlt{$-$12.0} & \cellcolor{ImpGreen!10}\underline{37.2}\dlt{+1.0} & \cellcolor{ImpGreen!11}\textbf{50.2}\dlt{+1.5} & \cellcolor{DegOrange!21}41.7\dlt{$-$6.7} & \cellcolor{DegOrange!24}18.0\dlt{$-$8.1} & \cellcolor{DegOrange!35}19.7\dlt{$-$13.4} & \cellcolor{DegOrange!16}30.3\dlt{$-$4.2} \\
 & SimpleMem & \cellcolor{DegOrange!32}60.2\dlt{$-$12.0} & \cellcolor{DegOrange!24}73.6\dlt{$-$7.8} & \cellcolor{DegOrange!20}\underline{84.0}\dlt{$-$6.0} & \cellcolor{DegOrange!11}34.7\dlt{$-$1.5} & \cellcolor{DegOrange!12}46.7\dlt{$-$2.0} & \cellcolor{DegOrange!15}45.1\dlt{$-$3.3} & \cellcolor{DegOrange!23}18.6\dlt{$-$7.5} & \cellcolor{DegOrange!35}19.9\dlt{$-$13.3} & \cellcolor{DegOrange!31}22.9\dlt{$-$11.6} \\
\midrule
\multirow{3}{*}{\clsHier{}\textbf{Hier/Lifecycle}} & MemoryOS & \cellcolor{DegOrange!23}64.6\dlt{$-$7.6} & \cellcolor{DegOrange!24}73.4\dlt{$-$8.0} & \cellcolor{DegOrange!28}80.0\dlt{$-$10.0} & \cellcolor{ImpGreen!10}37.0\dlt{+0.9} & \cellcolor{DegOrange!9}48.2\dlt{$-$0.5} & \cellcolor{DegOrange!19}43.0\dlt{$-$5.4} & \cellcolor{DegOrange!22}19.0\dlt{$-$7.2} & \cellcolor{DegOrange!22}26.3\dlt{$-$6.9} & \cellcolor{DegOrange!11}33.2\dlt{$-$1.3} \\
 & LightMem & \cellcolor{DegOrange!37}57.6\dlt{$-$14.6} & \cellcolor{DegOrange!13}78.8\dlt{$-$2.6} & \cellcolor{DegOrange!34}77.0\dlt{$-$13.0} & \cellcolor{ImpGreen!8}36.4\dlt{+0.2} & \cellcolor{DegOrange!11}47.0\dlt{$-$1.7} & \cellcolor{DegOrange!13}\underline{45.9}\dlt{$-$2.5} & \cellcolor{DegOrange!19}20.8\dlt{$-$5.4} & \cellcolor{DegOrange!28}23.0\dlt{$-$10.2} & \cellcolor{DegOrange!27}25.2\dlt{$-$9.3} \\
 & OpenViking & \cellcolor{DegOrange!75}31.8\dlt{$-$40.4} & \cellcolor{DegOrange!53}58.8\dlt{$-$22.6} & \cellcolor{DegOrange!62}63.0\dlt{$-$27.0} & \cellcolor{DegOrange!18}31.3\dlt{$-$4.9} & \cellcolor{DegOrange!28}38.7\dlt{$-$10.1} & \cellcolor{DegOrange!28}38.2\dlt{$-$10.2} & \cellcolor{DegOrange!40}10.2\dlt{$-$15.9} & \cellcolor{DegOrange!37}18.6\dlt{$-$14.6} & \cellcolor{DegOrange!29}24.2\dlt{$-$10.3} \\
\midrule
\multirow{3}{*}{\clsGraph{}\textbf{Graph-based}} & HippoRAG2 & \cellcolor{DegOrange!15}68.8\dlt{$-$3.4} & \cellcolor{DegOrange!14}78.2\dlt{$-$3.2} & \cellcolor{DegOrange!26}81.0\dlt{$-$9.0} & \cellcolor{DegOrange!13}33.9\dlt{$-$2.3} & \cellcolor{DegOrange!13}46.5\dlt{$-$2.3} & \cellcolor{DegOrange!14}45.3\dlt{$-$3.1} & \cellcolor{DegOrange!11}24.6\dlt{$-$1.5} & \cellcolor{DegOrange!27}23.5\dlt{$-$9.6} & \cellcolor{DegOrange!21}27.8\dlt{$-$6.7} \\
 & MAGMA & \cellcolor{DegOrange!31}60.8\dlt{$-$11.4} & \cellcolor{DegOrange!34}68.2\dlt{$-$13.2} & \cellcolor{DegOrange!36}76.0\dlt{$-$14.0} & \cellcolor{DegOrange!14}33.1\dlt{$-$3.1} & \cellcolor{DegOrange!25}40.5\dlt{$-$8.3} & \cellcolor{DegOrange!24}40.6\dlt{$-$7.8} & \cellcolor{DegOrange!13}23.7\dlt{$-$2.4} & \cellcolor{DegOrange!20}27.0\dlt{$-$6.2} & \cellcolor{DegOrange!18}29.5\dlt{$-$5.0} \\
 & PlugMem & \cellcolor{DegOrange!16}68.0\dlt{$-$4.2} & \cellcolor{DegOrange!19}75.8\dlt{$-$5.6} & \cellcolor{DegOrange!48}70.0\dlt{$-$20.0} & \cellcolor{DegOrange!10}35.0\dlt{$-$1.2} & \cellcolor{DegOrange!15}45.2\dlt{$-$3.5} & \cellcolor{DegOrange!25}39.8\dlt{$-$8.6} & \cellcolor{DegOrange!17}21.5\dlt{$-$4.7} & \cellcolor{ImpGreen!8}\textbf{33.3}\dlt{+0.1} & \cellcolor{ImpGreen!15}37.8\dlt{+3.3} \\
\midrule
\multirow{2}{*}{\clsTrain{}\textbf{Trained Policy}} & Mem-T & \cellcolor{DegOrange!10}\underline{71.0}\dlt{$-$1.2} & \cellcolor{DegOrange!21}75.0\dlt{$-$6.4} & \cellcolor{DegOrange!32}78.0\dlt{$-$12.0} & \cellcolor{DegOrange!16}32.1\dlt{$-$4.1} & \cellcolor{DegOrange!21}42.1\dlt{$-$6.6} & \cellcolor{DegOrange!23}41.0\dlt{$-$7.4} & \cellcolor{ImpGreen!10}\textbf{27.0}\dlt{+0.9} & \cellcolor{DegOrange!13}30.4\dlt{$-$2.7} & \cellcolor{ImpGreen!16}\underline{38.4}\dlt{+3.9} \\
 & Memory-R2 & \cellcolor{DegOrange!21}65.8\dlt{$-$6.4} & \cellcolor{DegOrange!25}73.0\dlt{$-$8.4} & \cellcolor{DegOrange!30}79.0\dlt{$-$11.0} & \cellcolor{ImpGreen!12}\textbf{38.2}\dlt{+2.0} & \cellcolor{DegOrange!9}48.4\dlt{$-$0.3} & \cellcolor{DegOrange!16}44.2\dlt{$-$4.2} & \cellcolor{DegOrange!18}21.3\dlt{$-$4.8} & \cellcolor{DegOrange!31}21.5\dlt{$-$11.6} & \cellcolor{DegOrange!31}22.9\dlt{$-$11.6} \\
\bottomrule
\end{tabular}
\end{adjustbox}
{\footnotesize\raggedright $^\S$LongMemEval is relatively retrieval-friendly, while the other two involve more complex demands, see \S\ref{sec:regimes}.

$^\dagger$Interestingly, on PersonaMem-v2, 5.6-luna usually scores lower than Q3.6-27B, opposite of the other two benchmarks, see~\ref{apdx:more_explanations}. 
\par}
\vspace{-0.4em}
\end{table*}

\begin{figure*}[ht]
\centering
\begin{subfigure}[b]{0.49\textwidth}
    \centering
    \includegraphics[width=\linewidth]{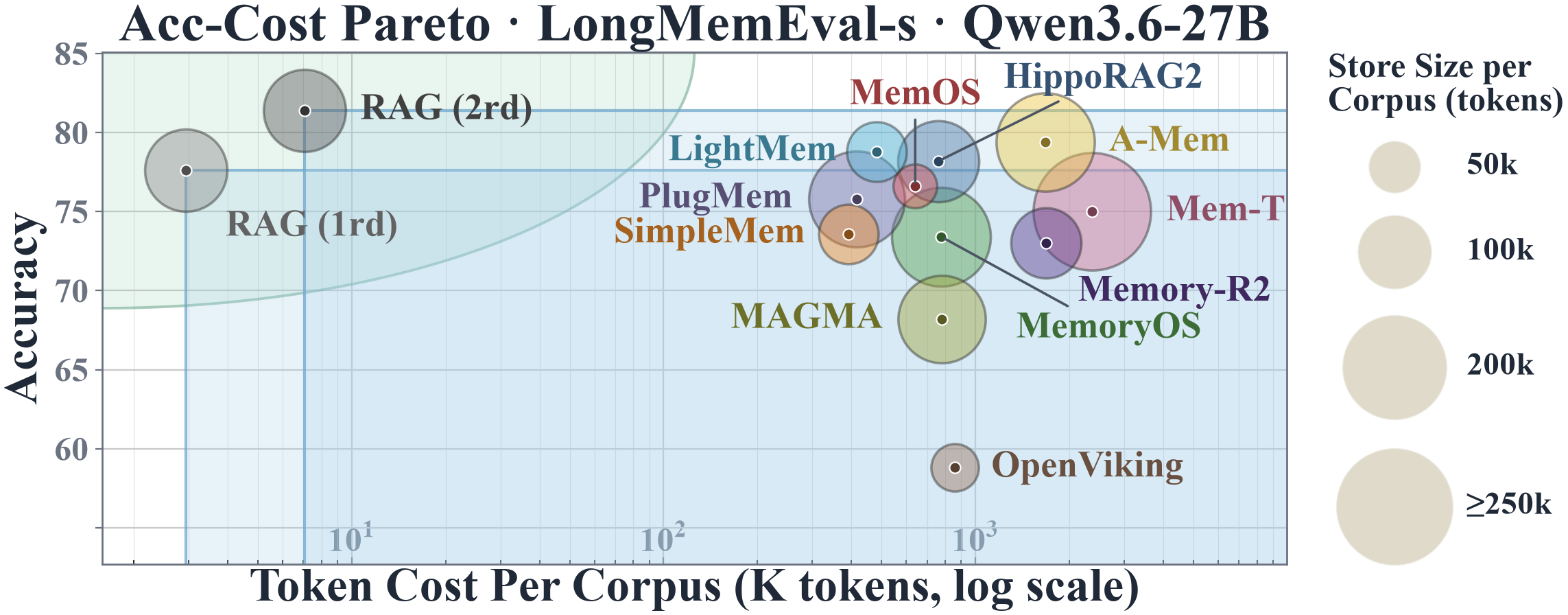}
    \phantomcaption\label{fig:pareto_lme}
\end{subfigure}\hfill
\begin{subfigure}[b]{0.49\textwidth}
    \centering
    \includegraphics[width=\linewidth]{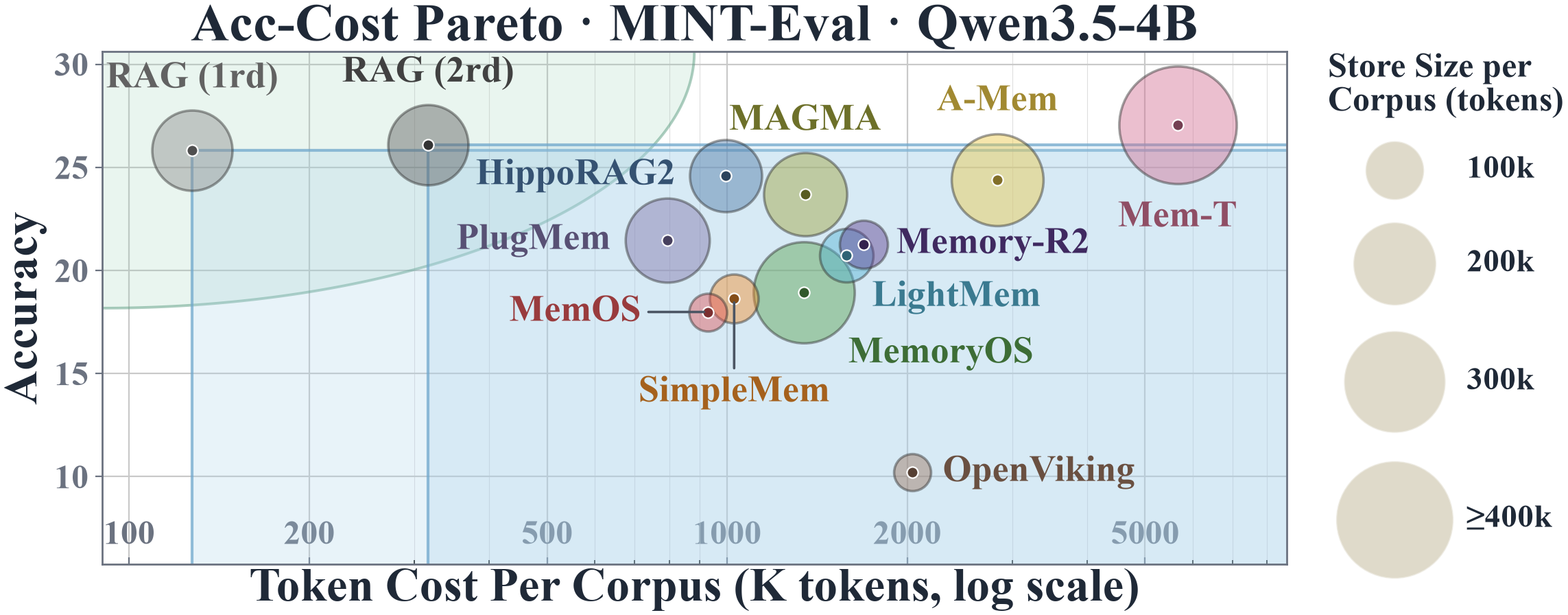}
    \phantomcaption\label{fig:pareto_mint}
\end{subfigure}
\vspace{-1.5em}
\caption{Pareto plot of Accuracy vs. Token-cost. Bubble size is the memory store size in tokens.}
\label{fig:pareto_27b}
\vspace{-1em}
\end{figure*}

Memory is increasingly used across different scenarios. We begin with its common application, \textbf{personal agent memory} or \textbf{conversational memory}, and later extend to more domains (\S\ref{sec:gen_experiential} and \S\ref{sec:gen_world_model}).

\textbf{Benchmarks.}
We evaluate on \textbf{LongMemEval-s}~\citep{wu2025longmemevalbenchmarkingchatassistants}, \textbf{PersonaMem-v2}~\citep{jiang2025personamemv2personalizedintelligencelearning}, and \textbf{MINT-Eval}~\citep{lee2026mintevalevaluatingmemorymultitarget}. Due to cost constraints, we evaluate 20\% of PersonaMem-v2 and 10\% of MINT-Eval. All subsets are selected with stratified sampling over question types. These benchmarks span direct recall to more complex queries involving aggregation or multi-interference reasoning. Appendix~\ref{apdx:benchmark_details} shows the full details.

\textbf{Baselines.}
We group the memory systems by how they organize what they store, and take open-source representatives of each group.
\catsq~\textbf{Flat store}: \textsc{A-Mem}~\citep{xu2026mem},
\textsc{SimpleMem}~\citep{liu2026simplememefficientlifelongmemory},
\textsc{MemOS}~\citep{li2025memosoperatingmemoryaugmentedgeneration}.
\catsq~\textbf{Hier/Lifecycle}: \textsc{MemoryOS}~\citep{kang2025memory}, \textsc{LightMem}~\citep{fang2026lightmem}, \textsc{OpenViking}~\citep{openviking2026}.
\catsq~\textbf{Graph-based}: 
\textsc{HippoRAG2}~\citep{gutierrez2025rag},
\textsc{PlugMem}~\citep{yang2026plugmem}, \textsc{MAGMA}~\citep{jiang2026magma}.
\catsq~\textbf{Trained policy}: \textsc{Mem-T}~\citep{yue2026mem}, \textsc{Memory-R2}~\citep{yan2026memory}.
We use RAG over the raw corpus as the main memory-free baseline, with both single-round (1rd) and two-round (2rd) variants. We also include several vanilla references, including No Context, Gold Context, etc. Appendix~\ref{apdx:baseline_details} shows the full details.

\textbf{Implementation details.}
We run every system on three backbones, \texttt{Qwen3.5-4B}, \texttt{Qwen3.6-27B} and \texttt{GPT-5.6-luna(low)}, sharing the same embedding model, \texttt{Qwen3-Embedding-0.6B}. Answers are scored by an LLM-as-Judge backed by \texttt{Qwen3.6-27B}, with judge prompts in Appendix~\ref{apdx:judge_prompts}. Alongside accuracy we report several cost metrics including token cost, end-to-end wall-clock time, number of LLM calls, and memory storage size. 

Since memory granularity and storage capacity vary substantially across systems, a fixed top-$k$ can correspond to very different amounts of retrieved information. For a fair comparison, we align retrieval budgets using \textbf{Retrieval Scope} (RS), defined as the tokens in the memory units exposed to the LLM from retrieval through answering, normalized by the total tokens within the system's retrievable store. We tune top-$k$ for each system (increasing it from the default in most cases) so its Retrieval Scope is roughly \textbf{no lower than} that of RAG (1-round), while roughly comparable across systems. The exception includes systems that contain only a few memory units, which retains a fixed top-$k$.
See full details in Appendix~\ref{apdx:eval_metrics} and \ref{apdx:implementation}.

\begin{figure*}[t]
\centering
\begin{minipage}[t]{0.46\textwidth}
\vspace{0pt}
\centering
\includegraphics[width=0.95\linewidth]{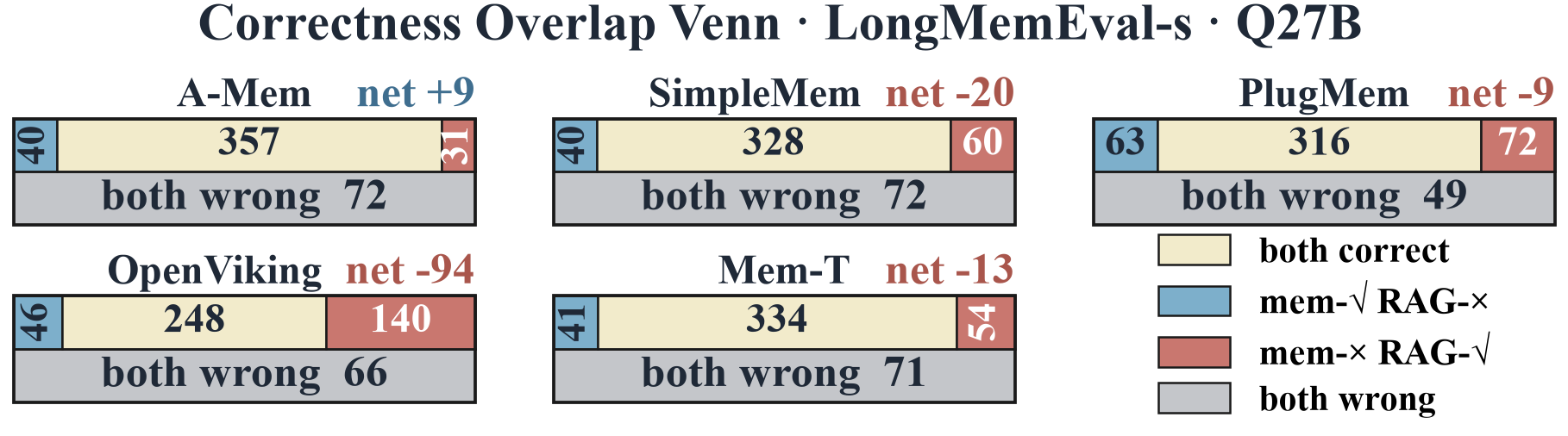}
\figsep
\includegraphics[width=0.95\linewidth]{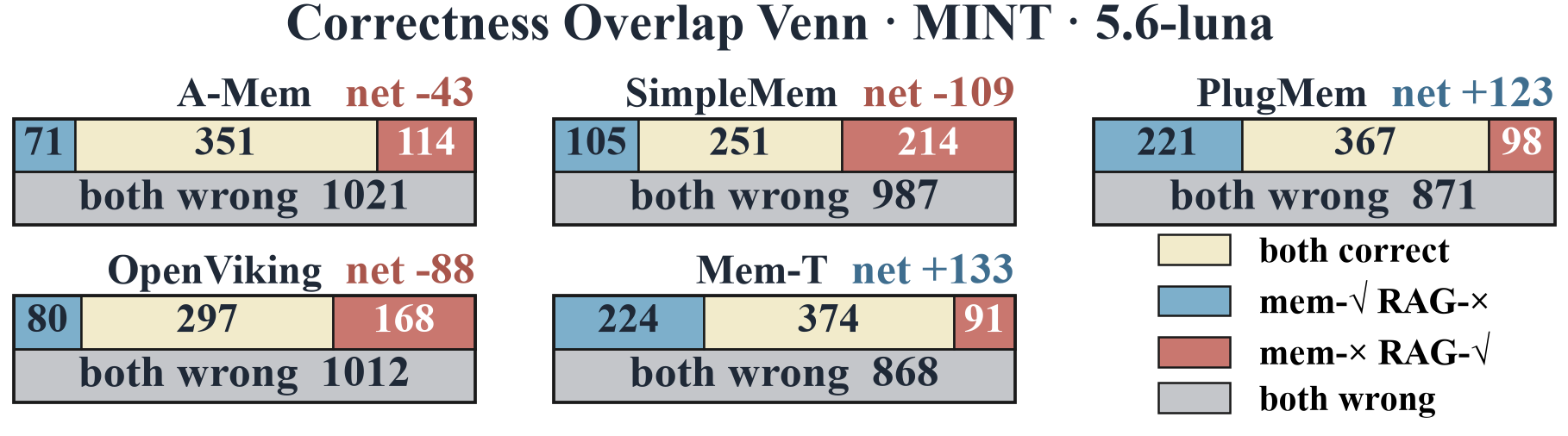}
\vspace{-0.6em}
\caption{Venn plot for per-query correction comparison between memory system and RAG (1rd).}
\label{fig:venn}
\end{minipage}\hfill
\begin{minipage}[t]{0.51\textwidth}
\vspace{0pt}
\centering
\includegraphics[width=\linewidth]{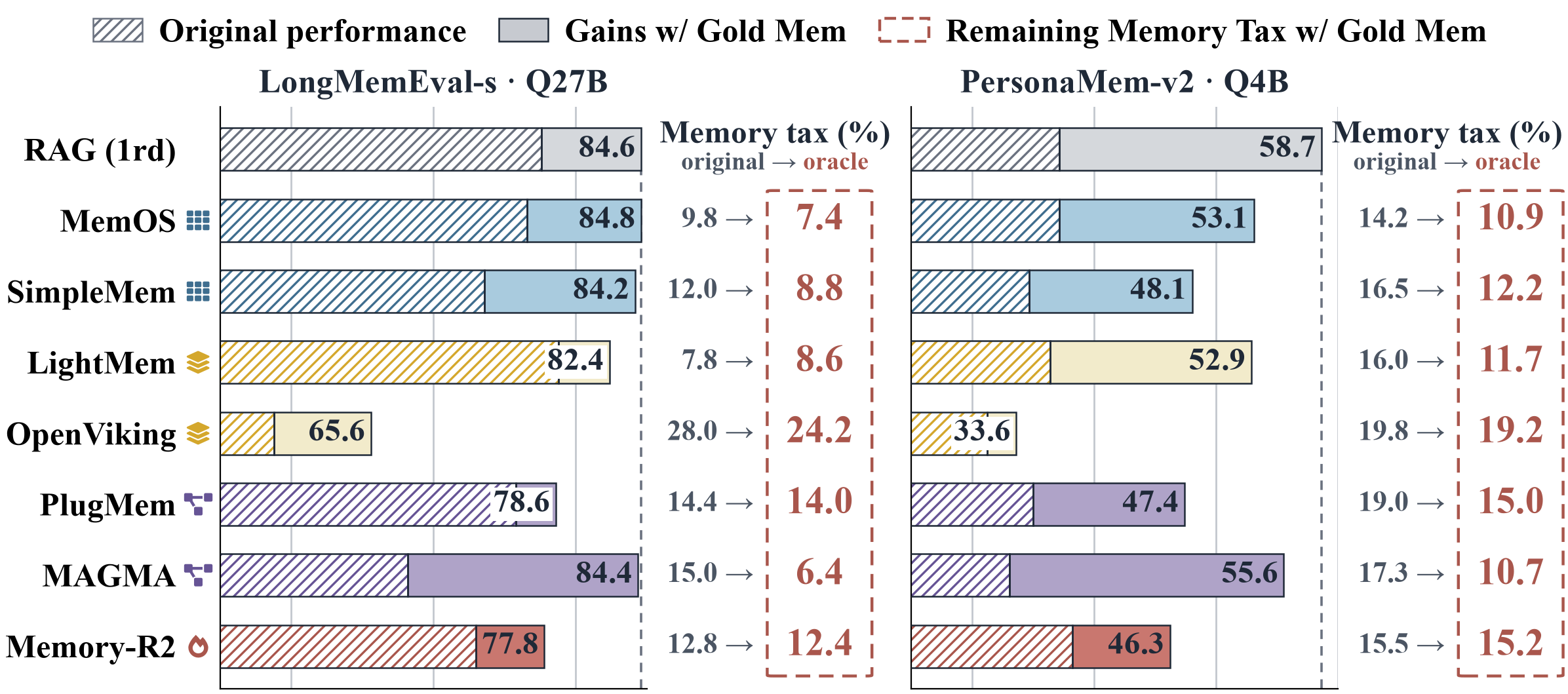}

\par\vspace{-0.6em}
\caption{Original performances vs.\ w/ gold memory units (bars), and the memory tax (quantified in scores) before and after having gold memory units (boxed numbers).}
\label{fig:oracle_compare}
\end{minipage}
\vspace{-1em}
\end{figure*}

\subsection{Results}
\label{sec:results}

Table~\ref{tab:main_results} reports the results of all the baselines on three benchmarks, and Figure~\ref{fig:pareto_27b} plots  the cost-performance trade-off more directly. Appendix~\ref{apdx:full_lme} to~\ref{apdx:full_mint} gives the full result. Given the \textbf{surprisingly strong performance} of naive RAG, we \textbf{further validate our baseline setup} in Appendix~\ref{apdx:rag_ablation} to~\ref{apdx:config_ablation}, including ablations on the RAG, reviewing RAG baselines in prior works, and validating the fairness of memory configurations. Overall, many memory systems fail to beat RAG in many cases, motivating a closer analysis in the next section.

\vspace{-0.1em}
\section{Memory Tax}
\label{sec:mem_tax}
\vspace{-0.1em}

\subsection{Overview}

\S\ref{sec:results} shows many memory systems underperform the naive RAG despite substantially higher costs. Since RAG accesses the same raw history with minimal processing, it provides a reference for assessing the \textbf{net value of building memory}. 

We dissect this net value through a per-query comparison with RAG (1rd) (Figure~\ref{fig:venn}). Memory does solve some queries RAG misses, benefiting from additional organization or processing, while losing others that RAG already solves, where evidence may become lost, obscured, or harder to access after memory processing. 
These two outcomes can be understood through the \textbf{inevitable deviations} introduced by memory from the raw history, as any nontrivial memory process can change how information is structured, preserved, or accessed.
These deviations give memory its \textbf{dual nature}. 
They are originally intended to better serve the tasks, yet may also have unintended harmful effects due to particular design choices, giving rise to \textbf{\textcolor{BenefitGreen}{Memory Benefit}} and \textbf{\textcolor{TaxRed}{Memory Tax}}.
Their balance determines the \textbf{net value} of memory, and for the above cases, the tax outweighs the benefit.
These quantities are reference-dependent, and throughout this work are defined relative to the naive RAG.

\begin{table}[t]
\definecolor{AFNoUnit}{HTML}{E4E6E9}
\centering
\caption{
Annotations for degree of information loss for sampled facts (\%): no loss (NL), partial loss (PL), major loss (ML), and no memory unit (NU).}
\vspace{-0.7em}
\label{tab:atomic_facts}
\small
\setlength{\tabcolsep}{2.6pt}
\renewcommand{\arraystretch}{0.92}
\begin{adjustbox}{max width=\linewidth}
\begin{tabular}{@{}c@{\hspace{2pt}}l ccc>{\columncolor{AFNoUnit}}c @{\hspace{8pt}} c@{\hspace{2pt}}l ccc>{\columncolor{AFNoUnit}}c@{}}
\toprule
\multicolumn{2}{l}{\textbf{Method}} & NL & PL & ML & NU & \multicolumn{2}{l}{\textbf{Method}} & NL & PL & ML & NU \\
\midrule
\clsFlat & \textsc{MemOS} & 85.8 & 9.0 & 2.4 & 2.8 & \clsHier & \textsc{OpenViking} & 39.4 & 9.6 & 13.0 & 38.0 \\
\clsFlat & \textsc{SimpleMem} & 65.6 & 27.6 & 2.2 & 4.6 & \clsGraph & \textsc{MAGMA} & 93.6 & 6.2 & 0.2 & 0.0 \\
\clsHier & \textsc{MemoryOS} & 98.4 & 0.0 & 0.0 & 1.6 & \clsGraph & \textsc{PlugMem} & 79.6 & 10.8 & 5.2 & 4.4 \\
\clsHier & \textsc{LightMem} & 76.2 & 13.6 & 0.6 & 9.6 & \clsTrain & \textsc{Memory-R2} & 72.6 & 14.8 & 6.0 & 6.6 \\

\bottomrule
\end{tabular}
\end{adjustbox}
\vspace{-1em}
\end{table}

\obsbox[obsbreakable]{\textit{Building memory has a dual nature and we call the two sides \textcolor{BenefitGreen}{Memory Benefit} and \textcolor{TaxRed}{Memory Tax}, with their balance defining net value of memory.}}

We further provide representative benefit and tax cases in Appendix~\ref{apdx:case_studies}. Memory benefit often appears when extraction, aggregation, preference inference or additional processing helps as shown in several typical types in Appendix~\ref{apdx:case_mem_benefit}, it largely aligns with the intended value of memory, while memory tax is underexplored. Its effects are visible in failed queries, but these outcomes do not explain how the losses arise within the memory. We next examine internal factors behind the memory tax.

\subsection{Memory Tax - \FaithHd{}: How Well Is the Information Preserved?}
\label{sec:faithfulness}

To trace where memory tax comes from, the first place to look is how well the memory system preserves past information. Memory operations like restructuring and compression may discard or distort evidence before retrieval begins, and such loss can not be recovered later, so we first investigate the \textbf{\faith{}} for a memory system, at both the macro level and micro level.

\begin{figure*}[t!]
\centering

\includegraphics[width=\textwidth]{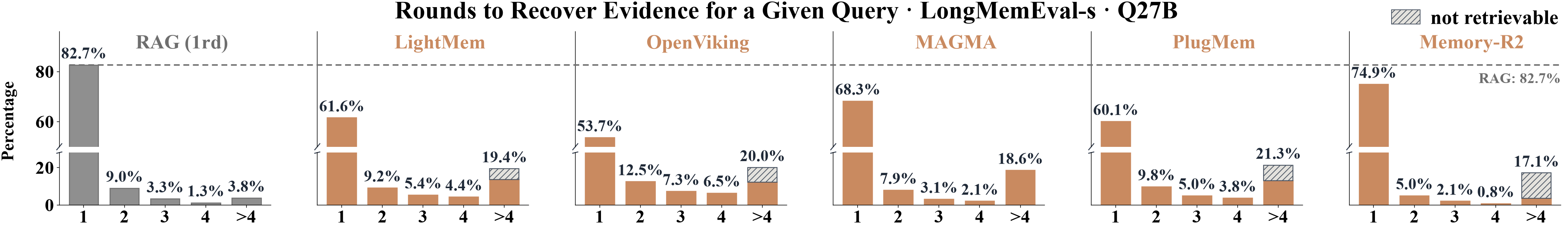}
\par\vspace{-0.6em}
\caption{Retrieval rounds replayed for the evidence to be fully retrieved, under raw chunks vs.\ memory units.}
\label{fig:access_rounds}
\vspace{6pt}
\includegraphics[width=\textwidth]{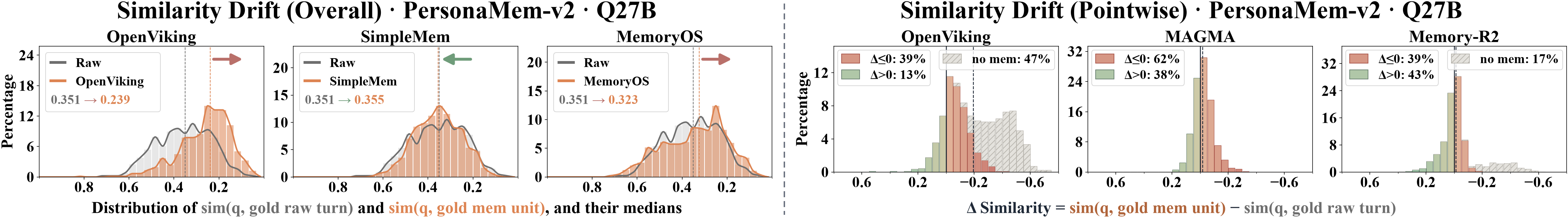}
\par\vspace{-0.6em}
\caption{Similarity comparison of \textbf{(query, gold raw turn)} vs.\ \textbf{\textcolor{SimMemOrange}{(query, gold memory unit)}}, with overall distribution shift (left) and point-wise similarity difference (right).}
\label{fig:accessibility}
\vspace{-1em}
\end{figure*}

For each query, we denote the raw conversation turns or steps with the evidence as \textbf{gold raw turns}, and the corresponding constructed memory units as \textbf{gold memory units} (null if the evidence is lost after construction). At the macro level, we \textbf{replay QA} with only the gold raw turns and, separately, only the gold memory units. Gold memory units remove retrieval failures, so the remaining memory tax reflects the losses introduced in memory construction, as shown by the red boxes in Figure~\ref{fig:oracle_compare}. At the micro level, we sample some atomic facts from the raw corpus and trace each fact to its corresponding memory units, annotating the degree of information loss, as shown in Table~\ref{tab:atomic_facts} and information loss does exist in many systems.  Appendix~\ref{apdx:faithfulness} and \ref{apdx:case_faithfulness} have more details and case studies.

\obsbox[obsbreakable]{\textit{Information loss with varying severity during the memory process becomes a direct cause of memory tax.}}

However, the reduced tax after having oracle memory also indicates degradation arises beyond construction. The performance shows a similar signal. Some systems (\textsc{MAGMA}, \textsc{LightMem}) achieve gold-mem scores close to gold-raw-turn scores, yet far above their original, suggesting the preserved evidence may become harder to access.

\subsection{Memory Tax - \AccessHd{}: How Easily Can Stored Information Be Accessed?}
\label{sec:accessibility}

\S\ref{sec:faithfulness} examines how much information survives memory construction. However, preserved information is useful only if it remains accessible at query time. So next we measure the \textbf{\access{}} of memory from both macro and micro levels.

At the macro level, we \textbf{replay retrieval iteratively} for multiple rounds over the constructed memory and record how many rounds are needed to recover all gold memory units for a given query. 
Evidence lost in construction is marked as not retrievable.
As shown in Figure~\ref{fig:access_rounds}, Memory systems recover fewer queries than RAG in the first round, suggesting that some relevant evidence is buried deeper after memory construction.

To understand why preserved evidence becomes harder to access, we further inspect some cases as the micro level analysis (Appendix~\ref{apdx:case_accessibility}). 
We observe several recurring patterns where evidence is \textbf{fragmented} across multiple memory units, \textbf{diluted} with unrelated information, or \textbf{rephrased}, weakening its accessibility compared with raw history.
To visualize this effect more broadly, we compare the (query, gold raw turn) similarity with the (query, gold memory unit) similarity, reporting the \textbf{point-wise changes and overall drift} in Figure~\ref{fig:accessibility}.
Since each system retrieves within its own memory store, absolute similarity values across systems are only indicative, the main signal is how memory construction shifts the query alignment of the same evidence.
Appendix~\ref{apdx:accessibility} provides more results.

\obsbox[obsbreakable]{\textit{Memory process can reduce evidence accessibility by fragmenting, diluting, or rephrasing it, creating memory tax even evidence is not lost.
}}

% At the micro level, we compare the (query, gold raw turn) similarity with the (query, gold memory unit) similarity, reporting the \textbf{point-wise changes and overall drift}. Since each system retrieves within its own memory store, absolute similarity values across systems are only indicative. The main signal is how construction shifts the query alignment of the same evidence, as shown in Figure~\ref{fig:accessibility}.

% Concretely, the results show that relevant evidence may require more retrieval rounds to be recalled, or become less aligned with the original query, even the information itself is preserved. Appendix~\ref{apdx:accessibility} and~\ref{apdx:case_accessibility} provides more results in different settings and also case studies showing what such accessibility degradation looks like.

\subsection{Memory Tax - \RobustHd{}: How Robust Is the Memory Across Backbones?}
\label{sec:robustness}

The above analyses examine memory under a fixed backbone setting. However, long-lived agents may construct memory with one model and later consume it with another, making backbone \textbf{robustness} another important property. We construct memory independently with three backbones and evaluate each memory with different QA backbones.

\begin{figure*}[t!]
\centering
\sbox0{\includegraphics[width=0.485\textwidth]{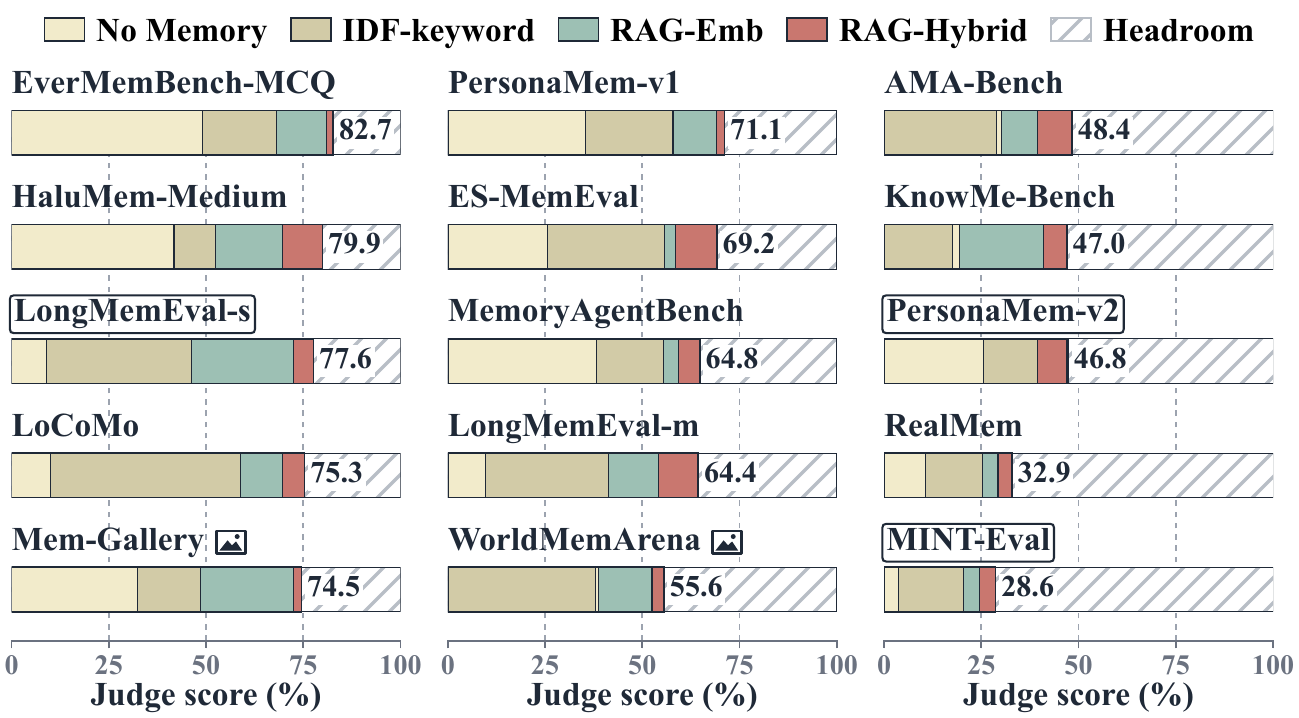}}%
\begin{minipage}[t]{0.485\textwidth}
\vspace{0pt}
\centering
\includegraphics[width=\linewidth]{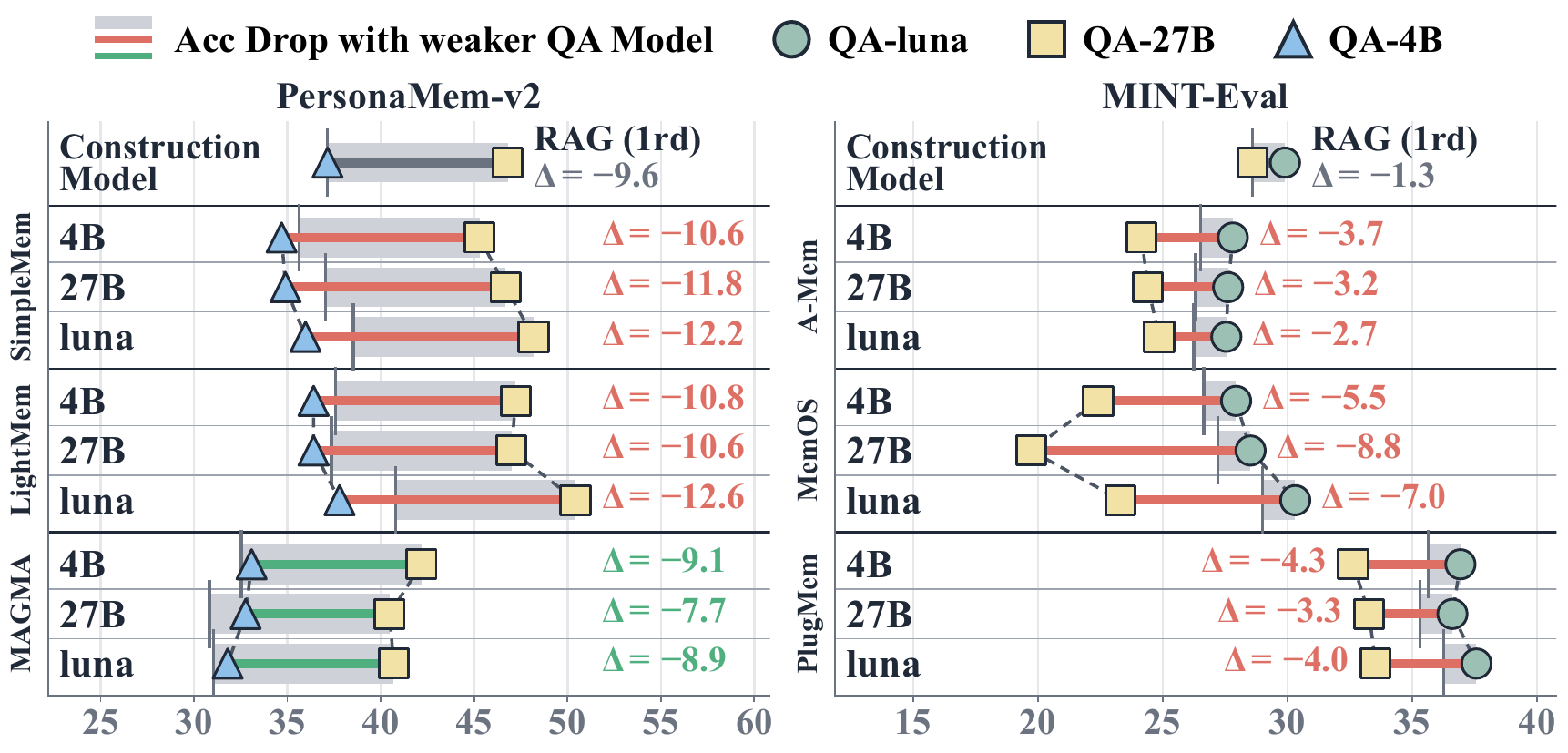}
\par\vspace{-0.6em}
\caption{Each row fixes the construction model, while different marker shapes denote different QA models. The segment shows Acc drop, \textcolor{TaxRed}{red} if the drop exceeds that of RAG (grey) and \textcolor{BenefitGreen}{green} otherwise.}
\label{fig:robustness}
\end{minipage}\hfill
\begin{minipage}[t]{0.485\textwidth}
\vspace{0pt}
\centering
\usebox0
\par\vspace{-0.6em}
\caption{Results of retrieval baselines on more memory benchmarks with Q27B. Details are in Appendix~\ref{apdx:regimes}.}
\label{fig:regimes}
\end{minipage}
\vspace{-1em}
\end{figure*}

Two patterns emerge. \boxednum{1} First, with the QA backbone fixed, changing the construction model leads to noticeable score variation. This is expected since many intermediate operations rely heavily on the backbones and can produce different memory items from the same raw history. \boxednum{2} More interestingly, with the construction fixed, changing the QA backbone causes larger performance variation than naive RAG for many systems, as shown in Figure~\ref{fig:robustness}.  This suggests many memory mechanisms can introduce \textbf{additional fragility} beyond the raw-history retrieval. In this sense, memory tax is reflected not only in point-wise performance loss, but also in the reduced robustness to backbone changes.
See Appendix~\ref{apdx:robustness} for more details. 

\obsbox[obsbreakable]{\textit{Memory process can amplify backbone sensitivity since model-dependent operations can introduce additional variability to raw history.}}

\subsection{Benchmark Headroom Further Exposes the Memory Tax}
\label{sec:regimes}

The analysis so far focuses on how memory tax arises within a memory system. Yet the balance between memory benefit and tax also depends on the task settings. Some tasks naturally favor direct recall from raw history, while others may benefit more from abstraction or distillation. To examine this evaluation-side effect, we run several naive baselines across a broader set of long-term memory benchmarks, including \boxednum{1} No Memory, \boxednum{2} IDF-keyword, a lexical matching baseline based on the keyword rarity, \boxednum{3} RAG-Emb, and \boxednum{4} RAG-Hybrid. Details are in Appendix~\ref{apdx:regimes}.

Figure~\ref{fig:regimes} shows that simple retrieval baselines already performs quite well on some of the benchmarks, including some widely-used ones. For example, the \textbf{extremely simplified IDF-keyword} based method can even achieve more than 50\% in absolute score on some of them, while RAG-Hybrid reaches about 70\%. Such results suggest some current evaluation settings are relatively retrieval-friendly, leaving quite \textbf{limited headroom} for heavy memory to improve over naive retrieval.

\obsbox[obsbreakable]{\textit{When naive RAG already solves many queries on certain benchmark, memory has less room for additional gains and even modest losses introduced by memory processing become more apparent in performance, making the tax more visible. 
}}

This also helps explain why some systems with lightweight or even no explicit memory construction can achieve strong performance and efficiency~\citep{yan2025general,
yu2026lazymemretrievebroadlyconstruct, 
xiao2026zeromemzerotokenmemoryoperations, xu2026chainofmemorylightweightmemoryconstruction, zhang2026learningqueryawarebudgettierrouting, wu2026basicsletconversationalagents}. When naive retrieval already provides a strong baseline, they \textbf{``inherit'' much of that performance while introducing little memory tax}, so lightweight design can be particularly competitive under limited benchmark headroom.

On the other hand, many real-world queries are also naturally retrieval-friendly, making such accurate recall tasks essential and meaningful in its own right. 
Current benchmarks contain a large share of such queries, and future evaluations could complement them with harder settings.
Moreover, reporting memory systems alongside naive RAG can \textbf{make remaining improvement room more explicit}, which also helps clarify the contribution of built memory. RAG scores can also serve as a reference for \textbf{calibrating difficulty and leaving meaningful room} for more reliable evaluations.

\vspace{-0.1em}
\section{Generalizing Memory Tax}
\label{sec:generalization}
\vspace{-0.1em}

% So far we have studied memory tax in \textbf{personal agent memory}. We next examine whether the same benefit-tax view extends to broader scenarios. So we study \textbf{experiential memory} and \textbf{world-model memory} as two representative settings.

Memory serves a wide range of application scenarios beyond personal-agent memory. We next examine whether the same benefit-tax view extends to two substantially different settings.

\subsection{Experiential Memory}
\label{sec:gen_experiential}

In this setting, agents accumulate experience from past task traces to guide future ones. We evaluate on several agentic scenarios covering embodied tasks (\textbf{ALFWorld}), web (\textbf{WebShop, Mind2Web}), coding (\textbf{BigCodeBench}) and tool use (\textbf{AppWorld, $\tau^2$-Bench}), with \texttt{Qwen3.5-4B} as backbone. Experience is accumulated on the training split and evaluated on the test split with memory frozen. We compare \textsc{Mem}$^p$~\citep{fang2026mempexploringagentprocedural}, \textsc{ReasoningBank}~\citep{ouyang2026reasoningbankscalingagentselfevolving}, \textsc{G-Memory}~\citep{zhang2025gmemorytracinghierarchicalmemory} and \textsc{SkillPro}~\citep{mi2026skillprolearningreusableskills} against a ReAct agent without memory and a naive RAG baseline, which retrieves the most similar past tasks by their descriptions and places the trajectories into the context, details are in Appendix~\ref{apdx:gen_experiential}.

 Figure~\ref{fig:generalization} shows the effect of memory varies across task domains. In web-based tasks, distilling long traces into workflows or strategies can outperform naive RAG, while for tool-use, several memory methods fall behind it, suggesting \textbf{memory tax also emerges in experiential memory}, but its magnitude depends strongly on the task settings.

\subsection{World-Model Memory}
\label{sec:gen_world_model}

Long-horizon world models need to retain information about past environment states to maintain temporal and scene consistency~\citep{zhang2026frame,li2026i3dm,he2026vgi,chen2026reworld,chen2026out,chen2026past}. Existing approaches have explored diverse memory representations and mechanisms, including raw frames, visual latents, and KV caches. We evaluate representative methods on two settings for long-horizon consistency: Context-as-Memory~\citep{yu2025context,king2026echo} and MemoBench~\citep{chen2026memobench}. Full details are in Appendix~\ref{apdx:gen_world_model}.

A similar pattern appears in these settings (Table~\ref{tab:world_model}). Directly retrieving and reinjecting past observations achieves strong consistency, while latent- or cache-based memories do not consistently outperform this simpler alternative. This suggests that \textbf{memory tax can also arise in world-model memory}, where additional memory operations may introduce losses that offset their intended benefits. It also motivates exploration of more unified memory carriers in video world models.

\begin{figure*}[t]
\centering
\includegraphics[width=\textwidth]{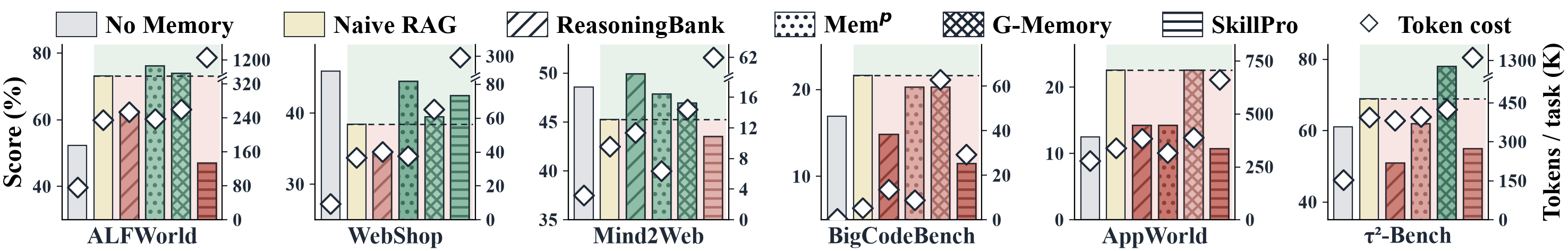}
\par\vspace{-0.7em}
\caption{Experiential memory results on agent task benchmarks, memory baselines are green when above RAG and red otherwise. See full version with memory tax part plotted in Appendix~\ref{apdx:gen_experiential}. }
\label{fig:generalization}
\vspace{-1em}
\end{figure*}

\obsbox[obsbreakable]{\textit{Memory process in a broader sense can also introduce both beneficial and harmful deviations from the original information, giving rise to memory benefit and tax, making the concepts generalizable across memory forms and scenarios.}}

Moreover, we also expect memory benefit/tax to extend beyond above settings along other axes~\citep{hu2026memoryageaiagents}, like \textit{omni-modality}, \textit{latents}, \textit{parameters}, etc. We leave the exploration for future work.

% 10-12 user: Table 3 with Fig 9 directly below it, one single-column float (Tables/world_model.tex)
\begin{table}[t]
\centering
\caption{Long-horizon consistency of world models with different memory representations and mechanisms. }
\vspace{-0.6em}
\label{tab:world_model}
\small
\setlength{\tabcolsep}{5pt}
\begin{adjustbox}{max width=\linewidth}
\begin{tabular}{l l cc cc}
\toprule
\multirow{2}{*}{\textbf{Method}} & \multirow{2}{*}{\begin{tabular}{@{}c@{}}\textbf{Mem.}\\\textbf{Form}\end{tabular}} & \multicolumn{2}{c}{\textbf{C-as-M-eval}} & \multicolumn{2}{c}{\textbf{MemoBench}} \\
\cmidrule(lr){3-4} \cmidrule(lr){5-6}
 & & SSIM $\uparrow$ & 1 - LPIPS $\uparrow$ & SSIM $\uparrow$ & 1 - LPIPS $\uparrow$ \\
\midrule
\textsc{I3DM} & naive frames & {\normalsize\textbf{.560}} & {\normalsize\textbf{.704}} & \underline{.430} & {\normalsize\textbf{.476}} \\
\arrayrulecolor{black!25}\specialrule{0.3pt}{1pt}{1pt}\arrayrulecolor{black}
\textsc{WorldKV} & raw KV & .372 & .459 & .379 & .420 \\
\textsc{ReWorld} & learned KV & .378 & \underline{.510} & .387 & \underline{.422} \\
\arrayrulecolor{black!25}\specialrule{0.3pt}{1pt}{1pt}\arrayrulecolor{black}
\textsc{MagicWorld} & visual latents & \underline{.388} & .341 & {\normalsize\textbf{.433}} & .412 \\
\textsc{FramePack} & visual latents & .217 & .312 & .279 & .303 \\
\arrayrulecolor{black!25}\specialrule{0.3pt}{1pt}{1pt}\arrayrulecolor{black}
\textsc{HyDRA} & mem. tokens & .320 & .319 & .355 & .326 \\
\bottomrule
\end{tabular}
\end{adjustbox}
\par\vspace{0.5em}
\includegraphics[width=\linewidth]{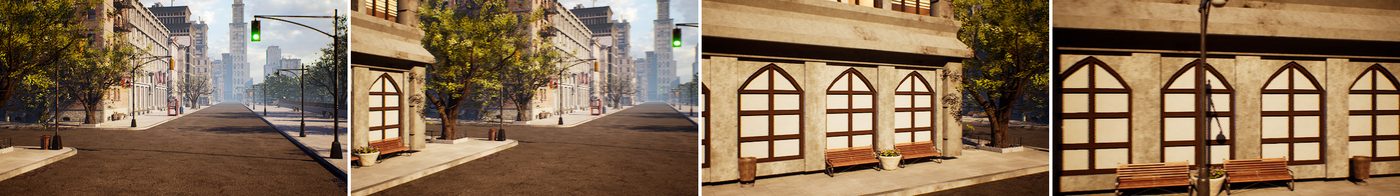}\\[2pt]
\includegraphics[width=\linewidth]{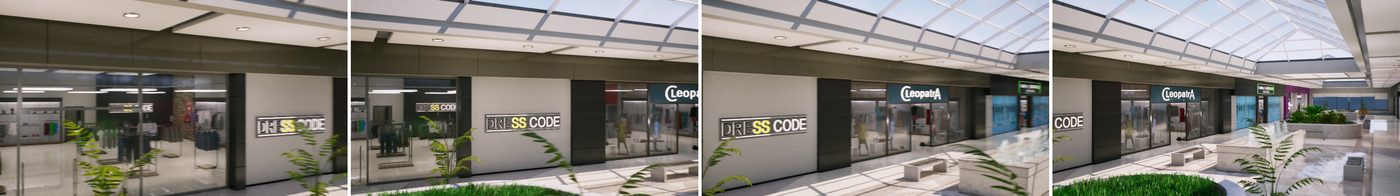}\\[2pt]
\includegraphics[width=\linewidth]{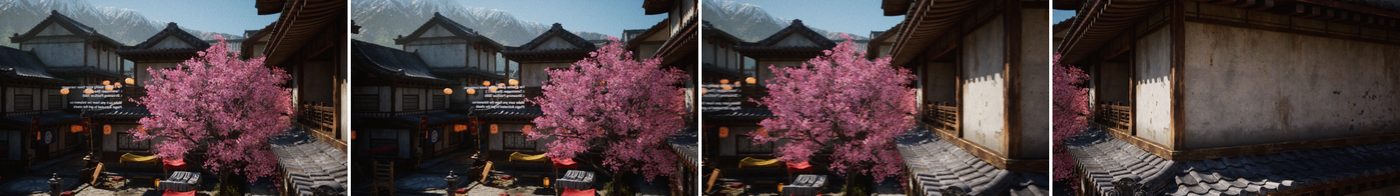}
\par\vspace{-0.6em}
\captionof{figure}{Example frame sequences from three evaluation scenes for testing long-horizon scene consistency.}
\label{fig:gen_video}
\vspace{-1em}
\end{table}

\vspace{-0.1em}
\section{Design Implications}
\label{sec:implications}
% 10-12 user: Table 4 waits until the page where Sec 6 starts is output, then floats to the top-left of the next page
% (independent of line breaks; Fig 9 keeps the right column next to Table 3)
\afterleftcol{% 10-12 user: Q27B only; under each benchmark an Accuracy and a Memory Tax column.
% 10-12 user: + rows shaded green, 14% + 31% x (improvement / largest improvement in the column), relative to the largest
% improvement in that column (acc: higher, tax: lower); no improvement -> light red background (10-12 user).
\begin{table}[t]
\centering
\caption{Accuracy and memory tax (\%) when applying the design implications, with \texttt{Qwen3.6-27B}.}
\vspace{-0.5em}
\label{tab:design_implications}
\small
\setlength{\tabcolsep}{4pt}
\renewcommand{\arraystretch}{1.1}   % 10-12 user: row spacing 1.1, this table only
\newcommand{\tg}[1]{\textcolor{BenefitGreen}{#1}}
\newcommand{\tr}[1]{\textcolor{TaxRed}{#1}}
\begin{adjustbox}{max width=0.95\linewidth}
\begin{tabular}{l cc cc}
\toprule
\multirow{2}{*}{\textbf{Method}} & \multicolumn{2}{c}{\textbf{LongMemEval-s}} & \multicolumn{2}{c}{\textbf{PersonaMem-v2}} \\
\cmidrule(lr){2-3} \cmidrule(lr){4-5}
 & \textbf{Acc} ($\uparrow$) & \textbf{Tax} ($\downarrow$) & \textbf{Acc} ($\uparrow$) & \textbf{Tax} ($\downarrow$) \\
\midrule
\rowcolor{black!7}\textsc{SimpleMem} & 73.6 & 12.0 & 46.7 & 11.2 \\
\arrayrulecolor{black!25}\specialrule{0.3pt}{1pt}{1pt}\arrayrulecolor{black}
\quad + src-aug & \cellcolor{BenefitGreen!45}82.8 & \cellcolor{BenefitGreen!45}5.2 & \cellcolor{BenefitGreen!27}49.6 & \cellcolor{BenefitGreen!29}7.2 \\
\quad + recursive & N/A$^\dagger$ & N/A$^\dagger$ & \cellcolor{BenefitGreen!16}46.8 & \cellcolor{TaxRed!22}12.1 \\
\midrule
\rowcolor{black!7}\textsc{MemoryOS} & 73.4 & 11.4 & 48.2 & 17.1 \\
\arrayrulecolor{black!25}\specialrule{0.3pt}{1pt}{1pt}\arrayrulecolor{black}
\quad + src-aug & \cellcolor{BenefitGreen!24}76.4 & \cellcolor{BenefitGreen!30}7.8 & \cellcolor{BenefitGreen!36}53.0 & \cellcolor{BenefitGreen!45}8.5 \\
\quad + recursive & N/A$^\dagger$ & N/A$^\dagger$ & \cellcolor{BenefitGreen!18}49.1 & \cellcolor{BenefitGreen!16}16.4 \\
\midrule
\rowcolor{black!7}\textsc{Mem-T} & 75.0 & 10.8 & 42.1 & 15.2 \\
\arrayrulecolor{black!25}\specialrule{0.3pt}{1pt}{1pt}\arrayrulecolor{black}
\quad + src-aug & \cellcolor{BenefitGreen!31}80.0 & \cellcolor{BenefitGreen!40}5.2 & \cellcolor{BenefitGreen!45}49.0 & \cellcolor{BenefitGreen!37}8.7 \\
\quad + recursive & N/A$^\dagger$ & N/A$^\dagger$ & \cellcolor{BenefitGreen!30}45.7 & \cellcolor{BenefitGreen!19}13.6 \\
\bottomrule
\end{tabular}
\end{adjustbox}
\vspace{-0.2em}
\par\vspace{2pt}\raggedright\footnotesize $^\dagger$LongMemEval-s has one query per corpus, so inter-query mechanisms do not apply.
\vspace{-1.2em}
\end{table}

}
\vspace{-0.1em}

In this section, we derive some simple yet effective design implications for \textbf{reducing or compensating for} the memory tax. We focus on improvements that can be used on existing memory systems without introducing another new architecture, and deliberately \textbf{exclude common test-time scaling techniques} like retrieval depth / width expansion, query refinement, using stronger backbones, etc. 

\obsbox[obsbreakable]{
6.1 \textit{(Memory Construction)} \textbf{Make memory deviations more recoverable. }}

As we discussed above, building memory can introduce changes and deviations from the raw history about how information is preserved and accessed through diverse memory operations. While no single design can predict and fit all tasks needs, and the harmful deviations are difficult to eliminate entirely, a more promising way to reduce memory tax is to make such deviations \textbf{recoverable}, preserving a lightweight path back to the source history when memory becomes insufficient.
As a simple proof of concept, we retrieve from source history alongside normal memory retrieval and jointly rerank them before QA. This lightweight recovery path can restore evidence lost during construction or missed in access. Results in Table~\ref{tab:design_implications} show consistent improvements and reduction of memory tax across several memory systems.

\obsbox[obsbreakable]{6.2 \textit{(Memory Access)} \textbf{Recursive or self-referential memory for retrieval.}
}

Memory tax can only be visible in actual use, when relevant evidences become harder to access and are recovered only after deep search. Such retrieval experience directly reflects the harmful deviations memory introduced, and can in turn provide feedback to improve future memory access, creating a \textbf{recursive} loop between memory access and construction.  
Inspired by progress on recursive self-improvement (RSI)~\citep{zhang2026darwin, karten2026primeagentselfimprovingrlm, zhang2026hyperagents}, where a system is expected to \textbf{``improve how it improves''}, memory can similarly \textbf{``remember how it remembers''}. 
Past queries, evidences, and retrieval paths \textbf{can themselves be source materials} for memory, and this is particularly useful for multi-round retrieval, since evidence found in deeper rounds can then be accessed directly by similar information needs. 
It also offers a new dimension of test-time scaling at query-side, allowing inter-query scaling beyond a single query.
As a simple proof of concept, we treat past queries and retrieved evidence as additional inputs into memory store, making them available for future retrieval. Results in Table~\ref{tab:design_implications} also validate its effectiveness.

\obsbox[obsbreakable]{6.3 \textit{(Future Direction and Beyond Memory Operations)} \textbf{Try modeling future demand beyond past information.}
}

Many methods build memory \textbf{ahead of time (AoT)} before future queries come. As we discussed before, memory makes deviations on the raw history, with the intention of better serving task queries, but may also add implicitly harmful effects. Without knowing future demand accurately, it is unclear what kind of deviations will be beneficial. To cover a broader range of possible queries, AoT systems may need to process more information in more elaborate ways, making it increasingly expensive.
The opposite is \textbf{just-in-time (JiT)} memory~\citep{yan2025generalagenticmemorydeep,
zhou2026justintimememorylearningcurate,
yan2026justintimeagentmemoryruntime}, which delays memory operations until queries arrive. With query already known, JiT can choose more targeted forms and directions of deviation to maximize the memory benefit and minimize the memory tax, but moves computation to the online part and increases user-facing latency, to which the users are highly sensitive.

A middle ground is to \textbf{model the future demand}. 
In personal agent settings, task queries are not purely random. 
Past interests, habits, and access patterns provide a prior over what information is likely to matter, and hence what kinds of deviations that memory introduces are likely useful. 
Such signals may help construction allocate effort more selectively, and reduces harmful deviations, without paying the full online cost of JiT. From this view, memory can partly be seen as an \textbf{implicit recommendation system} that estimates future relevance of past information. 
Exploring this perspective in real-world personal-agent settings may be a promising direction for future exploration.

\vspace{-0.1em}
\section{Related Works}
\label{sec:related}
\vspace{-0.1em}

\paragraph{Long-Term Agent Memory.}
Long-term memory has become an important component for long-horizon agents, supporting the retention of specific information, past interactions, and task experience. Existing systems explore a variety of memory representations, including relatively flat stores~\citep{liu2026simplememefficientlifelongmemory,yang-etal-2026-grounding,meng2026ensi}, hierarchical organizations, from short-/long-term memory layers~\citep{kang2025memory,fang2026lightmem} to filesystem-tree designs in recent industrial systems~\citep{openviking2026, tencentagentmemory2026}, and graph-based representations~\citep{jiang2026magma,yang2026plugmem}. Another line of variation lies in how memory operations are governed. Some systems rely on predefined harness for writing, updating, and using memories, while others allow an LLM agent to decide these operations dynamically~\citep{wang2025mirixmultiagentmemoryllmbased,xu2026mem,everos2026} or explore learned memory policies~\citep{yan2026memoryr1enhancinglargelanguage,yu2026memagentreshapinglongcontextllm,yan2026memory}. 
In this work, we study these memory systems through the lens of their \textbf{net value}, examining the gains and the losses introduced by memory operations.

\paragraph{Empirical Studies of Agent Memory.}
Recent empirical studies examine agent memory from different perspectives, including how memory systems are designed and deployed, their performance-cost trade-offs, and where computation and time are spent~\citep{jiang2026anatomyagenticmemorytaxonomy,omri2026agentmemorycharacterizationimplications,zhou2026readyagentnativememorysystem,huang2026harnessmemoryholisticevaluation,Wolff_2026}. These works mainly characterize existing memory designs and their empirical behavior, but our work focuses on the net value introduced by memory processing itself. We introduce memory tax to characterize its losses relative to direct retrieval, analyze how such losses arise from both system and evaluation sides, and derive system-agnostic design implications.
\vspace{-0.1em}
\section{Conclusion}
\label{sec:conclusion}
\vspace{-0.1em}

We revisit the agent memory by studying its net value over direct retrieval from raw history. 
Across the evaluation on three benchmarks with many retrieval-friendly queries alongside some complex ones, many memory methods underperform naive RAG, indicating a negative net value in these settings. 
We dissect this net value into memory benefit and tax, showing memory can improve how past information is organized and used, but may also have harmful effects. 
We analyze memory tax from both system sides and evaluation settings, and generalize it to broader scenarios like experiential memory and world-model memory. 
We hope memory tax can shift more attention from standalone performance to the net value of memory, and encourages more reliable memory design and evaluation.

% *ACL venues require a Limitations section after the conclusion.
\section*{Limitations}
Given the rapid development of this field, our baseline selection is representative rather than exhaustive. We also follow official configurations wherever possible, though further system-specific tuning may yield better performance.
Our benchmarks are primarily static, offline, and English-language, leaving long-lived deployments with interleaved queries and multilingual interactions underexplored. Extending the analysis to other forms of memory, such as omnimodal, latent, and parametric memory, is also left for future work.
Our evaluation is limited by computational and API-calling cost. We do not include stronger frontier models beyond \texttt{GPT-5.6 Luna (low)}, and use stratified subsets for some benchmarks (20\% of PersonaMem-v2, 10\% of MINT-Eval, and 20\% LongMemEval-s questions for GPT-5.6 Luna). 

\bibliography{references}

\begin{thebibliography}{89}
\providecommand{\natexlab}[1]{#1}

\bibitem[{Barres et~al.(2025)Barres, Dong, Ray, Si, and Narasimhan}]{barres2025tau2benchevaluatingconversationalagents}
Victor Barres, Honghua Dong, Soham Ray, Xujie Si, and Karthik Narasimhan. 2025.
\newblock \href {https://arxiv.org/abs/2506.07982} {$\tau^2$-bench: Evaluating conversational agents in a dual-control environment}.
\newblock \emph{Preprint}, arXiv:2506.07982.

\bibitem[{Bei et~al.(2026)Bei, Wei, Ning, Zhao, Liu, Lin, Zhu, Hamann, He, and Tong}]{bei2026memgallerybenchmarkingmultimodallongterm}
Yuanchen Bei, Tianxin Wei, Xuying Ning, Yanjun Zhao, Zhining Liu, Xiao Lin, Yada Zhu, Hendrik Hamann, Jingrui He, and Hanghang Tong. 2026.
\newblock \href {https://arxiv.org/abs/2601.03515} {Mem-gallery: Benchmarking multimodal long-term conversational memory for mllm agents}.
\newblock \emph{Preprint}, arXiv:2601.03515.

\bibitem[{Bian et~al.(2026)Bian, Yao, Hu, Xu, Zhang, Guo, Yang, Han, Wang, and Chen}]{bian2026realmembenchmarkingllmsrealworld}
Haonan Bian, Zhiyuan Yao, Sen Hu, Zishan Xu, Shaolei Zhang, Yifu Guo, Ziliang Yang, Xueran Han, Huacan Wang, and Ronghao Chen. 2026.
\newblock \href {https://arxiv.org/abs/2601.06966} {Realmem: Benchmarking llms in real-world memory-driven interaction}.
\newblock \emph{Preprint}, arXiv:2601.06966.

\bibitem[{Chen et~al.(2026{\natexlab{a}})Chen, Niu, Li, Liu, Zheng, Tang, Li, Xiong, and Li}]{chen2026halumemevaluatinghallucinationsmemory}
Ding Chen, Simin Niu, Kehang Li, Peng Liu, Xiangping Zheng, Bo~Tang, Xinchi Li, Feiyu Xiong, and Zhiyu Li. 2026{\natexlab{a}}.
\newblock \href {https://arxiv.org/abs/2511.03506} {Halumem: Evaluating hallucinations in memory systems of agents}.
\newblock \emph{Preprint}, arXiv:2511.03506.

\bibitem[{Chen et~al.(2026{\natexlab{b}})Chen, Zhou, Hua, Zhang, Qian, Ma, Chen, Liu, Zhao, Wang et~al.}]{chen2026memobench}
Haoyu Chen, Kaichen Zhou, Hang Hua, Kaile Zhang, Jingwen Qian, Wufei Ma, Haonan Chen, Chunjiang Liu, Yizhou Zhao, Xiaoyuan Wang, et~al. 2026{\natexlab{b}}.
\newblock Memobench: Benchmarking world modeling in dynamically changing environments.
\newblock In \emph{European Conference on Computer Vision}, pages 658--676. Springer.

\bibitem[{Chen et~al.(2026{\natexlab{c}})Chen, Guo, Lan, Shu, Zhang, Hu, Yao, Zhang, Zhang, Rao et~al.}]{chen2026past}
Harold~Haodong Chen, Rongjin Guo, Disen Lan, Wen-Jie Shu, Hongfei Zhang, Hanzhe Hu, Shengtao Yao, Zixin Zhang, Guibin Zhang, Zhefan Rao, et~al. 2026{\natexlab{c}}.
\newblock The past frames the future: Memory for autoregressive video generation.
\newblock \emph{arXiv preprint arXiv:2609.28466}.

\bibitem[{Chen et~al.(2026{\natexlab{d}})Chen, Liang, Zhou, Ding, Liu, Wan, and Bai}]{chen2026out}
Kaijin Chen, Dingkang Liang, Xin Zhou, Yikang Ding, Xiaoqiang Liu, Pengfei Wan, and Xiang Bai. 2026{\natexlab{d}}.
\newblock Out of sight but not out of mind: Hybrid memory for dynamic video world models.
\newblock \emph{arXiv preprint arXiv:2603.25716}.

\bibitem[{Chen et~al.(2026{\natexlab{e}})Chen, Lu, Shen, and Zhang}]{Chen_2026}
Tiantian Chen, Jiaqi Lu, Ying Shen, and Lin Zhang. 2026{\natexlab{e}}.
\newblock \href {https://doi.org/10.1145/3774904.3792143} {Es-memeval: Benchmarking conversational agents on personalized long-term emotional support}.
\newblock In \emph{Proceedings of the ACM Web Conference 2026}, page 5810–5821. ACM.

\bibitem[{Chen et~al.(2026{\natexlab{f}})Chen, Wang, Shen, Yan, Yang, Xu, Du, Wang, Gui, Huang et~al.}]{chen2026reworld}
Zhifei Chen, Luozhou Wang, Guibao Shen, Dongyu Yan, Shuai Yang, Tianshuo Xu, Yihua Du, Wei Wang, Tianyi Gui, Lianghua Huang, et~al. 2026{\natexlab{f}}.
\newblock Reworld: An interactive world model with long-horizon memory.
\newblock \emph{arXiv preprint arXiv:2608.23565}.

\bibitem[{Chhikara et~al.(2025{\natexlab{a}})Chhikara, Khant, Aryan, Singh, and Yadav}]{chhikara2025mem0}
Prateek Chhikara, Dev Khant, Saket Aryan, Taranjeet Singh, and Deshraj Yadav. 2025{\natexlab{a}}.
\newblock Mem0: Building production-ready ai agents with scalable long-term memory.
\newblock \emph{arXiv preprint arXiv:2504.19413}.

\bibitem[{Chhikara et~al.(2025{\natexlab{b}})Chhikara, Khant, Aryan, Singh, and Yadav}]{chhikara2025mem0buildingproductionreadyai}
Prateek Chhikara, Dev Khant, Saket Aryan, Taranjeet Singh, and Deshraj Yadav. 2025{\natexlab{b}}.
\newblock \href {https://arxiv.org/abs/2504.19413} {Mem0: Building production-ready ai agents with scalable long-term memory}.
\newblock \emph{Preprint}, arXiv:2504.19413.

\bibitem[{Deng et~al.(2023)Deng, Gu, Zheng, Chen, Stevens, Wang, Sun, and Su}]{deng2023mind2webgeneralistagentweb}
Xiang Deng, Yu~Gu, Boyuan Zheng, Shijie Chen, Samuel Stevens, Boshi Wang, Huan Sun, and Yu~Su. 2023.
\newblock \href {https://arxiv.org/abs/2306.06070} {Mind2web: Towards a generalist agent for the web}.
\newblock \emph{Preprint}, arXiv:2306.06070.

\bibitem[{Eslami et~al.(2026)Eslami, Gaiduk, Krimmel, Milliken, Wang, and Bykov}]{eslami2026diffusionpretraineddensecontextualembeddings}
Sedigheh Eslami, Maksim Gaiduk, Markus Krimmel, Louis Milliken, Bo~Wang, and Denis Bykov. 2026.
\newblock \href {https://arxiv.org/abs/2602.11151} {Diffusion-pretrained dense and contextual embeddings}.
\newblock \emph{Preprint}, arXiv:2602.11151.

\bibitem[{{EverMind-AI}(2026)}]{everos2026}
{EverMind-AI}. 2026.
\newblock {EverOS}: One portable memory layer for every ai agent: local-first, markdown-native, user-owned, and self-evolving across apps, tools, and workflows.
\newblock \url{https://github.com/EverMind-AI/EverOS}.

\bibitem[{Fang et~al.(2026{\natexlab{a}})Fang, Deng, Xu, Jiang, Tang, Xu, Deng, Yao, Wang, Qiao et~al.}]{fang2026lightmem}
Jizhan Fang, Xinle Deng, Haoming Xu, Ziyan Jiang, Yuqi Tang, Ziwen Xu, Shumin Deng, Yunzhi Yao, Mengru Wang, Shuofei Qiao, et~al. 2026{\natexlab{a}}.
\newblock Lightmem: Lightweight and efficient memory-augmented generation.
\newblock In \emph{International Conference on Learning Representations}, volume 2026, pages 98706--98729.

\bibitem[{Fang et~al.(2026{\natexlab{b}})Fang, Liang, Wang, Wu, Qiao, Xie, Huang, Chen, and Zhang}]{fang2026mempexploringagentprocedural}
Runnan Fang, Yuan Liang, Xiaobin Wang, Jialong Wu, Shuofei Qiao, Pengjun Xie, Fei Huang, Huajun Chen, and Ningyu Zhang. 2026{\natexlab{b}}.
\newblock \href {https://arxiv.org/abs/2508.06433} {Memp: Exploring agent procedural memory}.
\newblock \emph{Preprint}, arXiv:2508.06433.

\bibitem[{Gur et~al.(2024)Gur, Furuta, Huang, Safdari, Matsuo, Eck, and Faust}]{gur2024real}
Izzeddin Gur, Hiroki Furuta, Austin Huang, Mustafa Safdari, Yutaka Matsuo, Douglas Eck, and Aleksandra Faust. 2024.
\newblock A real-world webagent with planning, long context understanding, and program synthesis.
\newblock In \emph{International Conference on Learning Representations}, volume 2024, pages 52690--52717.

\bibitem[{Guti{\'e}rrez et~al.(2025)Guti{\'e}rrez, Shu, Qi, Zhou, and Su}]{gutierrez2025rag}
Bernal~Jim{\'e}nez Guti{\'e}rrez, Yiheng Shu, Weijian Qi, Sizhe Zhou, and Yu~Su. 2025.
\newblock From rag to memory: Non-parametric continual learning for large language models.
\newblock \emph{arXiv preprint arXiv:2502.14802}.

\bibitem[{He et~al.(2026)He, Wei, Cheng, Ma, Zhang, Li, Wen, Liu, Hao, Cai et~al.}]{he2026vgi}
Xuan He, Cong Wei, Yuhao Cheng, Linrui Ma, Yuxuan Zhang, Zuojun Li, Yuhao Wen, Zeyi Liu, Yuren Hao, Songcheng Cai, et~al. 2026.
\newblock Vgi-bench: Probing visual intelligence in video generation models.
\newblock \emph{arXiv preprint arXiv:2608.19583}.

\bibitem[{Hu et~al.(2026{\natexlab{a}})Hu, Li, Gao, Chen, Bai, Xu, Lin, Li, Han, Pei, and Deng}]{hu2026evaluatinglonghorizonmemorymultiparty}
Chuanrui Hu, Tong Li, Xingze Gao, Hongda Chen, Yi~Bai, Dannong Xu, Tianwei Lin, Xiaohong Li, Yunyun Han, Jian Pei, and Yafeng Deng. 2026{\natexlab{a}}.
\newblock \href {https://arxiv.org/abs/2602.01313} {Evaluating long-horizon memory for multi-party collaborative dialogues}.
\newblock \emph{Preprint}, arXiv:2602.01313.

\bibitem[{Hu et~al.(2025{\natexlab{a}})Hu, Chen, Chen, Mu, Shao, and Luo}]{hu-etal-2025-hiagent}
Mengkang Hu, Tianxing Chen, Qiguang Chen, Yao Mu, Wenqi Shao, and Ping Luo. 2025{\natexlab{a}}.
\newblock \href {https://doi.org/10.18653/v1/2025.acl-long.1575} {{H}i{A}gent: Hierarchical working memory management for solving long-horizon agent tasks with large language model}.
\newblock In \emph{Proceedings of the 63rd Annual Meeting of the Association for Computational Linguistics (Volume 1: Long Papers)}, pages 32779--32798, Vienna, Austria. Association for Computational Linguistics.

\bibitem[{Hu et~al.(2025{\natexlab{b}})Hu, Wang, and McAuley}]{hu2025evaluatingmemoryllmagents}
Yuanzhe Hu, Yu~Wang, and Julian McAuley. 2025{\natexlab{b}}.
\newblock \href {https://arxiv.org/abs/2507.05257} {Evaluating memory in llm agents via incremental multi-turn interactions}.
\newblock \emph{Preprint}, arXiv:2507.05257.

\bibitem[{Hu et~al.(2026{\natexlab{b}})Hu, Liu, Yue, Zhang, Liu, Zhu, Lin, Guo, Dou, Xi, Jin, Tan, Yin, Liu, Zhang, Sun, Zhu, Sun, Peng, Cheng, Fan, Guo, Yu, Zhou, Hu, Huo, Wang, Niu, Wang, Yin, Hu, Liao, Li, Wang, Zhou, Liu, Cheng, Zhang, Gui, Pan, Zhang, Torr, Dou, Wen, Huang, Jiang, and Yan}]{hu2026memoryageaiagents}
Yuyang Hu, Shichun Liu, Yanwei Yue, Guibin Zhang, Boyang Liu, Fangyi Zhu, Jiahang Lin, Honglin Guo, Shihan Dou, Zhiheng Xi, Senjie Jin, Jiejun Tan, Yanbin Yin, Jiongnan Liu, Zeyu Zhang, Zhongxiang Sun, Yutao Zhu, Hao Sun, Boci Peng, and 28 others. 2026{\natexlab{b}}.
\newblock \href {https://arxiv.org/abs/2512.13564} {Memory in the age of ai agents}.
\newblock \emph{Preprint}, arXiv:2512.13564.

\bibitem[{Huang et~al.(2026)Huang, Zhang, Wu, Chen, Jiang, Yang, Yang, Zou, Zhang, Wu, Wu, Chang, Yu, Liu, and Caliskan}]{huang2026harnessmemoryholisticevaluation}
Wei-Chieh Huang, Weizhi Zhang, Yuchen Wu, Yankai Chen, Eric~Hanchen Jiang, Wooseong Yang, Yiwei Yang, Henry~Peng Zou, Hanrong Zhang, Ying~Nian Wu, Haolun Wu, Kai-Wei Chang, Philip~S. Yu, Xue Liu, and Aylin Caliskan. 2026.
\newblock \href {https://arxiv.org/abs/2608.15008} {Harness the memory: A holistic evaluation of memory substrates in memory agents}.
\newblock \emph{Preprint}, arXiv:2608.15008.

\bibitem[{Jiang et~al.(2025{\natexlab{a}})Jiang, Hao, Cho, Li, Yuan, Chen, Ungar, Taylor, and Roth}]{jiang2025knowmerespondme}
Bowen Jiang, Zhuoqun Hao, Young-Min Cho, Bryan Li, Yuan Yuan, Sihao Chen, Lyle Ungar, Camillo~J. Taylor, and Dan Roth. 2025{\natexlab{a}}.
\newblock \href {https://arxiv.org/abs/2504.14225} {Know me, respond to me: Benchmarking llms for dynamic user profiling and personalized responses at scale}.
\newblock \emph{Preprint}, arXiv:2504.14225.

\bibitem[{Jiang et~al.(2025{\natexlab{b}})Jiang, Yuan, Shen, Hao, Xu, Chen, Liu, Vijjini, He, Yu, Poovendran, Wornell, Ungar, Roth, Chen, and Taylor}]{jiang2025personamemv2personalizedintelligencelearning}
Bowen Jiang, Yuan Yuan, Maohao Shen, Zhuoqun Hao, Zhangchen Xu, Zichen Chen, Ziyi Liu, Anvesh~Rao Vijjini, Jiashu He, Hanchao Yu, Radha Poovendran, Gregory Wornell, Lyle Ungar, Dan Roth, Sihao Chen, and Camillo~Jose Taylor. 2025{\natexlab{b}}.
\newblock \href {https://arxiv.org/abs/2512.06688} {Personamem-v2: Towards personalized intelligence via learning implicit user personas and agentic memory}.
\newblock \emph{Preprint}, arXiv:2512.06688.

\bibitem[{Jiang et~al.(2026{\natexlab{a}})Jiang, Li, Li, and Li}]{jiang2026magma}
Dongming Jiang, Yi~Li, Guanpeng Li, and Bingzhe Li. 2026{\natexlab{a}}.
\newblock Magma: A multi-graph based agentic memory architecture for ai agents.
\newblock In \emph{Proceedings of the 64th Annual Meeting of the Association for Computational Linguistics (Volume 1: Long Papers)}, pages 36848--36865.

\bibitem[{Jiang et~al.(2026{\natexlab{b}})Jiang, Li, Wei, Yang, Kishore, Zhao, Kang, Hu, Chen, Li, and Li}]{jiang2026anatomyagenticmemorytaxonomy}
Dongming Jiang, Yi~Li, Songtao Wei, Jinxin Yang, Ayushi Kishore, Alysa Zhao, Dingyi Kang, Xu~Hu, Feng Chen, Qiannan Li, and Bingzhe Li. 2026{\natexlab{b}}.
\newblock \href {https://arxiv.org/abs/2602.19320} {Anatomy of agentic memory: Taxonomy and empirical analysis of evaluation and system limitations}.
\newblock \emph{Preprint}, arXiv:2602.19320.

\bibitem[{Kang et~al.(2025)Kang, Ji, Zhao, and Bai}]{kang2025memory}
Jiazheng Kang, Mingming Ji, Zhe Zhao, and Ting Bai. 2025.
\newblock Memory os of ai agent.
\newblock In \emph{Proceedings of the 2025 Conference on Empirical Methods in Natural Language Processing}, pages 25972--25981.

\bibitem[{Karten et~al.(2026)Karten, Zhang, Thomas, Müller, Bakouch, Auras, Senghaas, Obeid, Dunas, Hagemann, and Jaghouar}]{karten2026primeagentselfimprovingrlm}
Seth Karten, Alex~L. Zhang, Kevin Thomas, Sebastian Müller, Elie Bakouch, Daniel Auras, Mika Senghaas, Fares Obeid, Konstantin Dunas, Johannes Hagemann, and Sami Jaghouar. 2026.
\newblock \href {https://arxiv.org/abs/2608.23552} {Prime agent: A self-improving rlm harness}.
\newblock \emph{Preprint}, arXiv:2608.23552.

\bibitem[{King et~al.(2026)King, Xue, Bian, Huang, Li, Li, Su, Li, Wang, Zhang et~al.}]{king2026echo}
Wayne King, Zeyue Xue, Yuxuan Bian, Jie Huang, Haoran Li, Yaowei Li, Yaofeng Su, Yuming Li, Haoyu Wang, Shiyi Zhang, et~al. 2026.
\newblock Echo-memory: A controlled study of memory in action world models.
\newblock \emph{arXiv preprint arXiv:2606.09803}.

\bibitem[{Lee et~al.(2025)Lee, Roy, Xu, Raiman, Shoeybi, Catanzaro, and Ping}]{lee2025nvembedimprovedtechniquestraining}
Chankyu Lee, Rajarshi Roy, Mengyao Xu, Jonathan Raiman, Mohammad Shoeybi, Bryan Catanzaro, and Wei Ping. 2025.
\newblock \href {https://arxiv.org/abs/2405.17428} {Nv-embed: Improved techniques for training llms as generalist embedding models}.
\newblock \emph{Preprint}, arXiv:2405.17428.

\bibitem[{Lee et~al.(2026)Lee, Chen, Singh, Khan, Stengel-Eskin, and Bansal}]{lee2026mintevalevaluatingmemorymultitarget}
Hyunji Lee, Justin Chih-Yao Chen, Joykirat Singh, Zaid Khan, Elias Stengel-Eskin, and Mohit Bansal. 2026.
\newblock \href {https://arxiv.org/abs/2605.18565} {Minteval: Evaluating memory under multi-target interference in long-horizon agent systems}.
\newblock \emph{Preprint}, arXiv:2605.18565.

\bibitem[{Li et~al.(2025{\natexlab{a}})Li, Li, Chen, Hu, Zhao, and Jiang}]{li2025magicworld}
Guangyuan Li, Bo~Li, Jinwei Chen, Xiaobin Hu, Lei Zhao, and Peng-Tao Jiang. 2025{\natexlab{a}}.
\newblock Magicworld: Towards long-horizon stability for interactive video world exploration.
\newblock \emph{arXiv preprint arXiv:2511.18886}.

\bibitem[{Li et~al.(2026)Li, Yan, Chen, Li, Song, Wang, Cai, Wong, and Ji}]{li2026i3dm}
Jia Li, Han Yan, Yihang Chen, Siqi Li, Xibin Song, Yifu Wang, Jianfei Cai, Tien-Tsin Wong, and Pan Ji. 2026.
\newblock I3dm: Implicit 3d-aware memory retrieval and injection for consistent video scene generation.
\newblock \emph{arXiv preprint arXiv:2603.23413}.

\bibitem[{Li et~al.(2025{\natexlab{b}})Li, Song, Wang, Niu, Chen, Yang, Xi, Lai, Zhao, Wang, Ren, Lin, Huo, Chen, Chen, Li, Yin, Yu, Tang, Yang, Xu, and Xiong}]{li2025memosoperatingmemoryaugmentedgeneration}
Zhiyu Li, Shichao Song, Hanyu Wang, Simin Niu, Ding Chen, Jiawei Yang, Chenyang Xi, Huayi Lai, Jihao Zhao, Yezhaohui Wang, Junpeng Ren, Zehao Lin, Jiahao Huo, Tianyi Chen, Kai Chen, Kehang Li, Zhiqiang Yin, Qingchen Yu, Bo~Tang, and 3 others. 2025{\natexlab{b}}.
\newblock \href {https://arxiv.org/abs/2505.22101} {Memos: An operating system for memory-augmented generation (mag) in large language models}.
\newblock \emph{Preprint}, arXiv:2505.22101.

\bibitem[{Liu et~al.(2026{\natexlab{a}})Liu, Yang, Pu, Liu, Long, Guo, Chen, Weng, Kochkina, Kaur, Smiley, Liu, Zou, Liu, Bu, Peng, and Wang}]{liu2026worldmemarenaevaluatingmultimodalagent}
Chengzhi Liu, Yuzhe Yang, Sophia~Xiao Pu, Yepeng Liu, Lin Long, Yichen Guo, Nuo Chen, Zhaotian Weng, Elena Kochkina, Simerjot Kaur, Charese Smiley, Xiaomo Liu, James Zou, Sheng Liu, Yuheng Bu, Songyou Peng, and Xin~Eric Wang. 2026{\natexlab{a}}.
\newblock \href {https://arxiv.org/abs/2605.29341} {Worldmemarena: Evaluating multimodal agent memory through action-world interaction}.
\newblock \emph{Preprint}, arXiv:2605.29341.

\bibitem[{Liu et~al.(2026{\natexlab{b}})Liu, Su, Xia, Han, Zheng, Xie, Ding, and Yao}]{liu2026simplememefficientlifelongmemory}
Jiaqi Liu, Yaofeng Su, Peng Xia, Siwei Han, Zeyu Zheng, Cihang Xie, Mingyu Ding, and Huaxiu Yao. 2026{\natexlab{b}}.
\newblock \href {https://arxiv.org/abs/2601.02553} {Simplemem: Efficient lifelong memory for llm agents}.
\newblock \emph{Preprint}, arXiv:2601.02553.

\bibitem[{Liu et~al.(2025)Liu, Li, Xiao, Shao, and Lian}]{liu2025llama2vecunsupervisedadaptationlarge}
Zheng Liu, Chaofan Li, Shitao Xiao, Yingxia Shao, and Defu Lian. 2025.
\newblock \href {https://arxiv.org/abs/2312.15503} {Llama2vec: Unsupervised adaptation of large language models for dense retrieval}.
\newblock \emph{Preprint}, arXiv:2312.15503.

\bibitem[{Ma et~al.(2026)Ma, Nan, Wu, and Chen}]{ma2026deservesmemoryadaptivememory}
Wenquan Ma, Jiayan Nan, Wenlong Wu, and Yize Chen. 2026.
\newblock \href {https://arxiv.org/abs/2508.03341} {What deserves memory: Adaptive memory distillation for llm agents}.
\newblock \emph{Preprint}, arXiv:2508.03341.

\bibitem[{Maharana et~al.(2024)Maharana, Lee, Tulyakov, Bansal, Barbieri, and Fang}]{maharana2024evaluatinglongtermconversationalmemory}
Adyasha Maharana, Dong-Ho Lee, Sergey Tulyakov, Mohit Bansal, Francesco Barbieri, and Yuwei Fang. 2024.
\newblock \href {https://arxiv.org/abs/2402.17753} {Evaluating very long-term conversational memory of llm agents}.
\newblock \emph{Preprint}, arXiv:2402.17753.

\bibitem[{Meng et~al.(2026)Meng, Sun, Parekh, and Han}]{meng2026ensi}
Xuanyu Meng, Jiashuo Sun, Jash~Rajesh Parekh, and Jiawei Han. 2026.
\newblock Ensi-rag: Entity-structure-indexed retrieval-augmented generation for long-document question answering.
\newblock \emph{arXiv preprint arXiv:2608.21252}.

\bibitem[{Mi et~al.(2026)Mi, Ma, Yang, Li, Wang, Zhang, and Wang}]{mi2026skillprolearningreusableskills}
Qirui Mi, Zhijian Ma, Mengyue Yang, Haoxuan Li, Yisen Wang, Haifeng Zhang, and Jun Wang. 2026.
\newblock \href {https://arxiv.org/abs/2602.01869} {Skill-pro: Learning reusable skills from experience via non-parametric ppo for llm agents}.
\newblock \emph{Preprint}, arXiv:2602.01869.

\bibitem[{Omri et~al.(2026)Omri, Gan, Broveak, Geens, He, Pentland, Verhelst, Weissman, and Tambe}]{omri2026agentmemorycharacterizationimplications}
Yasmine Omri, Ziyu Gan, Zachary Broveak, Robin Geens, Zexue He, Alex Pentland, Marian Verhelst, Tsachy Weissman, and Thierry Tambe. 2026.
\newblock \href {https://arxiv.org/abs/2606.06448} {Agent memory: Characterization and system implications of stateful long-horizon workloads}.
\newblock \emph{Preprint}, arXiv:2606.06448.

\bibitem[{Ouyang et~al.(2026)Ouyang, Yan, Hsu, Chen, Jiang, Wang, Han, Le, Daruki, Tang, Tirumalashetty, Lee, Rofouei, Lin, Han, Lee, and Pfister}]{ouyang2026reasoningbankscalingagentselfevolving}
Siru Ouyang, Jun Yan, I-Hung Hsu, Yanfei Chen, Ke~Jiang, Zifeng Wang, Rujun Han, Long~T. Le, Samira Daruki, Xiangru Tang, Vishy Tirumalashetty, George Lee, Mahsan Rofouei, Hangfei Lin, Jiawei Han, Chen-Yu Lee, and Tomas Pfister. 2026.
\newblock \href {https://arxiv.org/abs/2509.25140} {Reasoningbank: Scaling agent self-evolving with reasoning memory}.
\newblock \emph{Preprint}, arXiv:2509.25140.

\bibitem[{Packer et~al.(2024)Packer, Wooders, Lin, Fang, Patil, Stoica, and Gonzalez}]{packer2024memgptllmsoperatingsystems}
Charles Packer, Sarah Wooders, Kevin Lin, Vivian Fang, Shishir~G. Patil, Ion Stoica, and Joseph~E. Gonzalez. 2024.
\newblock \href {https://arxiv.org/abs/2310.08560} {Memgpt: Towards llms as operating systems}.
\newblock \emph{Preprint}, arXiv:2310.08560.

\bibitem[{Rasmussen et~al.(2025)Rasmussen, Paliychuk, Beauvais, Ryan, and Chalef}]{rasmussen2025zeptemporalknowledgegraph}
Preston Rasmussen, Pavlo Paliychuk, Travis Beauvais, Jack Ryan, and Daniel Chalef. 2025.
\newblock \href {https://arxiv.org/abs/2501.13956} {Zep: A temporal knowledge graph architecture for agent memory}.
\newblock \emph{Preprint}, arXiv:2501.13956.

\bibitem[{Shridhar et~al.(2021)Shridhar, Yuan, Côté, Bisk, Trischler, and Hausknecht}]{shridhar2021alfworldaligningtextembodied}
Mohit Shridhar, Xingdi Yuan, Marc-Alexandre Côté, Yonatan Bisk, Adam Trischler, and Matthew Hausknecht. 2021.
\newblock \href {https://arxiv.org/abs/2010.03768} {Alfworld: Aligning text and embodied environments for interactive learning}.
\newblock \emph{Preprint}, arXiv:2010.03768.

\bibitem[{{Tencent-Cloud}(2026)}]{tencentagentmemory2026}
{Tencent-Cloud}. 2026.
\newblock {TencentDB-Agent-Memory}: Tencentdb agent memory is a team-level memory hub for ai agents - turning conversations, docs, and code into four reusable memory assets (chat memory, skill, llm-wiki, code-graph) that are governed, shared, and equipped across agents and frameworks..
\newblock \url{https://github.com/TencentCloud/TencentDB-Agent-Memory}.

\bibitem[{Trivedi et~al.(2024)Trivedi, Khot, Hartmann, Manku, Dong, Li, Gupta, Sabharwal, and Balasubramanian}]{trivedi2024appworldcontrollableworldapps}
Harsh Trivedi, Tushar Khot, Mareike Hartmann, Ruskin Manku, Vinty Dong, Edward Li, Shashank Gupta, Ashish Sabharwal, and Niranjan Balasubramanian. 2024.
\newblock \href {https://arxiv.org/abs/2407.18901} {Appworld: A controllable world of apps and people for benchmarking interactive coding agents}.
\newblock \emph{Preprint}, arXiv:2407.18901.

\bibitem[{{Volcengine}(2026)}]{openviking2026}
{Volcengine}. 2026.
\newblock {OpenViking}: An open-source context database for ai agents.
\newblock \url{https://github.com/volcengine/OpenViking}.

\bibitem[{Wang et~al.(2025)Wang, Li, and Xiao}]{wang2025jinarerankerv3lateinteractiondocument}
Feng Wang, Yuqing Li, and Han Xiao. 2025.
\newblock \href {https://arxiv.org/abs/2509.25085} {jina-reranker-v3: Last but not late interaction for document reranking}.
\newblock \emph{Preprint}, arXiv:2509.25085.

\bibitem[{Wang et~al.(2023)Wang, Xie, Jiang, Mandlekar, Xiao, Zhu, Fan, and Anandkumar}]{wang2023voyager}
Guanzhi Wang, Yuqi Xie, Yunfan Jiang, Ajay Mandlekar, Chaowei Xiao, Yuke Zhu, Linxi Fan, and Anima Anandkumar. 2023.
\newblock Voyager: An open-ended embodied agent with large language models.
\newblock \emph{arXiv preprint arXiv:2305.16291}.

\bibitem[{Wang et~al.(2020)Wang, Wei, Dong, Bao, Yang, and Zhou}]{wang2020minilmdeepselfattentiondistillation}
Wenhui Wang, Furu Wei, Li~Dong, Hangbo Bao, Nan Yang, and Ming Zhou. 2020.
\newblock \href {https://arxiv.org/abs/2002.10957} {Minilm: Deep self-attention distillation for task-agnostic compression of pre-trained transformers}.
\newblock \emph{Preprint}, arXiv:2002.10957.

\bibitem[{Wang and Chen(2025)}]{wang2025mirixmultiagentmemoryllmbased}
Yu~Wang and Xi~Chen. 2025.
\newblock \href {https://arxiv.org/abs/2507.07957} {Mirix: Multi-agent memory system for llm-based agents}.
\newblock \emph{Preprint}, arXiv:2507.07957.

\bibitem[{Wolff and Bennati(2026)}]{Wolff_2026}
Benedict Wolff and Jacopo Bennati. 2026.
\newblock \href {https://doi.org/10.1109/compsac69091.2026.00015} {Cost and accuracy of long-term memory in distributed multi-agent systems based on large language models}.
\newblock In \emph{2026 IEEE 50th Annual Computers, Software, and Applications Conference (COMPSAC)}, page 11–21. IEEE.

\bibitem[{Wu et~al.(2025)Wu, Wang, Yu, Zhang, Chang, and Yu}]{wu2025longmemevalbenchmarkingchatassistants}
Di~Wu, Hongwei Wang, Wenhao Yu, Yuwei Zhang, Kai-Wei Chang, and Dong Yu. 2025.
\newblock \href {https://arxiv.org/abs/2410.10813} {Longmemeval: Benchmarking chat assistants on long-term interactive memory}.
\newblock \emph{Preprint}, arXiv:2410.10813.

\bibitem[{Wu et~al.(2026{\natexlab{a}})Wu, Chen, Weng, Wang, Li, Zhang, Hu, Wu, Lan, Wang, and Chen}]{wu2026knowmebenchbenchmarkingpersonunderstanding}
Tingyu Wu, Zhisheng Chen, Ziyan Weng, Shuhe Wang, Chenglong Li, Shuo Zhang, Sen Hu, Silin Wu, Qizhen Lan, Huacan Wang, and Ronghao Chen. 2026{\natexlab{a}}.
\newblock \href {https://arxiv.org/abs/2601.04745} {Knowme-bench: Benchmarking person understanding for lifelong digital companions}.
\newblock \emph{Preprint}, arXiv:2601.04745.

\bibitem[{Wu et~al.(2026{\natexlab{b}})Wu, Chen, Huang, Chen, Liu, Wang, Zhou, and Liang}]{wu2026basicsletconversationalagents}
Yuqian Wu, Wei Chen, Zhengjun Huang, Junle Chen, Qingxiang Liu, Kai Wang, Xiaofang Zhou, and Yuxuan Liang. 2026{\natexlab{b}}.
\newblock \href {https://arxiv.org/abs/2604.11628} {Back to basics: Let conversational agents remember with just retrieval and generation}.
\newblock \emph{Preprint}, arXiv:2604.11628.

\bibitem[{Xiao et~al.(2024)Xiao, Liu, Zhang, Muennighoff, Lian, and Nie}]{xiao2024cpackpackedresourcesgeneral}
Shitao Xiao, Zheng Liu, Peitian Zhang, Niklas Muennighoff, Defu Lian, and Jian-Yun Nie. 2024.
\newblock \href {https://arxiv.org/abs/2309.07597} {C-pack: Packed resources for general chinese embeddings}.
\newblock \emph{Preprint}, arXiv:2309.07597.

\bibitem[{Xiao et~al.(2026)Xiao, Zhu, Zhang, Chen, Hong, Zhuang, Zhang, Chen, Ouyang, Ren, and Huang}]{xiao2026zeromemzerotokenmemoryoperations}
Yilin Xiao, Zhehan Zhu, Yujing Zhang, Jin Chen, Zijin Hong, Luyao Zhuang, Qinggang Zhang, Shengyuan Chen, Xiaocao Ouyang, Lingfei Ren, and Xiao Huang. 2026.
\newblock \href {https://arxiv.org/abs/2607.29377} {Zero-mem: Zero-token memory operations for llm agents}.
\newblock \emph{Preprint}, arXiv:2607.29377.

\bibitem[{Xie et~al.(2024)Xie, Zhang, Chen, Li, Zhao, Cao, Hua, Cheng, Shin, Lei et~al.}]{xie2024osworld}
Tianbao Xie, Danyang Zhang, Jixuan Chen, Xiaochuan Li, Siheng Zhao, Ruisheng Cao, Toh~J Hua, Zhoujun Cheng, Dongchan Shin, Fangyu Lei, et~al. 2024.
\newblock Osworld: Benchmarking multimodal agents for open-ended tasks in real computer environments.
\newblock \emph{Advances in Neural Information Processing Systems}, 37:52040--52094.

\bibitem[{Xu et~al.(2026{\natexlab{a}})Xu, Liang, Mei, Gao, Tan, and Zhang}]{xu2026mem}
Wujiang Xu, Zujie Liang, Kai Mei, Hang Gao, Juntao Tan, and Yongfeng Zhang. 2026{\natexlab{a}}.
\newblock A-mem: Agentic memory for llm agents.
\newblock \emph{Advances in Neural Information Processing Systems}, 38:17577--17604.

\bibitem[{Xu et~al.(2026{\natexlab{b}})Xu, Xu, Tian, Huang, Chen, Li, and Shen}]{xu2026chainofmemorylightweightmemoryconstruction}
Xiucheng Xu, Bingbing Xu, Xueyun Tian, Zihe Huang, Rongxin Chen, Yunfan Li, and Huawei Shen. 2026{\natexlab{b}}.
\newblock \href {https://arxiv.org/abs/2601.14287} {Chain-of-memory: Lightweight memory construction with dynamic evolution for llm agents}.
\newblock \emph{Preprint}, arXiv:2601.14287.

\bibitem[{Yan et~al.(2025{\natexlab{a}})Yan, Li, Qian, Lu, and Liu}]{yan2025generalagenticmemorydeep}
B.~Y. Yan, Chaofan Li, Hongjin Qian, Shuqi Lu, and Zheng Liu. 2025{\natexlab{a}}.
\newblock \href {https://arxiv.org/abs/2511.18423} {General agentic memory via deep research}.
\newblock \emph{Preprint}, arXiv:2511.18423.

\bibitem[{Yan et~al.(2026{\natexlab{a}})Yan, Li, Qian, Lu, Li, and Liu}]{yan2026justintimeagentmemoryruntime}
Bingyu Yan, Chaofan Li, Hongjin Qian, Shuqi Lu, Chaozhuo Li, and Zheng Liu. 2026{\natexlab{a}}.
\newblock \href {https://arxiv.org/abs/2609.34385} {Just-in-time agent memory with runtime agentic research}.
\newblock \emph{Preprint}, arXiv:2609.34385.

\bibitem[{Yan et~al.(2025{\natexlab{b}})Yan, Li, Qian, Lu, and Liu}]{yan2025general}
BY~Yan, Chaofan Li, Hongjin Qian, Shuqi Lu, and Zheng Liu. 2025{\natexlab{b}}.
\newblock General agentic memory via deep research.
\newblock \emph{arXiv preprint arXiv:2511.18423}.

\bibitem[{Yan et~al.(2026{\natexlab{b}})Yan, Bahloul, Nie, Schwarzmann, Trivisonno, Tresp, and Ma}]{yan2026memory}
Sikuan Yan, Ahmed Bahloul, Ercong Nie, Susanna Schwarzmann, Riccardo Trivisonno, Volker Tresp, and Yunpu Ma. 2026{\natexlab{b}}.
\newblock Memory-r2: Fair credit assignment for long-horizon memory-augmented llm agents.
\newblock \emph{arXiv preprint arXiv:2605.21768}.

\bibitem[{Yan et~al.(2026{\natexlab{c}})Yan, Yang, Huang, Nie, Ding, Li, Ma, Bi, Kersting, Pan, Schütze, Tresp, and Ma}]{yan2026memoryr1enhancinglargelanguage}
Sikuan Yan, Xiufeng Yang, Zuchao Huang, Ercong Nie, Zifeng Ding, Zonggen Li, Xiaowen Ma, Jinhe Bi, Kristian Kersting, Jeff~Z. Pan, Hinrich Schütze, Volker Tresp, and Yunpu Ma. 2026{\natexlab{c}}.
\newblock \href {https://arxiv.org/abs/2508.19828} {Memory-r1: Enhancing large language model agents to manage and utilize memories via reinforcement learning}.
\newblock \emph{Preprint}, arXiv:2508.19828.

\bibitem[{Yang et~al.(2026{\natexlab{a}})Yang, Chen, He, Jiang, Galley, Wang, Gao, Han, and Zhai}]{yang2026plugmem}
Ke~Yang, Zixi Chen, Xuan He, Jize Jiang, Michel Galley, Chenglong Wang, Jianfeng Gao, Jiawei Han, and ChengXiang Zhai. 2026{\natexlab{a}}.
\newblock Plugmem: A task-agnostic plugin memory module for llm agents.
\newblock \emph{arXiv preprint arXiv:2603.03296}.

\bibitem[{Yang et~al.(2026{\natexlab{b}})Yang, Jiang, Jiang, Kargupta, Zhang, and Han}]{yang-etal-2026-grounding}
Ruozhen Yang, Yucheng Jiang, Yueqi Jiang, Priyanka Kargupta, Yunyi Zhang, and Jiawei Han. 2026{\natexlab{b}}.
\newblock \href {https://doi.org/10.18653/v1/2026.findings-acl.584} {Grounding agent memory in contextual intent}.
\newblock In \emph{Findings of the {A}ssociation for {C}omputational {L}inguistics: {ACL} 2026}, pages 12008--12042, San Diego, California, United States. Association for Computational Linguistics.

\bibitem[{Yao et~al.(2023{\natexlab{a}})Yao, Chen, Yang, and Narasimhan}]{yao2023webshopscalablerealworldweb}
Shunyu Yao, Howard Chen, John Yang, and Karthik Narasimhan. 2023{\natexlab{a}}.
\newblock \href {https://arxiv.org/abs/2207.01206} {Webshop: Towards scalable real-world web interaction with grounded language agents}.
\newblock \emph{Preprint}, arXiv:2207.01206.

\bibitem[{Yao et~al.(2023{\natexlab{b}})Yao, Zhao, Yu, Du, Shafran, Narasimhan, and Cao}]{yao2023reactsynergizingreasoningacting}
Shunyu Yao, Jeffrey Zhao, Dian Yu, Nan Du, Izhak Shafran, Karthik Narasimhan, and Yuan Cao. 2023{\natexlab{b}}.
\newblock \href {https://arxiv.org/abs/2210.03629} {React: Synergizing reasoning and acting in language models}.
\newblock \emph{Preprint}, arXiv:2210.03629.

\bibitem[{Yi et~al.(2026)Yi, Kim, Cho, Jang, Yun, and Kim}]{yi2026worldkv}
Jung Yi, Minjae Kim, Paul~Hyunbin Cho, Wooseok Jang, Sangdoo Yun, and Seungryong Kim. 2026.
\newblock Worldkv: Efficient world memory with world retrieval and compression.
\newblock \emph{arXiv preprint arXiv:2605.22718}.

\bibitem[{Yu et~al.(2026{\natexlab{a}})Yu, Chen, Feng, Chen, Dai, Yu, Zhang, Ma, Liu, Wang, and Zhou}]{yu2026memagentreshapinglongcontextllm}
Hongli Yu, Tinghong Chen, Jiangtao Feng, Jiangjie Chen, Weinan Dai, Qiying Yu, Ya-Qin Zhang, Wei-Ying Ma, Jingjing Liu, Mingxuan Wang, and Hao Zhou. 2026{\natexlab{a}}.
\newblock \href {https://arxiv.org/abs/2507.02259} {Memagent: Reshaping long-context llm with multi-conv rl-based memory agent}.
\newblock \emph{Preprint}, arXiv:2507.02259.

\bibitem[{Yu et~al.(2026{\natexlab{b}})Yu, Zhao, Zhang, and Li}]{yu2026lazymemretrievebroadlyconstruct}
Jing Yu, Yibo Zhao, Jiaming Zhang, and Xiang Li. 2026{\natexlab{b}}.
\newblock \href {https://arxiv.org/abs/2607.22690} {Lazymem: Retrieve broadly, construct selectively for efficient long-term agent memory}.
\newblock \emph{Preprint}, arXiv:2607.22690.

\bibitem[{Yu et~al.(2025)Yu, Bai, Qin, Liu, Wang, Wan, Zhang, and Liu}]{yu2025context}
Jiwen Yu, Jianhong Bai, Yiran Qin, Quande Liu, Xintao Wang, Pengfei Wan, Di~Zhang, and Xihui Liu. 2025.
\newblock Context as memory: Scene-consistent interactive long video generation with memory retrieval.
\newblock In \emph{Proceedings of the SIGGRAPH Asia 2025 Conference Papers}, pages 1--11.

\bibitem[{Yue et~al.(2026)Yue, Peng, Fan, Guo, Li, and Zhang}]{yue2026mem}
Yanwei Yue, Boci Peng, Xuanbo Fan, Jiaxin Guo, Qiankun Li, and Yan Zhang. 2026.
\newblock Mem-t: Densifying rewards for long-horizon memory agents.
\newblock \emph{arXiv preprint arXiv:2601.23014}.

\bibitem[{Zhang et~al.(2025{\natexlab{a}})Zhang, Fu, Wan, Yu, Wang, and Yan}]{zhang2025gmemorytracinghierarchicalmemory}
Guibin Zhang, Muxin Fu, Guancheng Wan, Miao Yu, Kun Wang, and Shuicheng Yan. 2025{\natexlab{a}}.
\newblock \href {https://arxiv.org/abs/2506.07398} {G-memory: Tracing hierarchical memory for multi-agent systems}.
\newblock \emph{Preprint}, arXiv:2506.07398.

\bibitem[{Zhang et~al.(2026{\natexlab{a}})Zhang, Long, Bao, Feng, Zhang, Yue, and Wang}]{zhang2026memskilllearningevolvingmemory}
Haozhen Zhang, Quanyu Long, Jianzhu Bao, Tao Feng, Weizhi Zhang, Haodong Yue, and Wenya Wang. 2026{\natexlab{a}}.
\newblock \href {https://arxiv.org/abs/2602.02474} {Memskill: Learning and evolving memory skills for self-evolving agents}.
\newblock \emph{Preprint}, arXiv:2602.02474.

\bibitem[{Zhang et~al.(2026{\natexlab{b}})Zhang, Yue, Feng, Long, Bao, Jin, Zhang, Li, You, Qin, and Wang}]{zhang2026learningqueryawarebudgettierrouting}
Haozhen Zhang, Haodong Yue, Tao Feng, Quanyu Long, Jianzhu Bao, Bowen Jin, Weizhi Zhang, Xiao Li, Jiaxuan You, Chengwei Qin, and Wenya Wang. 2026{\natexlab{b}}.
\newblock \href {https://arxiv.org/abs/2602.06025} {Learning query-aware budget-tier routing for runtime agent memory}.
\newblock \emph{Preprint}, arXiv:2602.06025.

\bibitem[{Zhang et~al.(2026{\natexlab{c}})Zhang, Hu, Lu, Lange, and Clune}]{zhang2026darwin}
Jenny Zhang, Shengran Hu, Cong Lu, Robert Lange, and Jeff Clune. 2026{\natexlab{c}}.
\newblock Darwin g{\"o}del machine: Open-ended evolution of self-improving agents.
\newblock In \emph{International Conference on Learning Representations}, volume 2026, pages 104223--104294.

\bibitem[{Zhang et~al.(2026{\natexlab{d}})Zhang, Zhao, Yang, Foerster, Clune, Jiang, Devlin, and Shavrina}]{zhang2026hyperagents}
Jenny Zhang, Bingchen Zhao, Wannan Yang, Jakob Foerster, Jeff Clune, Minqi Jiang, Sam Devlin, and Tatiana Shavrina. 2026{\natexlab{d}}.
\newblock \href {https://arxiv.org/abs/2603.19461} {Hyperagents}.
\newblock \emph{Preprint}, arXiv:2603.19461.

\bibitem[{Zhang et~al.(2026{\natexlab{e}})Zhang, Cai, Li, Wetzstein, and Agrawala}]{zhang2026frame}
Lvmin Zhang, Shengqu Cai, Muyang Li, Gordon Wetzstein, and Maneesh Agrawala. 2026{\natexlab{e}}.
\newblock Frame context packing and drift prevention in next-frame-prediction video diffusion models.
\newblock \emph{Advances in Neural Information Processing Systems}, 38:30546--30566.

\bibitem[{Zhang et~al.(2025{\natexlab{b}})Zhang, Li, Long, Zhang, Lin, Yang, Xie, Yang, Liu, Lin, Huang, and Zhou}]{zhang2025qwen3embeddingadvancingtext}
Yanzhao Zhang, Mingxin Li, Dingkun Long, Xin Zhang, Huan Lin, Baosong Yang, Pengjun Xie, An~Yang, Dayiheng Liu, Junyang Lin, Fei Huang, and Jingren Zhou. 2025{\natexlab{b}}.
\newblock \href {https://arxiv.org/abs/2506.05176} {Qwen3 embedding: Advancing text embedding and reranking through foundation models}.
\newblock \emph{Preprint}, arXiv:2506.05176.

\bibitem[{Zhao et~al.(2026)Zhao, Yuan, Huang, Yuan, Yu, Xu, Hu, Shankarampeta, Huang, Ni, Tian, and Zhao}]{zhao2026amabenchevaluatinglonghorizonmemory}
Yujie Zhao, Boqin Yuan, Junbo Huang, Haocheng Yuan, Zhongming Yu, Haozhou Xu, Lanxiang Hu, Abhilash Shankarampeta, Zimeng Huang, Wentao Ni, Yuandong Tian, and Jishen Zhao. 2026.
\newblock \href {https://arxiv.org/abs/2602.22769} {Ama-bench: Evaluating long-horizon memory for agentic applications}.
\newblock \emph{Preprint}, arXiv:2602.22769.

\bibitem[{Zhou et~al.(2026{\natexlab{a}})Zhou, Zhou, Han, Xu, Li, Li, Xiong, and Wu}]{zhou2026readyagentnativememorysystem}
Wei Zhou, Xuanhe Zhou, Shaokun Han, Hongming Xu, Guoliang Li, Zhiyu Li, Feiyu Xiong, and Fan Wu. 2026{\natexlab{a}}.
\newblock \href {https://arxiv.org/abs/2606.24775} {Are we ready for an agent-native memory system?}
\newblock \emph{Preprint}, arXiv:2606.24775.

\bibitem[{Zhou et~al.(2026{\natexlab{b}})Zhou, Li, Liu, Yavuz, and Joty}]{zhou2026justintimememorylearningcurate}
Yefan Zhou, Yang Li, Zeyu~Leo Liu, Semih Yavuz, and Shafiq Joty. 2026{\natexlab{b}}.
\newblock \href {https://arxiv.org/abs/2609.27334} {Just-in-time memory: Learning to curate task-adaptive memory for llm agents}.
\newblock \emph{Preprint}, arXiv:2609.27334.

\bibitem[{Zhuo et~al.(2025)Zhuo, Vu, Chim, Hu, Yu, Widyasari, Yusuf, Zhan, He, Paul, Brunner, Gong, Hoang, Zebaze, Hong, Li, Kaddour, Xu, Zhang, Yadav, Jain, Gu, Cheng, Liu, Liu, Wang, Hui, Muennighoff, Lo, Fried, Du, de~Vries, and Werra}]{zhuo2025bigcodebenchbenchmarkingcodegeneration}
Terry~Yue Zhuo, Minh~Chien Vu, Jenny Chim, Han Hu, Wenhao Yu, Ratnadira Widyasari, Imam Nur~Bani Yusuf, Haolan Zhan, Junda He, Indraneil Paul, Simon Brunner, Chen Gong, Thong Hoang, Armel~Randy Zebaze, Xiaoheng Hong, Wen-Ding Li, Jean Kaddour, Ming Xu, Zhihan Zhang, and 14 others. 2025.
\newblock \href {https://arxiv.org/abs/2406.15877} {Bigcodebench: Benchmarking code generation with diverse function calls and complex instructions}.
\newblock \emph{Preprint}, arXiv:2406.15877.

\end{thebibliography}

\clearpage
\appendix
\raggedbottom   % appendix: short columns keep their natural spacing instead of stretching the glue around headings

% Appendix-only table of contents (entries from here on are captured;
% shows up after the second compile)
\startcontents[appendix]
\section*{Appendix Contents}
\printcontents[appendix]{}{1}{\setcounter{tocdepth}{2}}

\clearpage
\section{Experiment Details}
\label{apdx:exp_details}

\subsection{Benchmark Details}
\label{apdx:benchmark_details}

Table~\ref{tab:bench_stats} summarizes the evaluation data.

\paragraph{LongMemEval-s~\citep{wu2025longmemevalbenchmarkingchatassistants}.}
Each question has its own chat history of about 50 sessions, with a few evidence sessions hidden among distractors. We use the full set, except a 100-corpus subset stratified by question type for \texttt{GPT-5.6 Luna}.

\paragraph{PersonaMem-v2~\citep{jiang2025personamemv2personalizedintelligencelearning}.}
Each corpus is one simulated user's chat history (128k-token variant), with four-way MCQs on the user's implicit persona and preferences. We evaluate a 20\% subset (40 of 200 corpora).

\paragraph{MINT-Eval~\citep{lee2026mintevalevaluatingmemorymultitarget}.}
Each corpus is an ordered stream of steps from one of four sub-domains, testing memory under multi-target interference. We draw a 10\% stratified sample (60 of 595 corpora).

\subsection{Baseline Details}
\label{apdx:baseline_details}

\paragraph{Vanilla.} These references use no retrieval or memory building. \textbf{No Context} answers from the question alone. 
\textbf{Full Context} provides the entire raw corpus in chronological order, truncating the earliest units if it exceeds the model's context window. 
\textbf{Gold Context} provides only the gold evidence (raw turns supporting the answer) for each question, serving as an oracle reference. We omit this baseline on MINT-Eval, where author-annotated and verified gold evidence is available for only about half of the questions (Appendix~\ref{apdx:faithfulness}).

\paragraph{Naive RAG.}
The raw corpus is first chunked and indexed. Chunking follows conversation turns (or observation steps) and never crosses session boundaries. Short turns are greedily merged into chunks of about 1024 characters, with the last turn carried over as overlap. Longer turns remain separate, while turns exceeding 8192 characters are split into 8192-character windows with 1024-character overlap. Each chunk is prefixed with a session date-time header.
At question time, BM25 and dense retrieval (\texttt{Qwen3-Embedding-0.6B}) independently return their top-$k$ chunks. Their deduplicated union is reranked by \texttt{Qwen3-Reranker-0.6B}, and the top $k$ chunks, roughly half of the candidate pool, are passed to the backbone with the shared QA prompt.

For \textsc{RAG (2-round)}, we add a second retrieval round. The backbone checks whether the first-round chunks suffice and, if not, issues a follow-up query against the remaining pool. Chunks from both rounds are used for answering.

\paragraph{Memory systems.} We group the memory systems as follows:

\catsq~\textbf{Flat store} keeps memory as a flat collection (extracted facts, notes, etc.), indexed for similarity search: \textsc{A-Mem}~\citep{xu2026mem}, \textsc{SimpleMem}~\citep{liu2026simplememefficientlifelongmemory}, \textsc{MemOS}~\citep{li2025memosoperatingmemoryaugmentedgeneration}.

\catsq~\textbf{Hier/Lifecycle} manages memory across tiers that differ in lifetime or abstraction, and moves, consolidates or updates units between them over time: \textsc{MemoryOS}~\citep{kang2025memory}, \textsc{LightMem}~\citep{fang2026lightmem}, and \textsc{OpenViking}~\citep{openviking2026}.

\catsq~\textbf{Graph-based} links memory units through explicit relations (e.g., entities, events or causal edges), and retrieval can expand along them: \textsc{HippoRAG2}~\citep{gutierrez2025rag}, \textsc{PlugMem}~\citep{yang2026plugmem}, \textsc{MAGMA}~\citep{jiang2026magma}.

\catsq~\textbf{Trained Policy} learns how memory is written, updated and used through reinforcement learning rather than a hand-designed pipeline, whatever the store layout (hierarchical for \textsc{Mem-T}, flat for \textsc{Memory-R2}): \textsc{Mem-T}~\citep{yue2026mem}, \textsc{Memory-R2}~\citep{yan2026memory}.
For these two systems, memory construction, update and retrieval are carried out by their own policy models (\texttt{Qwen3-4B} for \textsc{Mem-T}, \texttt{Qwen2.5-7B-Instruct} for \textsc{Memory-R2}), which we keep as released, replacing only the QA model with our backbones.

\begin{table}[t]
\centering
\caption{Statistics of the evaluation data. Tokens and questions per corpus are medians. $^\ast$Due to cost, we evaluate on a subset of the released benchmark (see text for the sampling procedure).}
\label{tab:bench_stats}
\vspace{-0.6em}
\small
\setlength{\tabcolsep}{6pt}
\resizebox{\linewidth}{!}{%
\begin{tabular}{l c c c c l}
\toprule
\textbf{Benchmark} & \textbf{\#Corpus} & \textbf{\#Q} & \textbf{Q/corpus} & \textbf{Tokens/corpus} & \textbf{Answer format} \\
\midrule
LongMemEval-s & 500 & 500 & 1 & $\sim$115k & short answer \\
\arrayrulecolor{black!35}\midrule\arrayrulecolor{black}
PersonaMem-v2$^\ast$ & 40 & 934 & 25 & $\sim$123k & 4-way MCQ \\
\arrayrulecolor{black!35}\midrule\arrayrulecolor{black}
MINT-Eval$^\ast$ & 60 & 1{,}557 & 26 & $\sim$139k & \multirow{5}{*}{short answer} \\
\quad Multi-turn dialogue & 10 & 690 & 70 & $\sim$370k &  \\
\quad State tracking & 10 & 563 & 50 & $\sim$0.2k &  \\
\quad GitHub commits & 20 & 154 & 7.5 & $\sim$56k &  \\
\quad Wiki revisions & 20 & 150 & 7 & $\sim$129k &  \\
\bottomrule
\end{tabular}%
}
\vspace{-1.5em}
\end{table}

\subsection{Evaluation Metrics}
\label{apdx:eval_metrics}

\paragraph{Performance.}
We score answers with an LLM-as-Judge backed by \texttt{Qwen3.6-27B} and \textbf{deliberately avoid lexical-overlap metrics such as F1 or BLEU}, which are sensitive to surface form and can penalize a correct but rephrased answer.
For general QA, the judge checks the generated answer against the gold answer. For MCQ, we match the parsed option exactly and send unparsable responses to the judge.
Judge prompts are in Appendix~\ref{apdx:judge_prompts}, and Table~\ref{tab:judge_reliability} reports the judge reliability against human labels.

\begin{table}[t]
\centering
\caption{Reliability of the LLM-as-Judge against human judgments, measured as the fraction of items on which the judge and human labels match.}
\label{tab:judge_reliability}
\vspace{-0.6em}
\small
\setlength{\tabcolsep}{4pt}
\begin{adjustbox}{max width=\linewidth}
\begin{tabular}{l c c c}
\toprule
 & \textbf{LongMemEval-s} & \textbf{PersonaMem-v2} & \textbf{MINT-Eval} \\
\midrule
\textbf{Agree. (\%)} & 95.1\% & 100.0\%$^\dagger$  & 97.3\% \\
\bottomrule
\end{tabular}
\end{adjustbox}
\par\vspace{2pt}{\footnotesize\raggedright $^\dagger$PersonaMem-v2 is MCQ, so answers are scored by the option letter alone, matching the human labels on every item.\par}
\vspace{-1em}
\end{table}

\paragraph{Cost.}
We report the following cost metrics for every system--benchmark--backbone triple.

\catsq~\textbf{Token cost.} All input and output LLM tokens, summed over ingestion (memory construction, maintenance, etc.) and QA (retrieval, intermediate reasoning, answering, etc.).

\catsq~\textbf{End-to-end wall-clock time (Equivalent single-concurrency time).} Systems run with different concurrency, so we normalize each runtime, covering construction, retrieval and answering, to its single-concurrency equivalent in hours. Times under the same backbone are broadly comparable, though endpoint throughput can still fluctuate.

\catsq~\textbf{LLM calls.} The total number of LLM calls across ingestion and QA, which, unlike wall-clock time, does not depend on endpoint load.

\catsq~\textbf{Memory store size.} The total tokens across all stored memory units after ingestion, averaged over corpora, reflecting how much a system expands or compresses the raw corpus.

\subsection{Implementation}
\label{apdx:implementation}

\paragraph{Retrieval scope.}
In existing comparisons of memory systems, the retrieval top-$k$ used at QA time is usually inherited from each repository's default configuration or aligned with that of the proposed system. 
Since memory units vary widely across systems, from a single turn to a merged chunk or session summary, the same top-$k$ can span from a few hundred to tens of thousands of tokens.
So we measure \textbf{Retrieval Scope} (RS), defined as the fraction of tokens in memory units seen by the LLM from retrieval through QA over the total tokens in all memory units within the system's retrieval range. We average this ratio over questions and report it as a percentage.
Tokens are counted with the \texttt{Qwen3.6-27B} tokenizer. 
We adjust each system's top-$k$ so that its RS is \textbf{\textit{no lower than}} that of ``RAG (1-round)'', while keeping RS reasonably comparable across systems.

We further clarify what counts as ``the docs LLM sees'' in RS. This depends on the LLM's involvement in retrieval, which falls into three levels: \boxednum{1} no involvement, where retrieval relies entirely on embeddings, indices, or other non-LLM components. \boxednum{2} pre-retrieval involvement, where LLM rewrites query or extracts keywords but reads no retrieved units. \boxednum{3} retrieval-time involvement, where the LLM reads retrieved units, e.g., to decide whether further retrieval is needed, reformulate the query, or assess whether a unit is useful.

Only level \boxednum{3} exposes the LLM to memory units beyond the final answer prompt. RS counts all such exposure, including every unit read from retrieval through QA regardless of whether it appears in the final answer context.
Both the numerator and denominator count only content that participates in retrieval. Metadata attached to memory units, such as topic, involved people, or date, is excluded when it is not searched over.

The RS requirement has some exceptions. Some memory systems produce only a few memory units, each covering a large part of the corpus. Forcing their RS to match would leave a very small top-$k$, as for OpenViking, so we fix a top-$k$ for them.

\paragraph{Inference settings.} We use NVIDIA H20 GPUs to locally served
\texttt{Qwen3.6-27B} and \texttt{Qwen3.5-4B}, and they are decoded with temperature $0.0$ and top-$p$ $1.0$ for every ingestion and QA call, with reasoning disabled. \texttt{GPT-5.6 Luna (low)} is accessed through the OpenAI API with the default sampling config and reasoning effort set to \texttt{low}.

For a fair comparison, we unify the embedding model and, where a system uses one, the reranker model across all baselines.
Unless otherwise stated, all baselines use \texttt{Qwen3-Embedding-0.6B}~\citep{zhang2025qwen3embeddingadvancingtext} as the embedding model and \texttt{Qwen3-Reranker-0.6B}~\citep{zhang2025qwen3embeddingadvancingtext} as the reranker if used in certain memory system.

\subsection{More Explanations}
\label{apdx:more_explanations}

\textbf{GPT-5.6 luna on PersonaMem-v2.}
Interestingly, on PersonaMem-v2 5.6-luna usually scores lower than Q3.6-27B for most systems, while on LongMemEval-s and MINT-Eval, it's usually better. Our inspection suggests this gap reflects answer preference of 5.6-luna, not the ability degradation. PersonaMem-v2 often contrasts a personalized answer with an almost identical generic distractor, and 5.6-luna tends to prefer the generic option. This preference appears to interact with 5.6-luna's tendency to avoid relying on stereotypes, health-related, or privacy-sensitive information. This pattern is specific to PersonaMem-v2, whose multiple-choice questions often reward explicit use of user preferences, unlike the other two benchmarks.

\textbf{Gold Context baseline on MINT-Eval.}
MINT-Eval provides author-annotated gold context for only about half of its queris, so we skip it. 
\clearpage
\section{Configuration Validation}
\label{apdx:baseline_validation}

The main results in Section~\ref{sec:experiments} show that naive RAG performs surprisingly well, outperforming many memory systems across all our three benchmarks. To further examine and validate it, we provide three complementary analyses: \boxednum{1} component-wise ablations of our RAG baseline to show its stability (\ref{apdx:rag_ablation}); \boxednum{2} a review of RAG implementations in prior memory works (\ref{apdx:rag_prior_works}); and \boxednum{3} a validation of memory baseline configurations, examining our deviations from the original repositories and their impact on performance (\ref{apdx:config_ablation}).

\subsection{Ablations on RAG Baseline}
\label{apdx:rag_ablation}

Since naive RAG serves as the reference point for our memory tax analysis throughout the paper, we examine how sensitive it is to its design choices. The default configuration is the single-round hybrid RAG (1rd), described in Section~\ref{sec:exp_setup} (details in Appendix~\ref{apdx:baseline_details}), which uses buffer-merged chunks, BM25 and dense retrieval combined and then reranked, with \texttt{Qwen3-Embedding-0.6B} as the embedding model and \texttt{Qwen3-Reranker-0.6B} as the reranker. 

\begin{table}[H]
\centering
\caption{Ablations of the RAG baseline on chunking method, retrieval method, embedding model and reranker. Scores are judge accuracy (\%).}
\label{tab:rag_ablation}
% 10-13 user: default rows synced to Table 1 RAG (1rd) (LME 70.2 / 77.6 after the preference re-judge, MINT Q27B 28.6); bold re-ranked per block
\vspace{-0.6em}
\small
\setlength{\tabcolsep}{5pt}
\resizebox{\linewidth}{!}{%
\begin{tabular}{l| c c| c c| c c}
\toprule
\multicolumn{1}{l}{\multirow{2}{*}{\textbf{Method}}} & \multicolumn{2}{c}{\textbf{LongMemEval-s}} & \multicolumn{2}{c}{\textbf{PersonaMem-v2}} & \multicolumn{2}{c}{\textbf{MINT-Eval}} \\
\cmidrule(lr){2-3} \cmidrule(lr){4-5} \cmidrule(lr){6-7}
\multicolumn{1}{l}{} & \texttt{\textbf{Q4B}} & \texttt{\textbf{Q27B}} & \texttt{\textbf{Q4B}} & \texttt{\textbf{Q27B}} & \texttt{\textbf{Q4B}} & \texttt{\textbf{Q27B}} \\
\midrule
\multicolumn{7}{c}{\textit{\textbf{Chunking Method}}} \\
\midrule
\textbf{\textit{Original (Buffer-1024)}} & 70.2 & \textbf{77.6} & \textbf{37.2} & 46.8 & \textbf{25.8} & \textbf{28.6} \\
Buffer-512 & \textbf{71.2} & 76.8 & 35.8 & 46.5 & 25.6 & 25.7 \\
Turn-level & 70.8 & 76.6 & 35.2 & \textbf{47.2} & \textbf{25.8} & 26.1 \\
Streaming-chunk & 68.4 & 69.4 & 35.1 & 46.2 & 25.1 & 24.8 \\
\midrule
\multicolumn{7}{c}{\textit{\textbf{Retrieval Method}}} \\
\midrule
\textbf{\textit{Original (Hybrid)}} & \textbf{70.2} & \textbf{77.6} & \textbf{37.2} & 46.8 & 25.8 & \textbf{28.6} \\
BM25 & 63.4 & 68.2 & 34.4 & 42.7 & 24.4 & 23.3 \\
Emb-Only & 67.2 & 72.6 & 35.4 & \textbf{47.3} & \textbf{26.5} & 24.7 \\
\midrule
\multicolumn{7}{c}{\textit{\textbf{Embedding Model}}} \\
\midrule
\textbf{\textit{Qwen3-Embedding-0.6B}} & 70.2 & 77.6 & 37.2 & 46.8 & 25.8 & \textbf{28.6} \\
all-MiniLM-L6-v2 (0.02B) & 68.0 & 74.6 & 35.7 & 45.7 & 25.3 & 25.3 \\
BGE-Large-en-v1.5 (0.3B) & \textbf{72.2} & 76.6 & \textbf{38.0} & 47.4 & 25.9 & 25.2 \\
% Nemotron3-Embed-1B & 70.4 & 78.4 & 35.7 & 46.4 & 25.4 & 25.7 \\
pplx-embed-v1 (4B) & 68.4 & 75.8 & 37.6 & 47.5 & \textbf{26.3} & 25.6 \\
% Qwen3-Embedding-4B & 69.4 & 77.8 & 34.9 & 46.5 & 26.7 & 26.1 \\
NV-Embed-v2 (8B) & 69.6 & \textbf{78.2} & 37.4 & \textbf{49.3} & 25.3 & 25.8 \\
% Qwen3-Embedding-8B & 69.8 & 77.2 & 34.9 & 47.1 & 25.6 & 25.8 \\
\midrule
\multicolumn{7}{c}{\textit{\textbf{Reranker}}} \\
\midrule
\textbf{\textit{Qwen3-Reranker-0.6B}} & 70.2 & \textbf{77.6} & 37.2 & 46.8 & 25.8 & \textbf{28.6} \\
BGE-Reranker-Large (0.6B) & 69.8 & 76.4 & 34.5 & 45.6 & 25.7 & 25.1 \\
Jina-Reranker-v3 (0.6B) & \textbf{71.4} & \textbf{77.6} & \textbf{37.3} & \textbf{47.4} & \textbf{26.7} & 26.3 \\
Qwen3-Reranker-4B & 70.0 & 77.2 & 35.3 & 46.3 & 26.2 & 26.2 \\
\bottomrule
\end{tabular}%
}
\vspace{-1em}
\end{table}

We vary one component at a time while keeping the rest at the default, along these dimensions:

\catsq~\textbf{Chunking method}:

\boxednum{1} buffer-merged chunks with a 1024-chars buffer \textbf{[default]},

\boxednum{2} a smaller 512-chars buffer (with long turns split into 4096-chars windows); 

\boxednum{3} turn-level chunks, where each conversation turn or observation step forms its own chunk; 

\boxednum{4} streaming chunks, where the history is cut into fixed 1024-chars windows with a 256-chars overlap, regardless of turn boundaries.

\catsq~\textbf{Retrieval method}: 

\boxednum{1} hybrid retrieval \textbf{[default]};

\boxednum{2} \textbf{BM25 only};

\boxednum{3} dense embedding retrieval only (\textbf{Emb-Only}), both without reranking.

\catsq~\textbf{Embedding model} (ordered by model size): 

\texttt{Qwen3-Embedding-0.6B} \textbf{[default]}

\texttt{all-MiniLM-L6-v2} (0.02B)~\citep{wang2020minilmdeepselfattentiondistillation} 

\texttt{BGE-Large-en-v1.5} (0.3B)~\citep{xiao2024cpackpackedresourcesgeneral}

% \texttt{Nemotron3-Embed} (1B)~\citep{nvidia2026nemotron3embed}

\texttt{pplx-embed-v1} (4B)~\citep{eslami2026diffusionpretraineddensecontextualembeddings}

% \texttt{Qwen3-Embedding-4B}~\citep{zhang2025qwen3embeddingadvancingtext}

\texttt{NV-Embed-v2} (8B)~\citep{lee2025nvembedimprovedtechniquestraining}.

% \texttt{Qwen3-Embedding-8B}~\citep{zhang2025qwen3embeddingadvancingtext}.

\catsq~\textbf{Reranker} (ordered by model size): 

\texttt{Qwen3-Reranker-0.6B} \textbf{[default]}

\texttt{BGE-Reranker-Large} (0.6B)~\citep{liu2025llama2vecunsupervisedadaptationlarge}

\texttt{Jina-Reranker-v3} (0.6B)~\citep{wang2025jinarerankerv3lateinteractiondocument}

\texttt{Qwen3-Reranker-4B}~\citep{zhang2025qwen3embeddingadvancingtext}.

All other settings, including the retrieval top-$k$ and the answer prompt, follow the default configuration.
The full results are shown in Table~\ref{tab:rag_ablation}.

Across these ablations, naive RAG remains strong under a range of configurations, suggesting that its performance is \textbf{not driven by any particular component, careful tuning, or a lucky configuration}. Most variants stay within a few points of the default across settings. The largest drops occur when fixed-size windows split turns (streaming-chunk) or retrieval relies solely on lexical matching (BM25), yet even these variants remain competitive with several memory systems in Table~\ref{tab:main_results}.

\subsection{RAG Implementation in Prior Works}
\label{apdx:rag_prior_works}

All above results also raise a natural question: \textbf{given its strong performance and simplicity, has the naive RAG been largely overlooked in prior memory methods and studies? }

Revisiting the papers from all of our evaluated baselines along with several other influential memory papers (\citet{chhikara2025mem0buildingproductionreadyai}, \citet{rasmussen2025zeptemporalknowledgegraph}, \citet{packer2024memgptllmsoperatingsystems}, 
\citet{ma2026deservesmemoryadaptivememory}, \citet{zhang2026memskilllearningevolvingmemory}), we find that \textbf{only three of them} (excluding originally RAG system like HippoRAG2) report RAG results (\textsc{LightMem}, \textsc{PlugMem}, and \textsc{Mem-T}), while the others primarily compare against other existing memory systems.
Specifically, \textsc{LightMem} segments corpus by turn and retrieves using all-MiniLM-L6-v2 embedding \textbf{without specifying retrieval budget and other settings}. \textsc{PlugMem} also uses turn-level chunks and retrieves using NV-Embed-v2 embedding (top-10). \textsc{Mem-T} adopts BGE-M3 with a top-5 retrieval, \textbf{without specifying its chunking strategy}. 

These observations suggest that the potential of naive RAG may have been underestimated and not fully realized in prior memory studies. With a widely-used standard hybrid retrieval and reranking pipeline, RAG becomes a strong baseline, highlighting the importance of adequately evaluating retrieval controls when assessing the added value of constructed memory.

\subsection{Memory Baseline Config Offset Validation}
\label{apdx:config_ablation}

We largely follow the official configurations of each memory baseline \textbf{as closely as possible}. When modifications are necessary for controlled comparison, we further examine their impact to ensure that they DO NOT artificially disadvantage the corresponding baseline.

\begin{table}[t]
\centering
\small
\setlength{\tabcolsep}{3pt}
\caption{The effect of retrieval top-$k$ on LME-s.}
\label{tab:config_ablation_topk}
\vspace{-0.6em}
\begin{adjustbox}{max width=\linewidth}
\begin{tabular}{@{}l ccc@{}}
\toprule
 & \textsc{A-Mem}\,{\footnotesize(\texttt{luna})} & \textsc{SimpleMem}\,{\footnotesize(\texttt{Q27B})} & \textsc{PlugMem}\,{\footnotesize(\texttt{Q27B})} \\
\midrule
Ours    & {\normalsize\textbf{84.0}} {\scriptsize($k$=15)} & 73.6 {\scriptsize($k$=25)} & {\normalsize\textbf{75.8}} {\scriptsize($k$=15)} \\
Default & 73.0 {\scriptsize($k$=5)} & = Ours & 63.6 {\scriptsize($k$=30)} \\
\midrule
 & \textsc{HippoRAG2}\,{\footnotesize(\texttt{Q27B})} & \textsc{Mem-T}\,{\footnotesize(\texttt{Q27B})} & \textsc{Memory-R2}\,{\footnotesize(\texttt{luna})} \\
\midrule
Ours    & {\normalsize\textbf{78.2}} {\scriptsize($k$=10)} & {\normalsize\textbf{75.0}} {\scriptsize($k$=15)} & 79.0 {\scriptsize($k$=50)} \\
Default & 68.2 {\scriptsize($k$=5)} & 67.2 {\scriptsize($k$=5)} & {\normalsize\textbf{79.6}} {\scriptsize($k$=30)} \\   % 10-13 user: Ours = Table 1 (79.0, was 80.0)
\bottomrule
\end{tabular}
\end{adjustbox}
\vspace{-0.5em}
\vspace{0.9em}
\caption{Retrieval top-$k$: ours vs.\ the default setting on PersonaMem-v2.}
\label{tab:config_ablation_topk_psn}
\vspace{-0.6em}
\begin{adjustbox}{max width=\linewidth}
\begin{tabular}{@{}l ccc@{}}
\toprule
 & \textsc{A-Mem}\,{\footnotesize(\texttt{Q4B})} & \textsc{SimpleMem}\,{\footnotesize(\texttt{Q4B})} & \textsc{PlugMem}\,{\footnotesize(\texttt{Q27B})} \\
\midrule
Ours    & {\normalsize\textbf{33.7}} {\scriptsize($k$=20)} & 34.7 {\scriptsize($k$=10)} & {\normalsize\textbf{45.2}} {\scriptsize($k$=15)} \\
Default & 33.1 {\scriptsize($k$=5)} & {\normalsize\textbf{37.5}} {\scriptsize($k$=25)} & 43.5 {\scriptsize($k$=30)} \\
\midrule
 & \textsc{HippoRAG2}\,{\footnotesize(\texttt{Q27B})} & \textsc{Mem-T}\,{\footnotesize(\texttt{Q4B})} & \textsc{Memory-R2}\,{\footnotesize(\texttt{Q4B})} \\
\midrule
Ours    & 46.5 {\scriptsize($k$=10)} & {\normalsize\textbf{32.1}} {\scriptsize($k$=10)} & {\normalsize\textbf{38.2}} {\scriptsize($k$=50)} \\
Default & {\normalsize\textbf{47.3}} {\scriptsize($k$=5)} & 31.9 {\scriptsize($k$=5)} & 35.8 {\scriptsize($k$=30)} \\
\bottomrule
\end{tabular}
\end{adjustbox}
\vspace{-1em}
\end{table}

\paragraph{Shared settings.}
For fair comparison, we standardize several evaluation components across baselines, including the QA prompt, QA backbone, judge prompt, judge backbone, and embedding model. These settings are shared by design to reduce confounding factors unrelated to the memory mechanism itself, so we do not ablate them separately for individual baselines.

\paragraph{Retrieval scope.}
We additionally adjust the retrieval top-$k$ to control the \textbf{relative retrieval scope} of each system. 
RS is not intended to equalize exact inference token budgets across systems, but to avoid overly restrictive retrieval caused by heterogeneous memory-unit granularity and store sizes.
Using the same top-$k$ across systems can be misleading because memory units differ substantially in granularity and store size. For example, one system may effectively access 10\% of its store while another accesses only 1\%. So we tune top-$k$ based on Retrieval Scope (Appendix~\ref{apdx:implementation}), ensuring that each memory baseline has a scope no lower than RAG (1rd). In practice, this adjustment \textbf{mostly increases the original top-$k$}, giving memory baselines access to more of their stored information.

Table~\ref{tab:config_ablation_topk} shows the effect of adjusting retrieval top-$k$ to align retrieval scope across systems. In most cases, we \textbf{increase top-$k$} beyond the repository defaults, which generally improves performance. In the few cases where top-$k$ must be reduced to control its retrieval scope, the resulting performance degradation is minor.

\paragraph{System-specific deviations.}
Beyond the shared settings and the retrieval scope, a few systems need individual changes:
\begin{table}[t]
\centering
\small
\setlength{\tabcolsep}{5pt}
\caption{\textsc{MemOS}: retrieval post-processing and reranking.}
\label{tab:dev_memos}
\vspace{-0.6em}
\begin{tabular}{l l cc}
\toprule
\textbf{Backbone} & \textbf{Setting} & \textbf{LME-s} & \textbf{PM-v2} \\
\midrule
\multirow{2}{*}{\texttt{luna}} & Ours (no post-proc.) & {\normalsize\textbf{78.0}} & {\normalsize\textbf{41.7}} \\
 & + official post-proc. & 78.0 & 37.8 \\
\midrule
\multirow{2}{*}{\texttt{Q27B}} & Ours (reranking on) & {\normalsize\textbf{76.6}} & {\normalsize\textbf{50.2}} \\
 & default (reranking off) & 45.6 & 42.7 \\
\bottomrule
\end{tabular}

\vspace{0.9em}
\caption{\textsc{LightMem}: extraction source on LongMemEval-s (\texttt{Q27B}).}
\label{tab:dev_lightmem}
\vspace{-0.6em}
\begin{tabular}{l c}
\toprule
\textbf{Setting} & \textbf{LME-s} \\
\midrule
User + assistant (ours) & {\normalsize\textbf{78.8}} \\
User only (default) & 70.6 \\
\bottomrule
\end{tabular}

\vspace{-1em}
\end{table}

(1) \textsc{MemOS}: we disable the retrieval post-processing of its official evaluation (relativity threshold of 0.45 and MMR), whose threshold is calibrated for its original embedding model and overly strict with ours, and enable the cosine reranking that degenerates to constant scores by default. Both changes favor \textsc{MemOS} (Table~\ref{tab:dev_memos}).

(2) \textsc{SimpleMem}: its default ingestion window of 40 utterances is set for LoCoMo and packs unrelated sessions together elsewhere, so we use 10 on all three benchmarks.

(3) \textsc{LightMem}: its official LongMemEval-s script uses user turns only. We compare this setting with user+assistant turns on LongMemEval-s and find the latter performs better (Table~\ref{tab:dev_lightmem}), so we use it across all three benchmarks.

(4) \textsc{Mem-T}: we stop the policy model after retrieval, before it answers at \texttt{finish}, and let the shared QA backbone answer.

(5) \textsc{Memory-R2}: we raise the output limits of both construction stages (2048 and 1024 tokens) to 4096, its training length, since the defaults truncate about 12\% of the calls on LongMemEval-s and drop whole blocks of memory operations.

\clearpage
\section{Analysis Details}
\label{apdx:analysis_details}

\subsection{Overview}
\label{apdx:tax_overview}

\begin{figure*}[p]
\centering
\includegraphics[width=\textwidth]{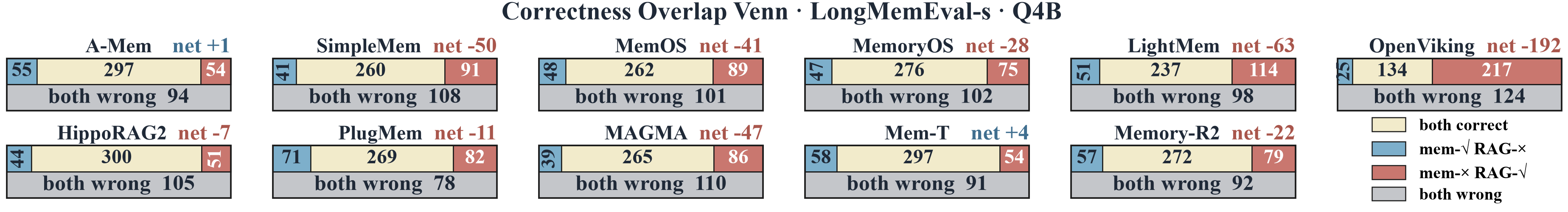}\\[0.5em]
\includegraphics[width=\textwidth]{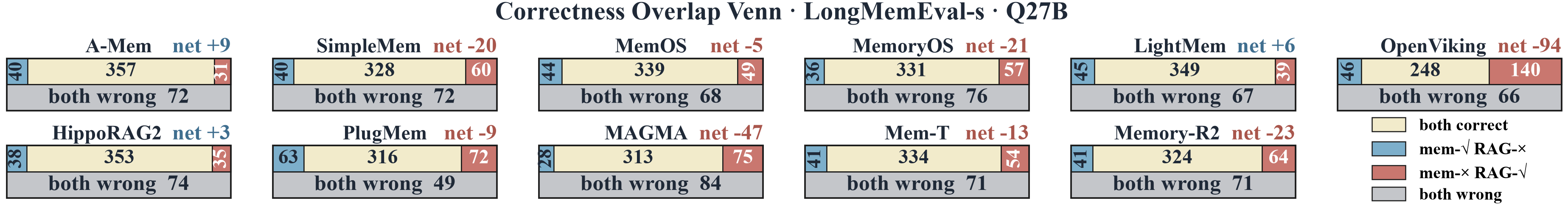}
\par\vspace{-0.6em}
\caption{Per-query correctness overlap between each memory system and RAG (1rd) on LongMemEval-s, with \texttt{Qwen3.5-4B} (top) and \texttt{Qwen3.6-27B} (bottom).}
\label{fig:venn_all11_longmemeval_s}
\par\vspace{1.2em}
\includegraphics[width=\textwidth]{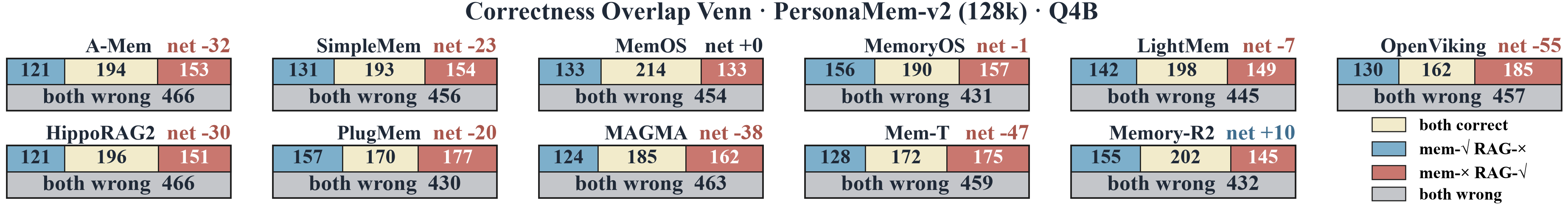}\\[0.5em]
\includegraphics[width=\textwidth]{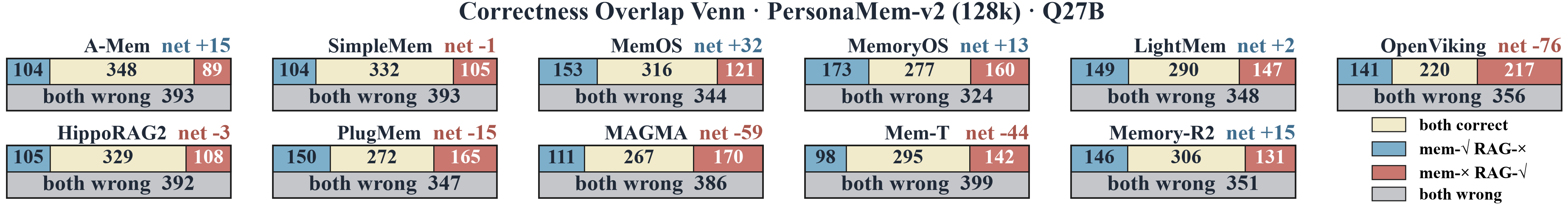}
\par\vspace{-0.6em}
\caption{Per-query correctness overlap between each memory system and RAG (1rd) on PersonaMem-v2, with \texttt{Qwen3.5-4B} (top) and \texttt{Qwen3.6-27B} (bottom).}
\label{fig:venn_all11_persona_mem_v2_128k}
\par\vspace{1.2em}
\includegraphics[width=\textwidth]{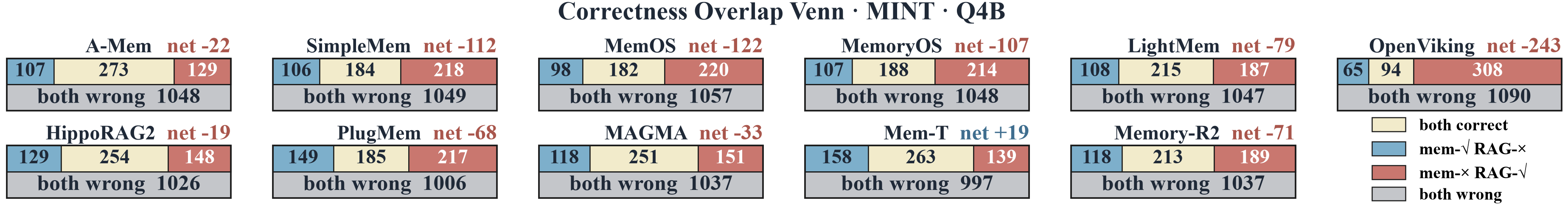}\\[0.5em]
\includegraphics[width=\textwidth]{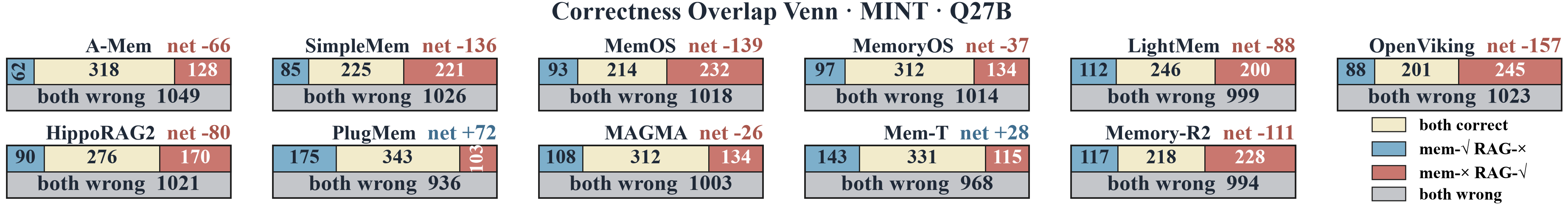}
\par\vspace{-0.6em}
\caption{Per-query correctness overlap between each memory system and RAG (1rd) on MINT-Eval, with \texttt{Qwen3.5-4B} (top) and \texttt{Qwen3.6-27B} (bottom). }
\label{fig:venn_all11_mint_eval_s}
\end{figure*}

Figures~\ref{fig:venn_all11_longmemeval_s} to \ref{fig:venn_all11_mint_eval_s} show the Venn plot of per-query correctness overlap with RAG (1rd) for all memory systems on the three main benchmarks.

\subsection{Memory Tax - Faithfulness }
\label{apdx:faithfulness}

\paragraph{Source gold context}

Table~\ref{tab:gold_stats} summarizes the gold evidence of the three main benchmarks. Queries without gold are excluded from all gold-based analyses.

\catsq~\textbf{LongMemEval-s.} 479 of the 500 queries have gold turns, and the other 21 are abstention questions with no supporting turn by design.

\catsq~\textbf{PersonaMem-v2.} 836 of the 934 queries have gold turns. The other 98 are sensitive-information queries, whose relevant snippet is in the query itself and needs no memory.

\catsq~\textbf{MINT-Eval.} 765 of the 1,557 queries have gold steps, covering all state-tracking queries, verified version-specific questions in wiki and github, and verifiable dialog queries.

\begin{table}[ht]
\centering
\caption{Gold evidence in the three main benchmarks. Gold items are summed over queries.}
\label{tab:gold_stats}
\vspace{-0.6em}
\small
\setlength{\tabcolsep}{5pt}
\resizebox{\linewidth}{!}{%
\begin{tabular}{l c c c}
\toprule
\textbf{Benchmark} & \textbf{\#Queries} & \textbf{\#Queries w/ gold} & \textbf{\#Gold items} \\
\midrule
LongMemEval-s & 500 & 479 & 896 \\
PersonaMem-v2 & 934 & 836 & 2618 \\
MINT-Eval & 1557 & 765 & 2153 \\
\bottomrule
\end{tabular}%
}
\vspace{-1em}
\end{table}

\renewcommand{\dbltopfraction}{0.95}  
\begin{figure*}[p]   % 10-11 user: Fig 13 + Fig 14 on a page of their own
\centering
\includegraphics[width=0.82\textwidth]{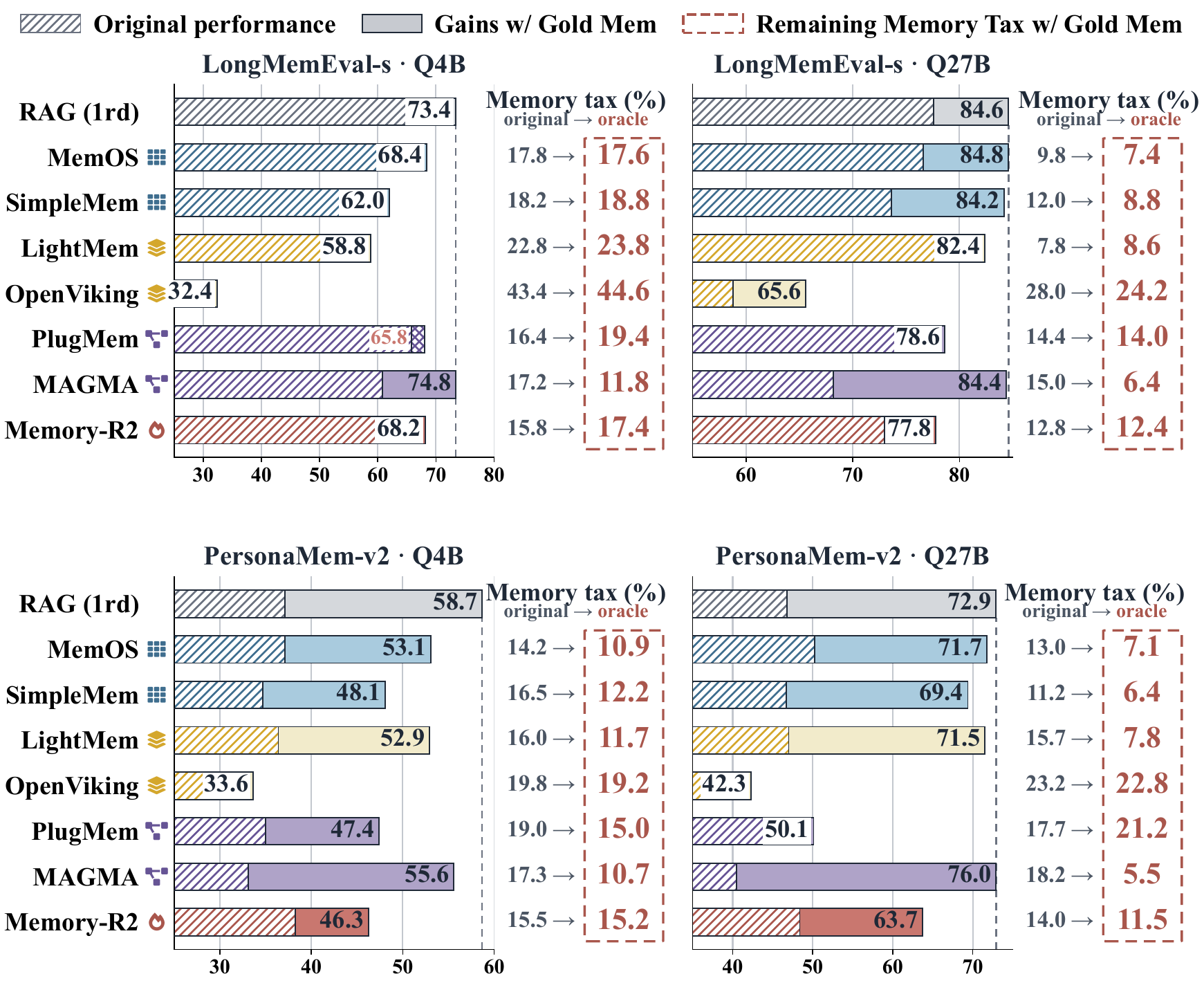}
\par\vspace{-0.6em}
\caption{QA accuracy with the oracle raw context vs.\ oracle memory units on LongMemEval-s and PersonaMem-v2 with \texttt{Qwen3.5-4B} and \texttt{Qwen3.6-27B}, with retrieval (hatched) as reference. Right of each panel: memory tax with normal retrieval vs.\ with oracle memory units.}
\label{fig:oracle_compare_all}
\vspace{-0.8em}
\end{figure*}

\begin{figure*}[p]
\centering
\includegraphics[width=\textwidth]{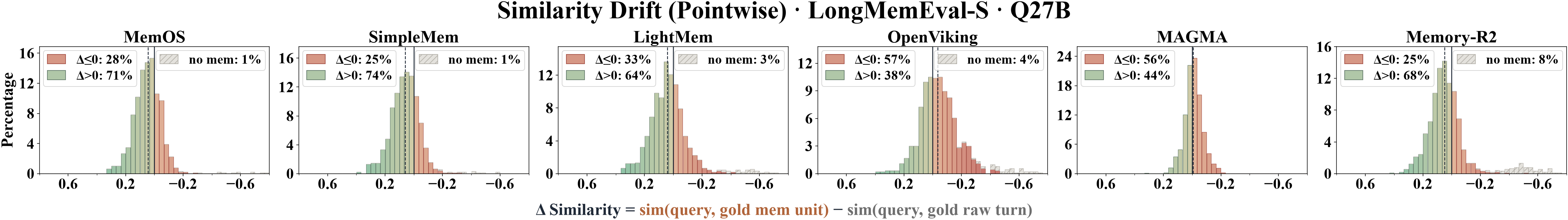}\\[1pt]
\includegraphics[width=\textwidth]{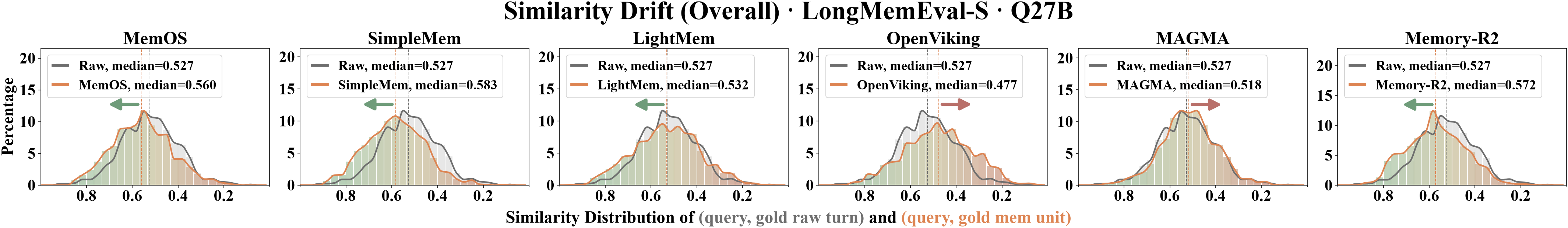}
\figsep
\includegraphics[width=\textwidth]{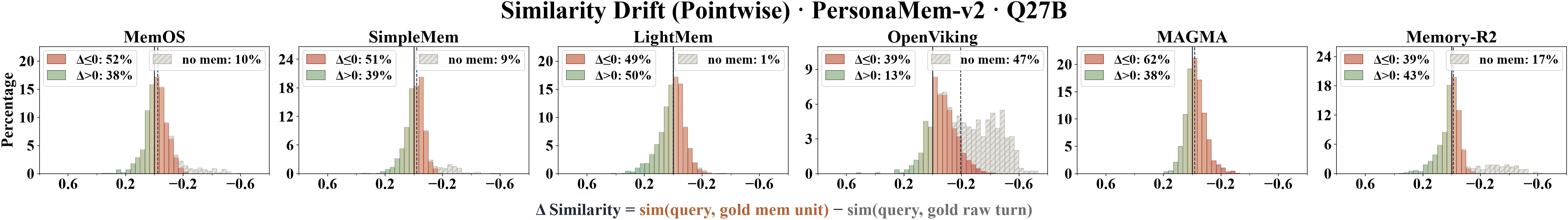}\\[1pt]
\includegraphics[width=\textwidth]{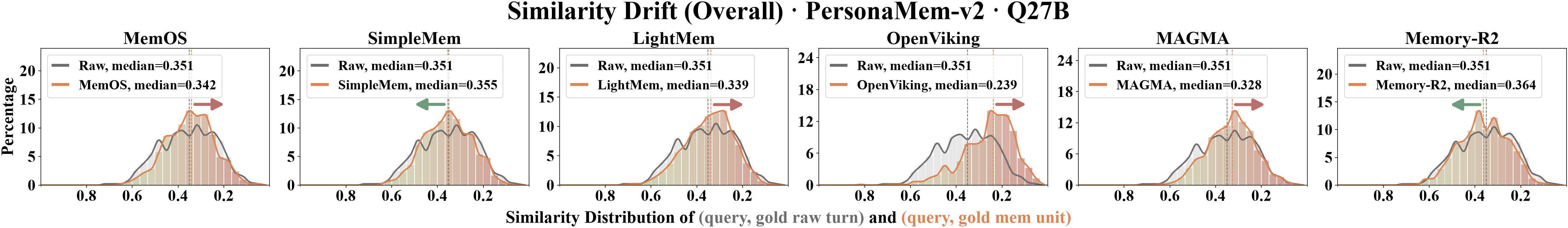}
\par\vspace{-0.6em}
\caption{Query similarity comparison of gold raw turns vs.\ gold memory units on LongMemEval-s (top) and PersonaMem-v2 (bottom) with \texttt{Qwen3.6-27B}. In each pair, point-wise similarity difference (top) and overall distributions (bottom).}
\label{fig:access_drift_all}
\vspace{-0.9em}
\end{figure*}

\paragraph{Mapping Gold Raw Turns to Gold Memory Units.}
Given the benchmark-provided gold raw turns, we map them to the corresponding units in each memory store. The mapping falls into three cases. A raw turn may correspond to \boxednum{1} a single deterministic memory unit, \boxednum{2} multiple candidate units, or \boxednum{3} no unit when the relevant information is lost during memory processing. \texttt{GPT-5.6 Terra} and \texttt{Claude Sonnet 5} independently select relevant memory units given the query, answer, and gold raw turn, with rare disagreements resolved by human adjudication.
\textsc{A-Mem}, \textsc{MemoryOS}, \textsc{HippoRAG2}, and \textsc{MAGMA} admit deterministic one-to-one mappings, while the remaining systems require this LLM-based filtering. Table~\ref{tab:gold_mapping} reports the resulting coverage and the average number of gold memory units per gold raw turn for some memory systems. To check the LLM-based selection, we manually inspect sampled cases for some systems and label each as correct, redundant, or missing (Table~\ref{tab:gold_filter_check}).

\begin{table}[t]
\centering
\caption{Mapping from gold raw turns to gold memory units. Coverage is the share of gold raw turns mapped to at least one memory unit, and Avg.\ units is the average number of gold memory units per gold raw turn.}
\label{tab:gold_mapping}
\vspace{-0.6em}
\small
\setlength{\tabcolsep}{3.5pt}
\resizebox{\linewidth}{!}{%
\begin{tabular}{l cccc}
\toprule
\multicolumn{5}{c}{\textbf{LongMemEval-s}} \\
\midrule
 & \textsc{MAGMA} & \textsc{A-Mem} & \textsc{MemoryOS} & \textsc{SimpleMem} \\
\cmidrule(lr){2-5}
Coverage & 100.0\% & 100.0\% & 99.7\% & 98.7\% \\
Avg.\ units & 1.00 & 1.00 & 0.99 & 1.59 \\
\midrule
 & \textsc{LightMem} & \textsc{OpenViking} & \textsc{PlugMem} & \textsc{Memory-R2} \\
\cmidrule(lr){2-5}
Coverage & 96.8\% & 95.5\% & 95.5\% & 92.3\% \\
Avg.\ units & 2.78 & 1.99 & 2.07 & 1.28 \\
\midrule\midrule
\multicolumn{5}{c}{\textbf{PersonaMem-v2}} \\
\midrule
 & \textsc{MAGMA} & \textsc{A-Mem} & \textsc{MemoryOS} & \textsc{SimpleMem} \\
\cmidrule(lr){2-5}
Coverage & 100.0\% & 100.0\% & 99.9\% & 90.9\% \\
Avg.\ units & 1.00 & 1.00 & 1.00 & 1.43 \\
\midrule
 & \textsc{LightMem} & \textsc{OpenViking} & \textsc{PlugMem} & \textsc{Memory-R2} \\
\cmidrule(lr){2-5}
Coverage & 99.0\% & 52.8\% & 75.9\% & 82.6\% \\
Avg.\ units & 5.78 & 0.88 & 1.99 & 1.41 \\
\bottomrule
\end{tabular}%
}
\vspace{-1em}
\end{table}

\begin{table}[t]
\centering
\caption{Human spot check of LLM-selected gold memory units (\%), labeled as correct, extraneous (irrelevant units are selected), or missing (relevant units left out).}
\label{tab:gold_filter_check}
\vspace{-0.6em}
\small
\setlength{\tabcolsep}{3.5pt}
\resizebox{\linewidth}{!}{%
\begin{tabular}{l ccc ccc}
\toprule
\multirow{2}{*}{\textbf{Method}} & \multicolumn{3}{c}{\textbf{LongMemEval-s}} & \multicolumn{3}{c}{\textbf{PersonaMem-v2}} \\
\cmidrule(lr){2-4} \cmidrule(lr){5-7}
 & Correct & Extraneous & Missing & Correct & Extraneous & Missing \\
\midrule
\textsc{SimpleMem} & 92.2 & 5.8 & 2.0 & 91.7 & 5.0 & 3.3 \\
\textsc{LightMem} & 85.0 & 14.2 & 0.8 & 88.5 & 8.2 & 3.3 \\
\textsc{MemOS} & 88.7 & 8.2 & 3.2 & 87.3 & 12.7 & 0.0 \\
\textsc{Memory-R2} & 83.3 & 15.8 & 0.8 & 85.9 & 12.5 & 1.6 \\
\bottomrule
\end{tabular}%
}
\vspace{-1em}
\end{table}

\paragraph{Oracle raw context vs. oracle memory units}
Figure~\ref{fig:oracle_compare_all} extends Figure~\ref{fig:oracle_compare} to both Qwen backbones on LongMemEval-s and PersonaMem-v2. \boxednum{1} \textbf{Some systems show a clear gap between oracle raw context and oracle memory units}, e.g., \textsc{SimpleMem}, \textsc{OpenViking} and \textsc{Memory-R2} on LongMemEval-s with Qwen3.5-4B and \textsc{PlugMem} on PersonaMem-v2 with Qwen3.6-27B, indicating information loss during memory processing. \boxednum{2} For \textsc{MAGMA} and \textsc{LightMem}, oracle-memory performance is close to oracle-raw performance yet much higher than their retrieval performance, suggesting that the evidence is preserved but harder to recover, which motivates the accessibility analysis.

\paragraph{Human Annotation of Information Preservation.}
For the micro-level analysis, we sample 500 raw turns from the corpus of three benchmarks and trace each to its corresponding memory unit(s), without restricting the samples to gold evidence. For one-to-many mappings, we present annotators with all corresponding memory units without LLM filtering, \textbf{avoiding any model-based preselection before human judgment.} Annotators compare the raw turn with the full set of memory units and label its information as fully preserved, partially missing, largely missing, or having no corresponding memory unit (Table~\ref{tab:atomic_facts}). Redundant memory units are allowed as long as the raw-turn information is fully preserved.

\clearpage   % flush Fig 13 + Fig 14 onto their own page before C.2
\subsection{Memory Tax - Accessibility }
\label{apdx:accessibility}

\paragraph{Overview}
We group the retrieval mechanisms of memory systems into four types in a representative but not exhaustive way: \catsq~\textbf{ranking} scores memory units, e.g., by embedding similarity, lexical scores or a reranker; \catsq~\textbf{matching} selects units by keywords or metadata filters; \catsq~\textbf{expansion} grows the retrieved set along the memory structure, e.g., via graph neighbors or personalized PageRank; and \catsq~\textbf{persistence} always places part of the memory, such as a user profile, preferences, knowledge entries, etc. into the context. Table~\ref{tab:retrieval_mechanisms} lists the types each system uses.

\begin{table}[ht]
\centering
\caption{Retrieval mechanisms used by each memory system at query time.}
\label{tab:retrieval_mechanisms}
\vspace{-0.6em}
\small
\setlength{\tabcolsep}{5pt}
\begin{adjustbox}{max width=\linewidth}
\begin{tabular}{l l c c c c}
\toprule
\textbf{Type} & \textbf{Method} & \textbf{Ranking} & \textbf{Matching} & \textbf{Expansion} & \textbf{Persistence} \\
\midrule
\multirow{3}{*}{\clsFlat{}Flat store} & \textsc{A-Mem} & \cmark & -- & -- & -- \\
 & \textsc{SimpleMem} & \cmark & \cmark & -- & -- \\
 & \textsc{MemOS} & \cmark & \cmark & -- & \cmark \\
\midrule
\multirow{3}{*}{\clsHier{}Hier/Lifecycle} & \textsc{MemoryOS} & \cmark & -- & -- & \cmark \\
 & \textsc{LightMem} & \cmark & -- & -- & -- \\
 & \textsc{OpenViking} & \cmark & -- & \cmark & -- \\
\midrule
\multirow{3}{*}{\clsGraph{}Graph-based} & \textsc{HippoRAG2} & \cmark & -- & \cmark & -- \\
 & \textsc{MAGMA} & \cmark & \cmark & \cmark & -- \\
 & \textsc{PlugMem} & \cmark & -- & \cmark & -- \\
\midrule
\multirow{2}{*}{\clsTrain{}Trained Policy} & \textsc{Mem-T} & \cmark & -- & -- & -- \\
 & \textsc{Memory-R2} & \cmark & -- & -- & -- \\
\bottomrule
\end{tabular}
\end{adjustbox}
\vspace{-1em}
\end{table}

\paragraph{Iterative Retrieval Replay.}
Following the definitions of gold raw turns and gold memory units in Appendix~\ref{apdx:faithfulness}, each query has one or more gold raw turns and their corresponding gold memory units. We repeatedly run retrieval to measure how many rounds are needed to recover all gold memory units. After each round, the retrieved units are excluded from the remaining memory store before the next round. The process continues until all gold units are recovered or the round limit is reached. For most baselines, this process can be reproduced offline from the stored memory and retrieval state. While for those systems involving graph expansion or LLM-based query refinement, we need to rerun the retrieval process.
Figure~\ref{fig:access_rounds} shows that memory systems generally recover fewer gold units in the first round than RAG.

\paragraph{Similarity Drift.}
At the micro level, we compare $\mathrm{sim}(q, g_{\mathrm{raw}})$ with $\mathrm{sim}(q, g_{\mathrm{mem}})$ for each gold raw turn and its gold memory unit(s). \textbf{This analysis is only indicative, since each system retrieves within its own memory store, so absolute similarity values are not directly comparable across systems.} Some memory systems also do not always rank units by embedding similarity only, so we focus on the within-system shift from raw evidence to its memory representation.

We report both point-wise changes and overall similarity distributions. For point-wise analysis, when multiple units correspond to the same raw turn, we use their median similarity to reduce the influence of extreme values.

\subsection{Memory Tax - Robustness }
\label{apdx:robustness}

\begin{figure}[t]
\centering
\includegraphics[width=\linewidth]{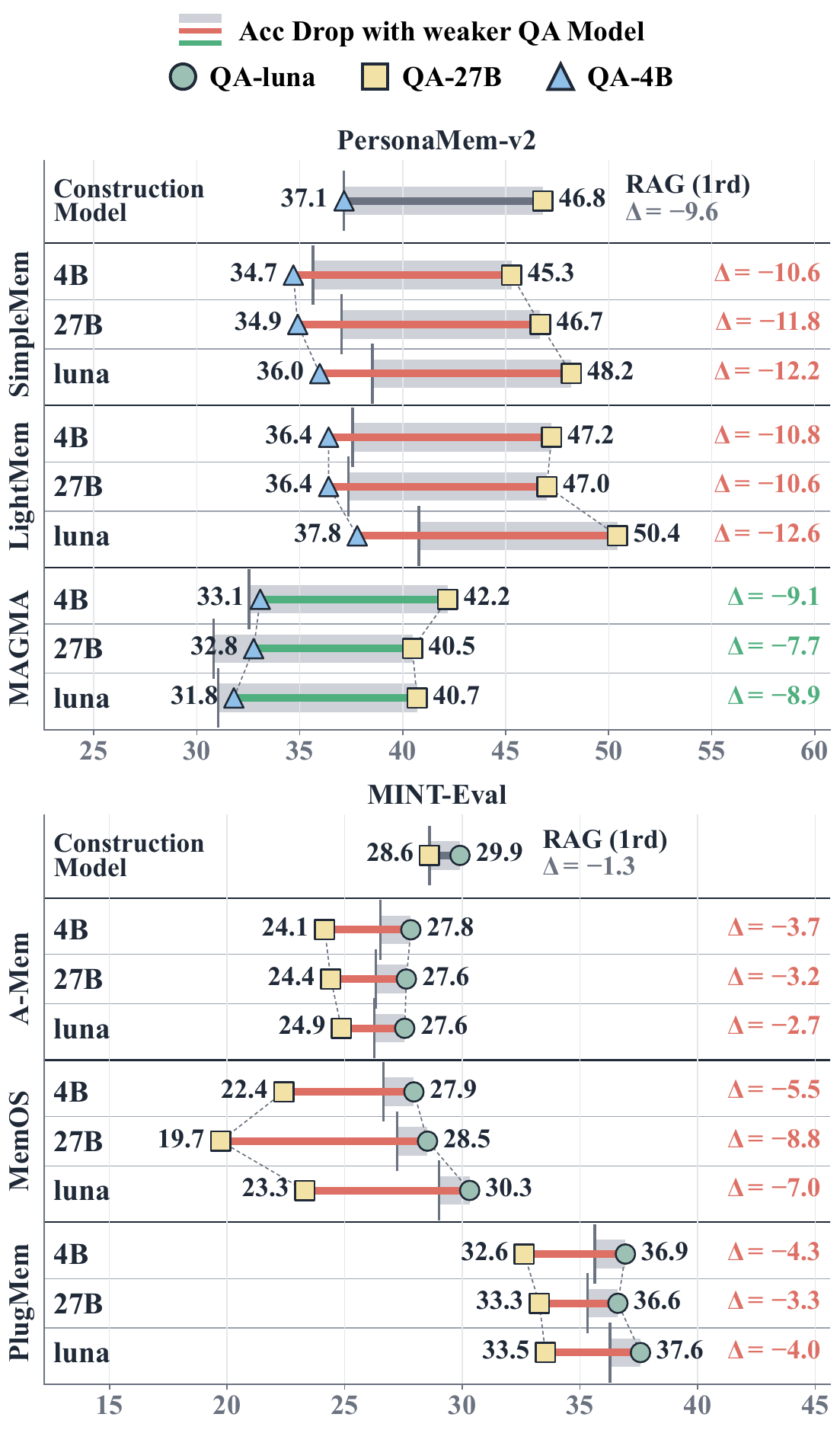}
\par\vspace{-0.6em}
\caption{Cross-backbone robustness of memory systems (vertical version of Figure~\ref{fig:robustness}). Segments are \textcolor{TaxRed}{red} when the drop exceeds that of RAG (grey block) and \textcolor{BenefitGreen}{green} otherwise.}
\label{fig:robustness_apdx}
\vspace{-0.8em}
\end{figure}

For each memory system, we construct memory with each of the three backbones and answer over each resulting memory with all three as the QA model, keeping the retrieved memory units fixed whenever retrieval does not call the LLM (otherwise it is replayed with the new QA model); naive RAG builds its index without any LLM, so it only varies with the QA model. Figure~\ref{fig:robustness_apdx} shows representative systems with scores annotated. Each arrow goes from the score with the stronger QA model to that with the weaker one, the grey bar behind it has the length of RAG's drop under the same downgrade, and the dashed lines connect the same QA model across construction models.

% Figure~\ref{fig:robust_all} gives the full cross-backbone results for several memory systems on the main benchmarks, where memory is built with one backbone (rows) and queried with another (columns).

\renewcommand{\dbltopfraction}{0.95}\clearpage
\begin{table*}[!b]
\centering
\caption{Score and retrieval scope (RS) of the retrieval-only baselines on each benchmark. Backbone is \texttt{Qwen3.6-27B}. All scores use the judge of Appendix~\ref{apdx:judge_prompts}. Benchmarks are ordered by the score of RAG-Hybrid.}
\label{tab:regimes_full}
\vspace{-0.6em}
\small
\setlength{\tabcolsep}{4pt}
\begin{adjustbox}{max width=0.9\textwidth}
\begin{tabular}{l cc cc cc cc}
\toprule
 & \multicolumn{2}{c}{\textbf{EverMemBench-MCQ} ($k$=30)} & \multicolumn{2}{c}{\textbf{HaluMem-Medium} ($k$=10)} & \multicolumn{2}{c}{\underline{\textbf{LongMemEval-s}} ($k$=10)} & \multicolumn{2}{c}{\textbf{LoCoMo} ($k$=10)} \\
 & \multicolumn{2}{c}{\footnotesize Avg Corpus Len (549k tokens)} & \multicolumn{2}{c}{\footnotesize Avg Corpus Len (164k tokens)} & \multicolumn{2}{c}{\footnotesize Avg Corpus Len (106k tokens)} & \multicolumn{2}{c}{\footnotesize Avg Corpus Len (20.4k tokens)} \\
\cmidrule(lr){2-3} \cmidrule(lr){4-5} \cmidrule(lr){6-7} \cmidrule(lr){8-9}
\textbf{Method} & Score & RS & Score & RS & Score & RS & Score & RS \\
\midrule
IDF-keyword & 68.2 & 0.40\% & 52.5 & 0.52\% & 46.2 & 2.49\% & 58.8 & 2.85\% \\
RAG-Emb & 80.9 & 0.50\% & 69.7 & 0.60\% & 72.6 & 1.58\% & 69.6 & 3.38\% \\
RAG-Hybrid & 82.7 & 0.50\% & 79.9 & 0.61\% & 77.6 & 1.96\% & 75.3 & 3.43\% \\
\midrule
 & \multicolumn{2}{c}{\textbf{Mem-Gallery} ($k$=5)} & \multicolumn{2}{c}{\textbf{PersonaMem-v1} ($k$=5)} & \multicolumn{2}{c}{\textbf{ES-MemEval} ($k$=3)} & \multicolumn{2}{c}{\textbf{MemoryAgentBench} ($k$=10)} \\
 & \multicolumn{2}{c}{\footnotesize Avg Corpus Len (27.1k tokens)} & \multicolumn{2}{c}{\footnotesize Avg Corpus Len (65.0k tokens)} & \multicolumn{2}{c}{\footnotesize Avg Corpus Len (11.7k tokens)} & \multicolumn{2}{c}{\footnotesize Avg Corpus Len (193k tokens)} \\
\cmidrule(lr){2-3} \cmidrule(lr){4-5} \cmidrule(lr){6-7} \cmidrule(lr){8-9}
\textbf{Method} & Score & RS & Score & RS & Score & RS & Score & RS \\
\midrule
IDF-keyword & 48.5 & 3.06\% & 57.9 & 0.69\% & 55.6 & 3.74\% & 55.5 & 2.42\% \\
RAG-Emb & 72.4 & 3.56\% & 69.2 & 0.60\% & 58.7 & 4.02\% & 59.4 & 4.04\% \\
RAG-Hybrid & 74.5 & 3.64\% & 71.1 & 0.61\% & 69.2 & 4.01\% & 64.8 & 4.10\% \\
\midrule
 & \multicolumn{2}{c}{\textbf{LongMemEval-m} ($k$=10)} & \multicolumn{2}{c}{\textbf{WorldMemArena} ($k$=5)} & \multicolumn{2}{c}{\textbf{AMA-Bench} ($k$=10)} & \multicolumn{2}{c}{\textbf{KnowMe-Bench} ($k$=20)} \\
 & \multicolumn{2}{c}{\footnotesize Avg Corpus Len (1044k tokens)} & \multicolumn{2}{c}{\footnotesize Avg Corpus Len (25.0k tokens)} & \multicolumn{2}{c}{\footnotesize Avg Corpus Len (32.4k tokens)} & \multicolumn{2}{c}{\footnotesize Avg Corpus Len (309k tokens)} \\
\cmidrule(lr){2-3} \cmidrule(lr){4-5} \cmidrule(lr){6-7} \cmidrule(lr){8-9}
\textbf{Method} & Score & RS & Score & RS & Score & RS & Score & RS \\
\midrule
IDF-keyword & 41.2 & 0.31\% & 38.0 & 15.71\% & 29.0 & 13.42\% & 17.6 & 0.43\% \\
RAG-Emb & 54.2 & 0.15\% & 52.5 & 23.14\% & 39.4 & 17.84\% & 40.9 & 0.95\% \\
RAG-Hybrid & 64.4 & 0.20\% & 55.6 & 26.14\% & 48.4 & 17.37\% & 47.0 & 0.94\% \\
\midrule
 & \multicolumn{2}{c}{\underline{\textbf{PersonaMem-v2}} ($k$=10)} & \multicolumn{2}{c}{\textbf{RealMem} ($k$=10)} & \multicolumn{2}{c}{\underline{\textbf{MINT-Eval}} ($k$=35/5/3/5)} & \multicolumn{2}{c}{} \\
 & \multicolumn{2}{c}{\footnotesize Avg Corpus Len (123k tokens)} & \multicolumn{2}{c}{\footnotesize Avg Corpus Len (210k tokens)} & \multicolumn{2}{c}{\footnotesize Avg Corpus Len (131k tokens)} & \multicolumn{2}{c}{} \\
\cmidrule(lr){2-3} \cmidrule(lr){4-5} \cmidrule(lr){6-7}
\textbf{Method} & Score & RS & Score & RS & Score & RS & &  \\
\midrule
IDF-keyword & 39.5 & 1.47\% & 25.2 & 1.40\% & 20.4 & 6.63\% & &  \\
RAG-Emb & 47.3 & 1.36\% & 29.3 & 1.08\% & 24.7 & 6.58\% & &  \\
RAG-Hybrid & 46.8 & 1.48\% & 32.9 & 1.22\% & 28.6 & 6.73\% & &  \\
\bottomrule
\end{tabular}
\end{adjustbox}
\vspace{-1em}
\end{table*}

\renewcommand{\dbltopfraction}{0.7}
\subsection{Benchmark Headroom Further Exposes the Memory Tax}
\label{apdx:regimes}

We probe the benchmark headroom with the retrieval-only controls on a broader set of long-term memory benchmarks, listed here in order of release date (including the three benchmarks used in our main experiments): LoCoMo~\citep{maharana2024evaluatinglongtermconversationalmemory}, LongMemEval-s and LongMemEval-m~\citep{wu2025longmemevalbenchmarkingchatassistants}, PersonaMem-v1~\citep{jiang2025knowmerespondme}, MemoryAgentBench~\citep{hu2025evaluatingmemoryllmagents}, HaluMem-Medium~\citep{chen2026halumemevaluatinghallucinationsmemory}, PersonaMem-v2~\citep{jiang2025personamemv2personalizedintelligencelearning}, MINT-Eval~\citep{lee2026mintevalevaluatingmemorymultitarget}, KnowMe-Bench~\citep{wu2026knowmebenchbenchmarkingpersonunderstanding}, RealMem~\citep{bian2026realmembenchmarkingllmsrealworld}, EverMemBench-MCQ~\citep{hu2026evaluatinglonghorizonmemorymultiparty}, AMA-Bench~\citep{zhao2026amabenchevaluatinglonghorizonmemory}, ES-MemEval~\citep{Chen_2026}, and the multimodal benchmarks Mem-Gallery~\citep{bei2026memgallerybenchmarkingmultimodallongterm} and WorldMemArena~\citep{liu2026worldmemarenaevaluatingmultimodalagent}.

\paragraph{Simple baselines.}
We run the four naive baselines of Section~\ref{sec:regimes}. For IDF-keyword, we select for each query the two terms with the highest inverse document frequency (IDF) in the corpus and prioritize chunks containing both terms, followed by those containing either one. Within each tier, chunks are sampled uniformly at random until the top-$k$ budget is filled.
All retrieval baselines follow the default settings of Section~\ref{sec:exp_setup}. Their top-$k$ and retrieval scope are listed in Table~\ref{tab:regimes_full}.
\begin{figure*}[t]
\centering
\includegraphics[width=0.7\textwidth]{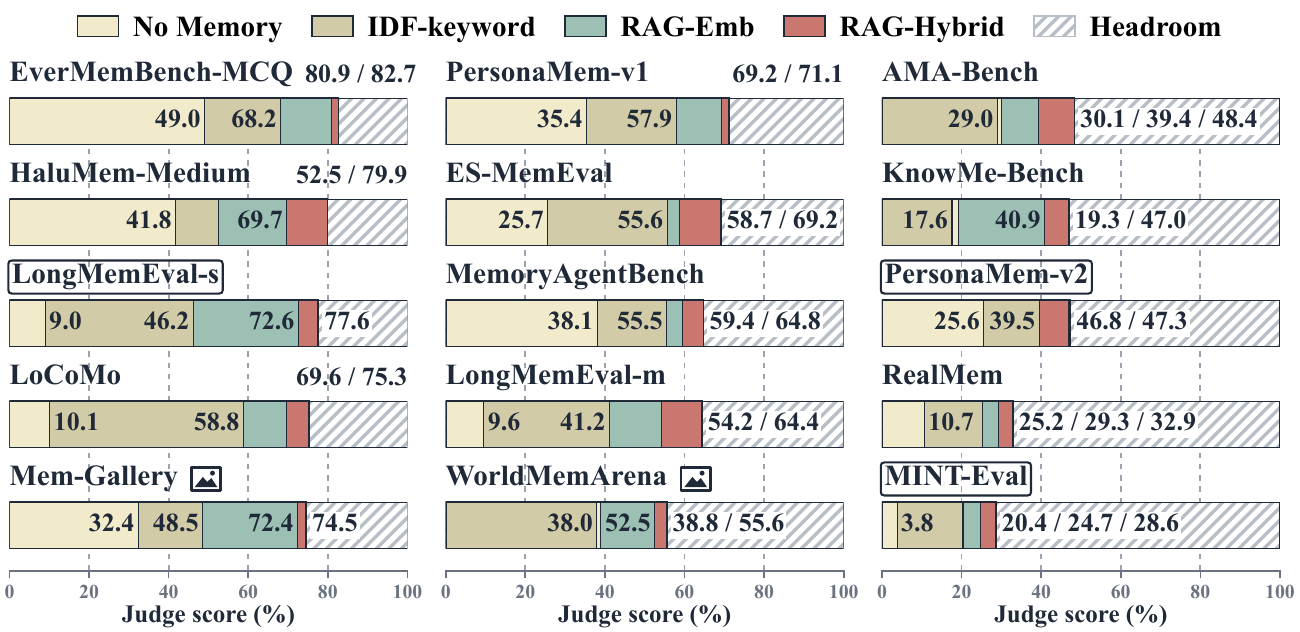}
\par\vspace{-0.6em}
\caption{Simple retrieval baselines across benchmarks with \texttt{Qwen3.6-27B}. \faImage[regular]~marks multimodal benchmarks.}
\label{fig:regimes_apdx}
\vspace{-0.8em}
\end{figure*}

\paragraph{Evaluation protocol.}
All predictions are judged against the reference answers with the binary LLM judge of Appendix~\ref{apdx:eval_metrics}. Deviations from the official evaluation setups are summarized below.

\catsq~\textbf{PersonaMem-v2, MINT-Eval, WorldMemArena}: To reduce evaluation cost, we sample a subset of examples while approximately preserving the original distribution of question types.

\catsq~\textbf{LoCoMo}: Following prior works, we exclude the adversarial category (category 5, abstention) and evaluate the remaining four categories.

\catsq~\textbf{ES-MemEval}: We evaluate only the QA task, as the other two task types do not have deterministic reference answers.

\catsq~\textbf{Multimodal benchmarks} (Mem-Gallery, WorldMemArena): We use a text-only retrieval setting for controlled comparison. Each image is first converted into a roughly 200-token caption by a VLM (\texttt{\textbf{Qwen3.6-27B}}), after which all retrieval baselines operate on the resulting captioned history.

% See more case studies demonstrating where the limited headroom comes from in Appendix~\ref{apdx:case_regimes}.

\renewcommand{\dbltopfraction}{0.95}
\begin{table*}[b]
\centering
\caption{Full results of experiential memory with \texttt{Qwen3.5-4B}: task score, tokens per task (in thousands, with the cost of building the experience library amortised over the test tasks) and average steps (dialogue turns for $\tau^2$-Bench). Best score per benchmark in \textbf{bold}.}
\label{tab:generalization_full}
\vspace{-0.6em}
\small
\setlength{\tabcolsep}{4pt}
\resizebox{0.88\textwidth}{!}{%
\begin{tabular}{l|ccc|ccc|ccc}
\toprule
\multirow{2}{*}{\textbf{Method}} & \multicolumn{3}{c}{\textbf{ALFWorld} \textit{(Embodied)}} & \multicolumn{3}{c}{\textbf{WebShop} \textit{(Web)}} & \multicolumn{3}{c}{\textbf{Mind2Web} \textit{(Web)}} \\
\cmidrule(lr){2-4} \cmidrule(lr){5-7} \cmidrule(lr){8-10}
 & SR & Avg Tokens (K) & Avg Steps & Score & Avg Tokens (K) & Avg Steps & StepSR & Avg Tokens (K) & Avg Steps \\
\midrule
No Memory & 52.2 & 75.9 & 32.1 & \textbf{45.9} & 9.4 & 7.3 & 48.5 & 3.2 & N/A$^\ddagger$ \\
Naive RAG & 73.1 & 234.8 & 22.7 & 38.4 & 36.8 & 8.8 & 45.3 & 9.6 & N/A$^\ddagger$ \\
\midrule
\textsc{ReasoningBank} & 62.7 & 254.0 & 29.7 & 34.3 & 40.1 & 9.7 & \textbf{49.9} & 11.4 & N/A$^\ddagger$ \\
\textsc{Mem}$^p$ & \textbf{76.1} & 237.7 & 23.4 & 44.5 & 37.6 & 7.8 & 47.9 & 6.4 & N/A$^\ddagger$ \\
\textsc{G-Memory} & 73.9 & 259.6 & 24.4 & 39.5 & 65.1 & 8.4 & 47.0 & 14.4 & N/A$^\ddagger$ \\
\textsc{SkillPro} & 47.0 & 1226.3 & 38.9 & 42.5 & 297.7 & 8.2 & 43.6 & 62.0 & N/A$^\ddagger$ \\
\midrule\midrule
\multirow{2}{*}{\textbf{Method}} & \multicolumn{3}{c}{\textbf{BigCodeBench} \textit{(Coding)}} & \multicolumn{3}{c}{\textbf{AppWorld} \textit{(Tool Use)}} & \multicolumn{3}{c}{\textbf{$\tau^2$-Bench} \textit{(Tool Use)}} \\
\cmidrule(lr){2-4} \cmidrule(lr){5-7} \cmidrule(lr){8-10}
 & pass@1 & Avg Tokens (K) & Avg Steps & TGC & Avg Tokens (K) & Avg Steps & pass\^\ 1 & Avg Tokens (K) & Avg Steps \\
\midrule
No Memory & 16.9 & 0.6 & 1$^\dagger$ & 12.5 & 278.3 & 22.4 & 61.0 & 154.2 & 16.7 \\
Naive RAG & \textbf{21.6} & 5.2 & 1$^\dagger$ & \textbf{22.6} & 339.6 & 20.0 & 69.0 & 394.5 & 14.3 \\
\midrule
\textsc{ReasoningBank} & 14.9 & 13.7 & 1$^\dagger$ & 14.3 & 383.7 & 22.1 & 51.0 & 381.8 & 15.0 \\
\textsc{Mem}$^p$ & 20.3 & 8.9 & 1$^\dagger$ & 14.3 & 317.1 & 20.9 & 62.0 & 396.2 & 14.6 \\
\textsc{G-Memory} & 20.3 & 62.9 & 1$^\dagger$ & \textbf{22.6} & 388.7 & 21.0 & \textbf{78.0} & 424.3 & 13.8 \\
\textsc{SkillPro} & 11.5 & 29.3 & 1$^\dagger$ & 10.7 & 661.9 & 22.7 & 55.0 & 1334.3 & 15.5 \\
\bottomrule
\end{tabular}%
}
\par\vspace{2pt}\parbox{0.86\textwidth}{\footnotesize\raggedright $^\dagger$BigCodeBench is single-shot generation, where each task takes one model call, with no interaction with an environment. $^\ddagger$Mind2Web is evaluated offline step by step on the gold trajectories, so the number of steps is fixed by the dataset, not chosen by the agent.}
\vspace{-1em}
\end{table*}

\begin{figure*}[ht]
\centering
\includegraphics[width=\textwidth]{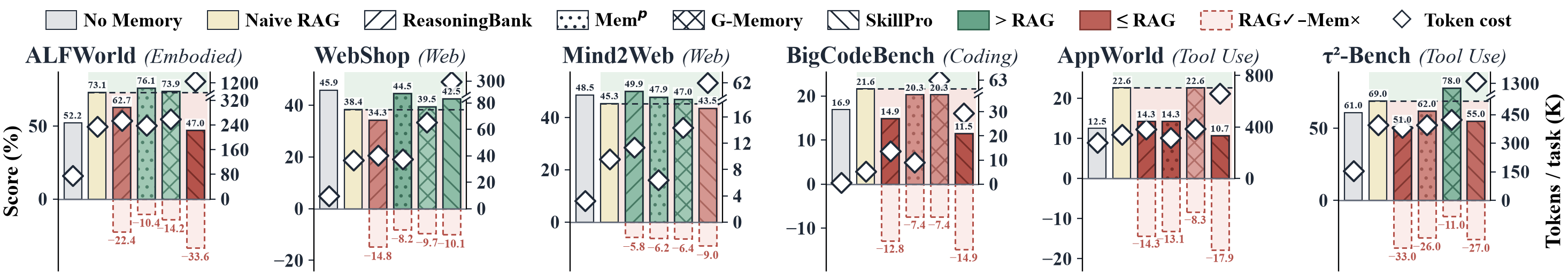}
\par\vspace{-0.6em}
\caption{Experiential memory on agentic tasks, task score (bars, left axis) and tokens per task (diamonds, right axis). Memory baselines are green when above Naive RAG and red otherwise, with darker shades for larger relative gaps. Dashed boxes below zero show the memory tax, the score that Naive RAG obtains where the memory fails.}
\label{fig:generalization_apdx}
\vspace{-0.6em}
\end{figure*}

\renewcommand{\dbltopfraction}{0.7}
\subsection{Generalizing Memory Tax}
\label{apdx:generalization}

\subsubsection{Experiential Memory}
\label{apdx:gen_experiential}

\paragraph{Benchmark Details.}
Experience is accumulated on the training (or dev) split and evaluated on test split, with task counts provided (experience / test).

\textbf{ALFWorld}~\citep{shridhar2021alfworldaligningtextembodied}: text-based household tasks (300 / 134).

\textbf{WebShop}~\citep{yao2023webshopscalablerealworldweb}: searching for and buying a product on a simulated e-commerce website (1,000 / 500).

\textbf{Mind2Web}~\citep{deng2023mind2webgeneralistagentweb}: navigation on real-world websites, tested on the cross-task, cross-website and cross-domain splits (1,009 / 1,341).

\textbf{BigCodeBench}~\citep{zhuo2025bigcodebenchbenchmarkingcodegeneration}: Python programming with calls to diverse libraries, tested on the Hard subset (992 / 148).

\textbf{AppWorld}~\citep{trivedi2024appworldcontrollableworldapps}: everyday tasks solved by writing code that calls app APIs, tested on test-normal and test-challenge (147 / 585).

\textbf{$\tau^2$-Bench}~\citep{barres2025tau2benchevaluatingconversationalagents}: customer service with domain policies, tools and a simulated user in the airline, retail and telecom domains (178 / 100).

\paragraph{Baseline Details.}

\catsq~\textbf{No Memory}: the agent interacts with the environment purely in the ReAct ~\citep{yao2023reactsynergizingreasoningacting}, with no memory across tasks.

\catsq~\textbf{Naive RAG}: At the start of each task, we retrieve the top-2 previously accumulated tasks based on their description similarity to the current task and include their raw trajectories in the prompt. We use the same embedding model as in Section~\ref{sec:exp_setup}, \texttt{Qwen3-Embedding-0.6B}.

\catsq~\textbf{\textsc{Mem}$^p$}~\citep{fang2026mempexploringagentprocedural}: distills past trajectories into procedural memory, i.e., step-by-step workflows, and retrieves those of similar tasks as guidelines.

\catsq~\textbf{\textsc{ReasoningBank}}~\citep{ouyang2026reasoningbankscalingagentselfevolving}: induces generalizable reasoning strategies from both successful and failed trajectories as structured memory items, retrieved by task similarity.

\catsq~\textbf{\textsc{G-Memory}}~\citep{zhang2025gmemorytracinghierarchicalmemory}: organizes experience in a three-tier graph hierarchy of insights, queries and interaction trajectories, and retrieves both high-level insights and condensed past trajectories from it.

\catsq~\textbf{\textsc{SkillPro}}~\citep{mi2026skillprolearningreusableskills}: maintains an evolving pool of skills, from which the agent selects one to follow at each step, and the skills are refined from the feedback of the tasks.

\paragraph{Implementation Details.}
We follow the default task protocols of each benchmark consistent with prior work, including a step limit of 50 for ALFWorld, 15 for Webshop, 50 for AppWorld and 50 for $\tau^2$-Bench, while Mind2Web imposes no explicit step limit. We follow each benchmark's official metric: success rate for ALFWorld, task score for WebShop, step success rate for Mind2Web, pass@1 for BigCodeBench, pass\^\ 1 for $\tau^2$-Bench, and Task Goal Completion (TGC) for AppWorld, where a task succeeds only if all associated unit tests pass. Although pass@$k$ and pass\^\ k differ in definition, they are equivalent when $k=1$. 

\subsubsection{World-Model Memory}
\label{apdx:gen_world_model}

\paragraph{Benchmark Details.}
2 benchmarks adopted: 

\catsq~\textbf{Context-as-Memory Test set}~\citep{yu2025context}. We use 60 samples from it. Starting from the first-frame input image, each sample generates four video segments in sequence that complete a revisit, e.g., with instruction of turning right by 45$^\circ$, right by 45$^\circ$, left by 45$^\circ$ and left by 45$^\circ$, which return the camera to its original position. We then compare the final frame with the original view and report PSNR, SSIM and 1 $-$ LPIPS, all higher is better.

\catsq~\textbf{MemoBench}~\citep{chen2026memobench}. A target object undergoes a physical process (e.g., dissolution), leaves the view through camera motion, and must reappear in its updated state. We test 25\% of it, i.e., 90 samples.

\paragraph{Baseline Details.} 6 baselines adopted: 

\catsq~\textbf{\textsc{I3DM}}~\citep{li2026i3dm} retrieves the past frames most relevant to the current viewpoint with implicit 3D-aware matching and injects them as conditioning, so its memory is the naive past frames.
\catsq~\textbf{\textsc{WorldKV}}~\citep{yi2026worldkv} keeps the key-value cache of past frames as world memory, and retrieves and compresses the cached entries for the current generation step.
\catsq~\textbf{\textsc{ReWorld}}~\citep{chen2026reworld} an interactive world model that carries long-horizon memory in a learned, bounded key-value memory updated as the video is generated.
\catsq~\textbf{\textsc{MagicWorld}}~\citep{li2025magicworld} stores the visual latents of past frames and reuses them as context to keep long-horizon exploration stable.
\catsq~\textbf{\textsc{FramePack}}~\citep{zhang2026frame} packs the history into a fixed-length context of visual latents, compressing older frames more aggressively, with sampling that prevents drift.
\catsq~\textbf{\textsc{HyDRA}}~\citep{chen2026out} summarizes the history into a small set of memory tokens that the generator attends to.

\begin{table}[t]
\centering
\caption{Full long-horizon consistency results of world models on C-as-M-eval and MemoBench.}
\vspace{-0.6em}
\label{tab:world_model_full}
\small
\setlength{\tabcolsep}{5pt}
\begin{adjustbox}{max width=\linewidth}
\begin{tabular}{l l ccc}
\toprule
\textbf{Method} & \begin{tabular}{@{}c@{}}\textbf{Mem}\\\textbf{Carrier}\end{tabular} & PSNR $\uparrow$ & SSIM $\uparrow$ & 1 - LPIPS $\uparrow$ \\
\midrule
\multicolumn{5}{c}{\cellcolor{black!6}\textbf{C-as-M-eval}} \\
\midrule
\textsc{I3DM} & naive frames & {\normalsize\textbf{19.8}} & {\normalsize\textbf{.560}} & {\normalsize\textbf{.704}} \\
\textsc{WorldKV} & raw KV & 13.5 & .372 & .459 \\
\textsc{ReWorld} & learned KV & \underline{14.1} & .378 & \underline{.510} \\
\textsc{MagicWorld} & visual latents & 13.4 & \underline{.388} & .341 \\
\textsc{FramePack} & visual latents & 9.1 & .217 & .312 \\
\textsc{HyDRA} & mem. tokens & 11.7 & .320 & .319 \\
\midrule
\multicolumn{5}{c}{\cellcolor{black!6}\textbf{MemoBench}} \\
\midrule
\textsc{I3DM} & naive frames & {\normalsize\textbf{14.4}} & \underline{.430} & {\normalsize\textbf{.476}} \\
\textsc{WorldKV} & raw KV & 12.6 & .379 & .420 \\
\textsc{ReWorld} & learned KV & 12.9 & .387 & \underline{.422} \\
\textsc{MagicWorld} & visual latents & \underline{13.9} & {\normalsize\textbf{.433}} & .412 \\
\textsc{FramePack} & visual latents & 9.7 & .279 & .303 \\
\textsc{HyDRA} & mem. tokens & 11.9 & .355 & .326 \\
\bottomrule
\end{tabular}
\end{adjustbox}
\vspace{-1em}
\end{table}

\paragraph{Implementation Details.}
All methods run on NVIDIA H100 GPUs with 50 inference steps.

\newpage   % 10-11 user: column break only, so Appendix D starts in the right column next to C.7

\section{Case Studies}
\label{apdx:case_studies}

In this section, we show some representative case studies for both \textcolor{BenefitGreen}{Memory Benefit} (\ref{apdx:case_mem_benefit}) and \textcolor{TaxRed}{Memory Tax} (
\ref{apdx:case_faithfulness}
, \ref{apdx:case_accessibility}
% , \ref{apdx:case_robustness}
), and also some comparison examples of benefit and tax within the same memory mechanism to directly show the dual nature of memory (\ref{apdx:case_dual}).

\definecolor{CaseFrame}{HTML}{6B7280}
\definecolor{RawFrame}{HTML}{8F8F8F}
\definecolor{RawBg}{HTML}{F1F1F1}
\definecolor{MemFrame}{HTML}{5E877A}
\definecolor{MemBg}{HTML}{F2F6F3}
\definecolor{MemCard}{HTML}{B3C9B6}
\definecolor{KeyHL}{HTML}{E6E1C8}
\definecolor{OkGreen}{HTML}{4E7D6B}
\definecolor{BadRed}{HTML}{9C7A3C}

\newtcolorbox{casebox}[1]{enhanced, colback=white, colframe=CaseFrame, boxrule=0.6pt,
  arc=4pt, left=5pt, right=5pt, top=4pt, bottom=4pt, width=\linewidth, before skip=0pt, after skip=0pt,
  title={\textbf{#1}}, fonttitle=\small\bfseries, coltitle=white, colbacktitle=CaseFrame}
% same card without a title bar (D.2)
\newtcolorbox{caseplain}[1][]{enhanced, colback=white, colframe=CaseFrame, boxrule=0.6pt,
  arc=4pt, left=5pt, right=5pt, top=4pt, bottom=4pt, width=\linewidth, before skip=0pt, after skip=0pt, #1}
% question / gold answer line
\newcommand{\caseq}[2]{{\small\textbf{Question:} #1}\par\vspace{4pt}}   % #2 (gold answer) kept in the source, not shown
% raw side (left) and memory side (right): a plain coloured header, then the items
\newenvironment{rawpanel}[1][Raw dialogue]{{\scriptsize\bfseries\color{RawFrame}#1}\par\vspace{1pt}}{}
\newenvironment{mempanel}[1]{{\scriptsize\bfseries\color{MemFrame}#1}\par\vspace{1pt}}{}
% chat bubble for a raw turn: \bubble{speaker icon + date}{text}  (filled, no frame)
\newcommand{\bubble}[2]{\begin{tcolorbox}[colback=RawBg, colframe=RawBg, boxrule=0pt, arc=5pt,
  left=4pt, right=4pt, top=2pt, bottom=2pt, before skip=2pt, after skip=2pt]
  {\tiny\color{black!60}#1}\\[-1pt]{\scriptsize\raggedright #2\par}\end{tcolorbox}}
% memory unit: filled bubble like \bubble, no frame; \otherunit = a non-gold unit (grey bar, dimmed)
\newcommand{\memunit}[2]{\begin{tcolorbox}[colback=MemBg, colframe=MemBg, boxrule=0pt, arc=5pt,
  left=4pt, right=4pt, top=2pt, bottom=2pt, before skip=2pt, after skip=2pt]
  {\scriptsize\raggedright #2\par}\end{tcolorbox}}   % #1 (unit type) kept in the source, not shown
\newcommand{\otherunit}[2]{\begin{tcolorbox}[enhanced, colback=black!3, colframe=black!25, boxrule=0pt, leftrule=1.6pt,
  sharp corners, left=4pt, right=4pt, top=2pt, bottom=2pt, before skip=2pt, after skip=2pt]
  {\tiny\color{black!45}#1}\\[-1pt]{\scriptsize\raggedright\color{black!60}#2\par}\end{tcolorbox}}
\newcommand{\key}[1]{{\setlength{\fboxsep}{0.6pt}\colorbox{KeyHL}{#1}}}
\newcommand{\lost}[1]{{\scriptsize\color{BadRed}\faExclamationTriangle\ \textit{#1}}}
% 10-11 user: D.1 / D.4 notes in dark grey text (icon stays BadRed)
\newcommand{\lostk}[1]{{\scriptsize\textcolor{BadRed}{\faExclamationTriangle}\ \color{black!75}\textit{#1}}}
\newcommand{\caseok}{\textcolor{OkGreen}{\faCheck}}
\newcommand{\casebad}{\textcolor{BadRed}{\faTimes}}
% answers: \caseans{method}{answer}{\caseok|\casebad}
\newcommand{\caseans}[3]{{\small #3\ \textbf{#1:} #2}\par}
\newcommand{\casesides}[2]{\vspace{2pt}\noindent
  \begin{minipage}[t]{0.485\linewidth}\vspace{0pt}#1\end{minipage}\hfill
  \begin{minipage}[t]{0.485\linewidth}\vspace{0pt}#2\end{minipage}\par\vspace{4pt}}

% faithfulness level badges and one row per system (D.2)
\definecolor{LvNoLoss}{HTML}{4E7D6B}
\definecolor{LvPartial}{HTML}{B8913A}
\definecolor{LvMajor}{HTML}{B5534B}
\definecolor{LvNoMem}{HTML}{6B7280}
\newcommand{\lvbadge}[2]{{\setlength{\fboxsep}{1.5pt}\colorbox{#1}{\makebox[4.6em]{\tiny\bfseries\color{white}#2}}}}
\newcommand{\lvnoloss}{\lvbadge{LvNoLoss}{NO LOSS}}
\newcommand{\lvpartial}{\lvbadge{LvPartial}{PARTIAL LOSS}}
\newcommand{\lvmajor}{\lvbadge{LvMajor}{MAJOR LOSS}}
\newcommand{\lvnomem}{\lvbadge{LvNoMem}{NO MEM UNIT}}
% \sysrow{badge}{system}{memory unit(s)}: badge + system name on one line, the units below at full width
\newcommand{\sysrow}[3]{\noindent#1\ {\scriptsize\bfseries #2}\par\vspace{1pt}#3\par\vspace{4pt}}
\newcommand{\nounit}[1]{{\scriptsize\color{black!55}\textit{#1}}\par}

\subsection{Memory Benefit}
\label{apdx:case_mem_benefit}

\newtcolorbox{benbox}[2][]{enhanced, colback=white, colframe=CaseFrame, boxrule=0.6pt,
  arc=4pt, left=5pt, right=5pt, top=3pt, bottom=4pt, width=\linewidth, before skip=0pt, after skip=0pt,
  title={\benfit{#2}}, fonttitle=\small\bfseries, coltitle=white, colbacktitle=CaseFrame, toptitle=1pt, bottomtitle=1pt, #1}   % 10-11 user: larger title
% 10-11 user: one-line card titles, shrunk only when longer than the card
\newcommand{\benfit}[1]{\sbox0{#1}\ifdim\wd0>\linewidth\resizebox{\linewidth}{!}{\usebox0}\else\usebox0\fi}
% 10-11 user: frame colours = Sec 4 BenefitGreen / TaxRed
\colorlet{BenefitFrame}{BenefitGreen!60!CaseFrame}   % 10-11 user: less saturated than the text colour
\colorlet{TaxFrame}{TaxRed!60!CaseFrame}
\tcbset{benefit/.style={colframe=BenefitFrame, colbacktitle=BenefitFrame}, tax/.style={colframe=TaxFrame, colbacktitle=TaxFrame}}
% 10-12 user: D.1 pairs share one card (one title); left / right case split by a dashed vertical line (\tcblower)
\tcbset{benpair/.style={sidebyside, sidebyside align=top seam, sidebyside gap=12pt, lower separated=true,
  segmentation style={dashed, draw=tcbcolframe, line width=0.5pt}}}
\newcommand{\benlost}[1]{{\raggedright\scriptsize\lostk{#1}\par}}   % 10-13: \scriptsize outside, so wrapped notes keep tight lines
\newcommand{\bennote}[1]{{\raggedright\scriptsize\color{OkGreen}\faCheck\ \textit{#1}\par}}
\newcommand{\benbad}{\textcolor{LvMajor}{\faTimes}}   % 10-11 user: red cross in D.5
% 10-11 user: retrieved RAG chunk without the "retrieved chunk" header
\newcommand{\benchunk}[1]{\begin{tcolorbox}[colback=RawBg, colframe=RawBg, boxrule=0pt, arc=5pt,
  left=4pt, right=4pt, top=2pt, bottom=2pt, before skip=2pt, after skip=2pt]{\scriptsize\raggedright #1\par}\end{tcolorbox}}
\newcommand{\benq}[2]{{\small\textbf{Question:} #1}\par\vspace{1pt}{\small\textbf{Gold:} #2}\par\vspace{3pt}}

Even the main results show memory underperform RAG in the overall performance in many cases, memory systems can still answer some questions correctly where naive RAG fails and we still need them in such scenarios. We inspect a sample of such questions and group the wins of memory into several representative types (Figures~\ref{fig:case_benefit_1} and \ref{fig:case_benefit_2}):

\catsq~\textbf{Focused atomic extraction}: each fact becomes a short unit, so the gold facts fit into the retrieved context, while RAG's top-$k$ is filled by long chunks that other irrelevant information, making the evidence hard to recover. 

\catsq~\textbf{Multi-hop linking}: facts from different sessions are retrieved together and combined, while for naive RAG the evidences are scattered. 

\catsq~\textbf{Implicit preference exposure}: a trait mentioned only in passing becomes an explicit unit in the memory. 

\catsq~\textbf{Temporal anchoring}: relative time expressions are anchored to absolute dates during construction. 

\catsq~\textbf{Explicit forgetting}: a forget request updates the stored state by removing or deactivating the targeted content.

\catsq~\textbf{Processing for heterogeneous inputs}: non-dialogue inputs such as code diffs are described in natural language, making it easier to be recalled with semantic retrieval.  

\begin{figure*}[p]
\centering
\begin{benbox}[benefit, benpair]{Memory Benefit \textperiodcentered\ Focused Atomic Extraction \textperiodcentered\ LME-s}
\benq{How many properties did I view before making an offer on the townhouse in the Brookside neighborhood?}{4}
\caseans{RAG (1rd)}{1}{\benbad}
\benchunk{I actually had a home inspection done on the 3-bedroom \key{townhouse} I'm buying in the \key{Brookside} neighborhood \ldots}
\benlost{every retrieved chunk is about the Brookside townhouse}
\vspace{3pt}
\caseans{\textsc{Memory-R2}}{4}{\caseok}
\memunit{}{User recently saw a beautiful 3-bedroom \key{bungalow} in the Oakwood neighborhood on January 22nd \ldots}
\memunit{}{User has seen some properties that didn't fit their budget, like one in \key{Cedar Creek} on February 1st}
\memunit{}{User viewed a \key{1-bedroom condo} on February 10th but the noise from the highway was a deal-breaker}
\memunit{}{User's offer on the \key{2-bedroom condo} got rejected on the 17th due to a higher bid}
\tcblower
\benq{What new kitchen gadget did I invest in before getting the Air Fryer?}{Instant Pot}
\caseans{RAG (1rd)}{not in the memories}{\benbad}
\benchunk{What a great idea! \key{Air Fryer} sweet potato fries are a fantastic side dish \ldots}
\benlost{top chunks are long assistant replies about the Air Fryer}
\vspace{3pt}
\caseans{\textsc{SimpleMem}}{Instant Pot}{\caseok}
\memunit{}{The user stated an intention to use a new \key{Instant Pot} to make soups and stews \ldots\ Time: 2023-05-21T05:48:00}
\caseans{\textsc{LightMem}}{Instant Pot}{\caseok}
\memunit{}{2023-05-21T05:48:01.000 Sun The user is thinking of using a new \key{Instant Pot} to make soups and stews.}
\end{benbox}

\vspace{8pt}
\begin{benbox}[benefit, benpair]{Memory Benefit \textperiodcentered\ Multi-hop Linking \textperiodcentered\ LME-s}
\benq{How many years will I be when my friend Rachel gets married? (asked on 2022/09/01)}{33}
\caseans{RAG (1rd)}{information not available}{\benbad}
\bubble{user \textperiodcentered\ 2022/09/01}{\ldots\ By the way, my friend \key{Rachel's getting married next year} \ldots}
\benlost{the turn with the user's age is not retrieved}
\vspace{3pt}
\caseans{\textsc{MemoryOS}}{33}{\caseok}
\memunit{}{Friend's Event: Friend named Rachel is getting married \key{next year} (relative to the conversation date of 2022/09/01).}
\memunit{}{Age: \key{32 years old}.}
\tcblower
\benq{What is the average age of me, my parents, and my grandparents?}{59.6}
\caseans{RAG (1rd)}{information not available}{\benbad}
\benchunk{\ldots\ \key{my mom is 55 and my dad is 58} \ldots}
\benlost{the grandparents' turn is not retrieved}
\vspace{3pt}
\caseans{\textsc{PlugMem}}{59.6}{\caseok}
\memunit{}{(2023/05/27) I'm \key{32} now}
\memunit{}{User's mother is \key{55} years old and their father is \key{58} years old.}
\memunit{}{(2023/05/22) My grandma is \key{75} and my grandpa is \key{78}}
\end{benbox}

\vspace{8pt}
\begin{benbox}[benefit, benpair]{Memory Benefit \textperiodcentered\ Implicit Preference Exposure \textperiodcentered\ PM-v2}
\benq{What are some cozy, hands-on weekend projects I could try that are relaxing and satisfying?}{a suggestion that builds on the user's weekend bread baking}
\caseans{RAG (1rd)}{a generic suggestion}{\benbad}
\benchunk{You could curl up with a cup of tea and work on a new scarf pattern \ldots}
\benlost{baking is mentioned only in an email about rescheduling a meeting}
\vspace{3pt}
\caseans{\textsc{MemOS}}{the bread-baking option}{\caseok}
\memunit{}{On February 8, 2025, Miguel mentioned that his recent weekend \ldots\ included balancing spreadsheets, spending time with family, and \key{kneading dough on a quiet Sunday afternoon as a personal ritual}.}
\tcblower
\benq{What are some good tips for making long hikes more comfortable?}{tips that account for the user's old knee injury}
\caseans{RAG (1rd)}{generic hiking tips}{\benbad}
\benchunk{If local pollen sets off your allergies, try wearing wraparound sunglasses and a wide-brimmed hat \ldots}
\benlost{the knee injury is mentioned only inside a translation request}
\vspace{3pt}
\caseans{\textsc{MemOS}}{the knee-injury option}{\caseok}
\memunit{}{The user recalled \key{twisting their knee} on a steep section of the trail years ago. As a result of this injury, they now \key{prefer taking the gentler path} \ldots}
\end{benbox}
\caption{Examples of memory benefit: (a) focused atomic extraction, (b) multi-hop linking, and (c) implicit preference exposure.}
\label{fig:case_benefit_1}
\end{figure*}

\begin{figure*}[p]
\centering
\begin{benbox}[benefit, benpair]{Memory Benefit \textperiodcentered\ Temporal Anchoring \textperiodcentered\ LME-s}
\benq{Which task did I complete first, fixing the fence or trimming the goats' hooves?}{fixing the fence}
\caseans{RAG (1rd)}{trimming the goats' hooves}{\benbad}
\bubble{user \textperiodcentered\ 2023/05/25}{I just fixed that broken fence on the east side of my property \key{three weeks ago} \ldots}
\bubble{user \textperiodcentered\ 2023/05/25}{\ldots\ The last time I did it was about \key{two weeks ago} \ldots}
\vspace{3pt}
\caseans{\textsc{SimpleMem}}{fixing the fence}{\caseok}
\memunit{}{The user fixed a broken fence on the east side of the user's property \key{three weeks prior to 2023-05-25}. Time: 2023-05-04}
\memunit{}{The user successfully trimmed the goats' hooves \key{two weeks before 2023-05-25} \ldots}
\tcblower
\benq{Which device did I set up first, the smart thermostat or the mesh network system?}{smart thermostat}
\caseans{RAG (1rd)}{the mesh network system}{\benbad}
\benchunk{\ldots\ since I \key{just upgraded} to a mesh network system \ldots}
\benlost{the turn with ``3 weeks ago'' is not retrieved}
\vspace{3pt}
\caseans{\textsc{OpenViking}}{the smart thermostat}{\caseok}
\memunit{}{events/2023/05/25/mesh\_network\_upgrade.md: The user upgraded their home Wi-Fi router to a new mesh network system 3 weeks ago \key{(around 2023-05-06)} \ldots}
\memunit{}{events/2023/05/25/smart\_thermostat\_setup.md: The user set up a new smart thermostat approximately one month prior to 2023-05-25 \key{(around 2023-04-25)}.}
\end{benbox}

\vspace{5pt}
\begin{benbox}[benefit, benpair]{Memory Benefit \textperiodcentered\ Explicit Forgetting \textperiodcentered\ PM-v2}
\benq{Do you have any recommendations for fascinating and informative science-themed films or series that would make a long flight feel shorter and more engaging?}{a non-science series (\textit{Ted Lasso}), as the user asked to forget their interest in science documentaries}
\caseans{RAG (1rd)}{\textit{Cosmos: Possible Worlds}, ``since you enjoy science documentaries''}{\benbad}
\benchunk{Since you enjoy science documentaries, you might love `Cosmos: A Spacetime Odyssey' \ldots\ [user]: \key{Please forget} that I enjoy watching science documentaries.}
\benlost{the request is retrieved but next to the reply it retracts}
\vspace{3pt}
\caseans{\textsc{MAGMA}}{\textit{Ted Lasso}}{\caseok}
\memunit{}{The user \key{does not want} recommendations for science documentaries.}
\tcblower
\benq{What are some effective ways to discover lesser-known book fairs or estate sales when travelling to a new city?}{a general tip, as the user asked to forget that they collect rare books}
\caseans{RAG (1rd)}{asking bookshop owners, for ``fellow collectors of rare volumes''}{\benbad}
\benchunk{When hunting for \key{rare books} in small, local shops, check the less obvious shelves \ldots\ [user]: \key{Please forget} that I collect rare books from second-hand shops.}
\vspace{3pt}
\caseans{\textsc{Mem-T}}{the general tip}{\caseok}
\memunit{}{(experience) To find unique items in small, local shops: visit on weekdays or after new arrivals; check less obvious shelves and tucked-away corners; ask the shop owner about special or off-display items.}
\bennote{the stored advice no longer mentions rare books}
\end{benbox}

\vspace{5pt}
\begin{benbox}[benefit, benpair]{Memory Benefit \textperiodcentered\ Processing for Heterogeneous Inputs \textperiodcentered\ MINT}
\benq{At the commit immediately after the \texttt{tabulate} dependency was first added to the project's \texttt{requirements.txt}, what version of \texttt{spotipy} was pinned in that same file?}{2.12.0}
\caseans{RAG (1rd)}{information not available}{\benbad}
\benchunk{\texttt{[step 50] (2021-02-12) diff {-}{-}git a/requirements.txt \ldots\ -spotipy==2.12.0 +spotipy==2.16.1}}
\benlost{the raw commit that adds \texttt{tabulate} is not retrieved}
\vspace{3pt}
\caseans{\textsc{MemOS}}{2.12.0}{\caseok}
\memunit{}{On January 31, 2021, the user \key{added the `tabulate' library} (version 0.8.7) to the project's requirements.txt file.}
\memunit{}{On February 12, 2021, the user updated the `spotipy' library version in the `requirements.txt' file \key{from 2.12.0 to 2.16.1}.}
\caseans{\textsc{Memory-R2}}{2.12.0}{\caseok}
\memunit{}{2021-01-31T20:23:10Z --- User: User \key{added tabulate==0.8.7} to requirements.txt}
\tcblower
\benq{For how many days was the file \texttt{modelci/hub/deployer/pytorch/docker-env.env} present in the repository before it was removed?}{2 days}
\caseans{RAG (1rd)}{28 days}{\benbad}
\benchunk{\texttt{\ldots\ diff {-}{-}git a/modelci/hub/deployer/pytorch/docker-env.env \ldots\ deleted file \ldots\ -CUDA\_DEVICE\_ORDER=PCI\_BUS\_ID \ldots}}
\benlost{the deletion is buried in a multi-file diff, the creation is not retrieved}
\vspace{3pt}
\caseans{\textsc{SimpleMem}}{2 days}{\caseok}
\memunit{}{At \key{2021-05-26T03:48:51Z}, a new file modelci/hub/deployer/pytorch/docker-env.env \key{was created} containing CUDA environment variables \ldots}
\memunit{}{At \key{2021-05-28T07:49:45Z}, the file modelci/hub/deployer/pytorch/docker-env.env \key{was deleted} \ldots}
\end{benbox}
\caption{Examples of memory benefit-2: (d) temporal anchoring, (e) explicit forgetting, and (f) processing for heterogeneous inputs.}
\label{fig:case_benefit_2}
\end{figure*}

\subsection{Memory Tax - Faithfulness}
\label{apdx:case_faithfulness}

Figures~\ref{fig:case_faith_lme} to \ref{fig:case_faith_lme3} show examples from the annotation in Section~\ref{sec:faithfulness}, with varying degrees of information loss: 

\catsq~\textbf{No Loss}. 

\catsq~\textbf{Partial Loss}. 

\catsq~\textbf{Major Loss}. 

\catsq~\textbf{No Corresponding Memory Unit}.

\begin{figure*}[p]
\centering
\begin{minipage}[t]{0.485\textwidth}
\vspace{0pt}
\begin{benbox}[tax, equal height group=faithA]{Memory Tax \textperiodcentered\ Faithfulness}
\caseq{How much will I save by taking the train from the airport to my hotel instead of a taxi?}{\$50}
\begin{rawpanel}
\bubble{user \textperiodcentered\ 2023/05/26 (Fri) 22:35}{By the way, I was told that taking a taxi from the airport to my hotel would cost \key{around \$60}, which is a bit pricey for me.}
\end{rawpanel}
\vspace{3pt}
\sysrow{\lvnoloss}{\textsc{LightMem}}{\memunit{memory entry}{The user was told that taking a taxi from the airport to their hotel would cost \key{around \$60}.}}
\sysrow{\lvnoloss}{\textsc{MemOS}}{\memunit{memory node}{On May 26, 2023, the user expressed concern that taking a taxi from the airport to their hotel in Tokyo would cost \key{around \$60}, which they considered too expensive.}}
\sysrow{\lvmajor}{\textsc{OpenViking}}{\memunit{preferences/user/travel\_transportation.md}{Prefers cost-effective transportation options due to budget concerns. Considers convenience and luggage handling when choosing transport. Open to asking hotels for transportation recommendations.}\lostk{dropped: the \$60 taxi fare}}
\sysrow{\lvnomem}{\textsc{Memory-R2}}{}
\end{benbox}
\caption{Faithfulness case studies-1.}
\label{fig:case_faith_lme}
\end{minipage}\hfill
\begin{minipage}[t]{0.485\textwidth}
\vspace{0pt}
\begin{benbox}[tax, equal height group=faithA]{Memory Tax \textperiodcentered\ Faithfulness}
\caseq{How can I gracefully turn down frequent family gatherings where the conversation always shifts to my personal life, without damaging relationships or losing precious time for my creative work?}{a reply that does not bring up the detail the user asked to forget}
\begin{rawpanel}
\bubble{user}{Please \key{forget the detail} about me feeling pressure from family to marry conflicting with my desire to focus on art and career.}
\end{rawpanel}
\vspace{3pt}
\sysrow{\lvnoloss}{\textsc{SimpleMem}}{\memunit{memory entry}{The user requested to \key{forget the detail} about feeling pressure from family to marry conflicting with a desire to focus on art and career.}}
\sysrow{\lvpartial}{\textsc{MAGMA}}{\memunit{event node \textperiodcentered\ summary}{The user instructs the assistant to \key{disregard previous details} about family pressure conflicting with their desire to pursue art and a career.}\memunit{event node \textperiodcentered\ semantic facts}{The user is prioritizing art and career over marriage. The user is ignoring family pressure related to marriage.}\lostk{the request to forget is turned into new facts about the same detail}}
\sysrow{\lvmajor}{\textsc{PlugMem}}{\memunit{semantic memory}{User mentioned feeling pressure from family to marry, which conflicts with their desire to focus on art and career.}\lostk{dropped: the request to forget, so the store keeps exactly the detail to be forgotten}}
\sysrow{\lvnomem}{\textsc{OpenViking}}{}
\end{benbox}
\caption{Faithfulness case studies-2.}
\label{fig:case_faith_psn}
\end{minipage}

\vspace{10pt}
\begin{minipage}[t]{0.485\textwidth}
\vspace{0pt}
\begin{benbox}[tax, equal height group=faithB]{Memory Tax \textperiodcentered\ Faithfulness}
\caseq{How long did Alex marinate the BBQ ribs in special sauce?}{24 hours}
\begin{rawpanel}
\bubble{user \textperiodcentered\ 2023/05/26 (Fri) 10:26}{\ldots\ By the way, I was thinking of trying out some BBQ ribs for my party, and I remember Alex telling me he marinated them \key{in a special sauce} for \key{24 hours} before grilling them to perfection. Do you have any tips on how to choose the right type of wood for smoking meats?}
\end{rawpanel}
\vspace{3pt}
\sysrow{\lvnoloss}{\textsc{PlugMem}}{\memunit{semantic memory}{User knows someone named Alex who marinates ribs \key{in a special sauce} for \key{24 hours} before grilling.}}
\sysrow{\lvpartial}{\textsc{Memory-R2}}{\memunit{memory item}{User considers trying out some BBQ ribs for the party and remembers Alex's advice about marinating them for \key{24 hours}}\lostk{dropped: ``in a special sauce'', and what Alex did becomes advice}}
\end{benbox}
\caption{Faithfulness case studies-3.}
\label{fig:case_faith_lme2}
\end{minipage}\hfill
\begin{minipage}[t]{0.485\textwidth}
\vspace{0pt}
\begin{benbox}[tax, equal height group=faithB]{Memory Tax \textperiodcentered\ Faithfulness}
\caseq{How many days did it take for my laptop backpack to arrive after I bought it?}{5 days}
\begin{rawpanel}
\bubble{user \textperiodcentered\ 2023/01/24 (Tue) 09:37}{\ldots\ By the way, I just started using my new laptop backpack and it's been great. It \key{arrived on 1/20} and has been a lifesaver for carrying my work laptop back and forth.}
\end{rawpanel}
\vspace{3pt}
\sysrow{\lvnoloss}{\textsc{MAGMA}}{\memunit{event node \textperiodcentered\ summary}{The user requests recommendations for a new wireless mouse due to issues with their current one and mentions that their new laptop backpack, which \key{arrived on 1/20}, is performing well.}}
\sysrow{\lvpartial}{\textsc{SimpleMem}}{\memunit{memory entry}{The user \key{started using} a new laptop backpack \key{on 2023-01-20}, which has been effective for carrying a work laptop.}\lostk{the arrival date becomes the date the user started using it}}
\end{benbox}
\caption{Faithfulness case studies-4.}
\label{fig:case_faith_lme3}
\end{minipage}
\end{figure*}

\newcommand{\rdbadge}[2]{{\setlength{\fboxsep}{1.5pt}\colorbox{#1}{\makebox[5.6em]{\tiny\bfseries\color{white}#2}}}}
\newcommand{\rdok}[2]{\rdbadge{LvNoLoss}{ROUND #1 / #2}}
\newcommand{\rdbad}[2]{\rdbadge{LvMajor}{ROUND #1 / #2}}

\subsection{Memory Tax - Accessibility}
\label{apdx:case_accessibility}

% 10-13 user: D.3 redone with three accessibility patterns (fragmenting / diluting / rephrasing), two cases each, Q27B.
% Source: Rethink_Cost/rethink_user_mem/analysis/2_accessibility/cases_v2/ (10-13 user: six cases, five systems):
% F1 LightMem gpt4_d84a3211, F2 PlugMem 06f04340, D1 OpenViking 118b2229, D5 MemOS persona256#4, MAGMA 59524333, R1 LightMem persona101#10 (10-13 user: no second PlugMem).
% Unit texts verbatim, trims marked; MAGMA replays 4 rounds (beyond = ROUND > 4).
\newcommand{\sysname}[1]{{\scriptsize\bfseries #1}\par\vspace{1pt}}
\newcommand{\rdout}{\rdbadge{LvMajor}{ROUND > 4}}   % MAGMA replays only 4 rounds
\newcommand{\rdunit}[2]{\noindent#1\par\vspace{-1pt}\memunit{}{#2}}
\newcommand{\rdlost}[1]{{\raggedright\scriptsize\color{BadRed}\faExclamationTriangle\ \textit{#1}\par}}   % \lost with its own paragraph, so wrapped notes keep tight lines

\begin{figure*}[p]
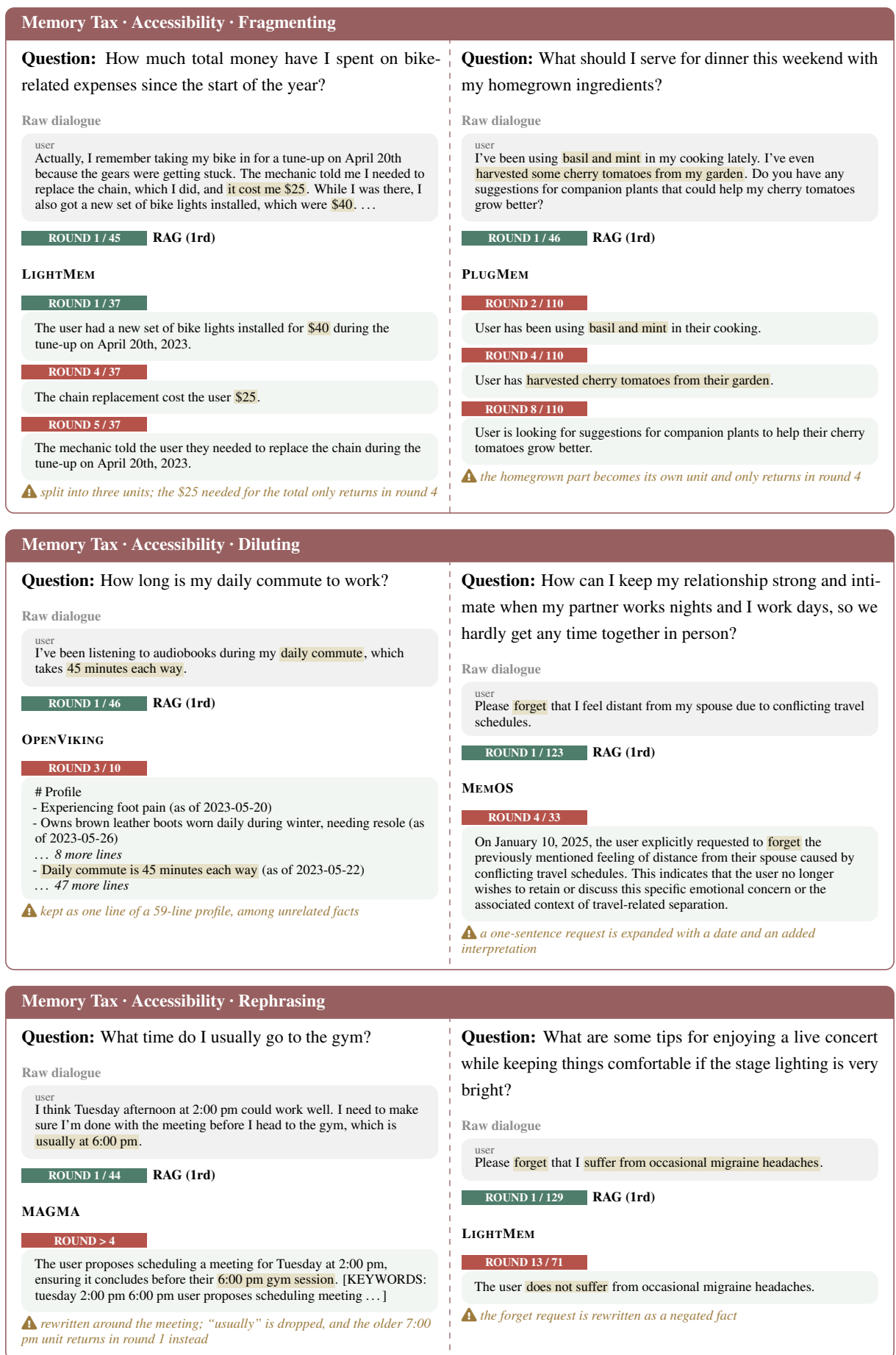

\centering
\begin{benbox}[tax, benpair]{Memory Tax \textperiodcentered\ Accessibility \textperiodcentered\ Fragmenting}
\caseq{How much total money have I spent on bike-related expenses since the start of the year?}{\$185}
\begin{rawpanel}
\bubble{user}{Actually, I remember taking my bike in for a tune-up on April 20th because the gears were getting stuck. The mechanic told me I needed to replace the chain, which I did, and \key{it cost me \$25}. While I was there, I also got a new set of bike lights installed, which were \key{\$40}. \ldots}
\end{rawpanel}
\vspace{3pt}
\sysrow{\rdok{1}{45}}{RAG (1rd)}{}
\sysname{\textsc{LightMem}}
\rdunit{\rdok{1}{37}}{The user had a new set of bike lights installed for \key{\$40} during the tune-up on April 20th, 2023.}
\rdunit{\rdbad{4}{37}}{The chain replacement cost the user \key{\$25}.}
\rdunit{\rdbad{5}{37}}{The mechanic told the user they needed to replace the chain during the tune-up on April 20th, 2023.}
\rdlost{split into three units; the \$25 needed for the total only returns in round 4}
\tcblower
\caseq{What should I serve for dinner this weekend with my homegrown ingredients?}{dinner ideas built on the user's homegrown cherry tomatoes, basil and mint}
\begin{rawpanel}
\bubble{user}{I've been using \key{basil and mint} in my cooking lately. I've even \key{harvested some cherry tomatoes from my garden}. Do you have any suggestions for companion plants that could help my cherry tomatoes grow better?}
\end{rawpanel}
\vspace{3pt}
\sysrow{\rdok{1}{46}}{RAG (1rd)}{}
\sysname{\textsc{PlugMem}}
\rdunit{\rdbad{2}{110}}{User has been using \key{basil and mint} in their cooking.}
\rdunit{\rdbad{4}{110}}{User has \key{harvested cherry tomatoes from their garden}.}
\rdunit{\rdbad{8}{110}}{User is looking for suggestions for companion plants to help their cherry tomatoes grow better.}
\rdlost{the homegrown part becomes its own unit and only returns in round 4}
\end{benbox}

\vspace{8pt}
\begin{benbox}[tax, benpair]{Memory Tax \textperiodcentered\ Accessibility \textperiodcentered\ Diluting}
\caseq{How long is my daily commute to work?}{45 minutes each way}
\begin{rawpanel}
\bubble{user}{I've been listening to audiobooks during my \key{daily commute}, which takes \key{45 minutes each way}.}
\end{rawpanel}
\vspace{3pt}
\sysrow{\rdok{1}{46}}{RAG (1rd)}{}
\sysname{\textsc{OpenViking}}
\rdunit{\rdbad{3}{10}}{\# Profile\\- Experiencing foot pain (as of 2023-05-20)\\- Owns brown leather boots worn daily during winter, needing resole (as of 2023-05-26)\\\textit{\ldots\ 8 more lines}\\- \key{Daily commute is 45 minutes each way} (as of 2023-05-22)\\\textit{\ldots\ 47 more lines}}
\rdlost{kept as one line of a 59-line profile, among unrelated facts}
\tcblower
\caseq{How can I keep my relationship strong and intimate when my partner works nights and I work days, so we hardly get any time together in person?}{a reply that does not bring up the detail the user asked to forget}
\begin{rawpanel}
\bubble{user}{Please \key{forget} that I feel distant from my spouse due to conflicting travel schedules.}
\end{rawpanel}
\vspace{3pt}
\sysrow{\rdok{1}{123}}{RAG (1rd)}{}
\sysname{\textsc{MemOS}}
\rdunit{\rdbad{4}{33}}{On January 10, 2025, the user explicitly requested to \key{forget} the previously mentioned feeling of distance from their spouse caused by conflicting travel schedules. This indicates that the user no longer wishes to retain or discuss this specific emotional concern or the associated context of travel-related separation.}
\rdlost{a one-sentence request is expanded with a date and an added interpretation}
\end{benbox}

\vspace{8pt}
\begin{benbox}[tax, benpair]{Memory Tax \textperiodcentered\ Accessibility \textperiodcentered\ Rephrasing}
\caseq{What time do I usually go to the gym?}{6:00 pm}
\begin{rawpanel}
\bubble{user}{I think Tuesday afternoon at 2:00 pm could work well. I need to make sure I'm done with the meeting before I head to the gym, which is \key{usually at 6:00 pm}.}
\end{rawpanel}
\vspace{3pt}
\sysrow{\rdok{1}{44}}{RAG (1rd)}{}
\sysname{\textsc{MAGMA}}
\rdunit{\rdout}{The user proposes scheduling a meeting for Tuesday at 2:00 pm, ensuring it concludes before their \key{6:00 pm gym session}. [KEYWORDS: tuesday 2:00 pm 6:00 pm user proposes scheduling meeting \ldots]}
\rdlost{rewritten around the meeting; ``usually'' is dropped, and the older 7:00 pm unit returns in round 1 instead}
\tcblower
\caseq{What are some tips for enjoying a live concert while keeping things comfortable if the stage lighting is very bright?}{a reply that does not bring up the detail the user asked to forget}
\begin{rawpanel}
\bubble{user}{Please \key{forget} that I \key{suffer from occasional migraine headaches}.}
\end{rawpanel}
\vspace{3pt}
\sysrow{\rdok{1}{129}}{RAG (1rd)}{}
\sysname{\textsc{LightMem}}
\rdunit{\rdbad{13}{71}}{The user \key{does not suffer} from occasional migraine headaches.}
\rdlost{the forget request is rewritten as a negated fact}
\end{benbox}
\caption{Examples of accessibility degradation: (a) fragmenting, (b) diluting, and (c) rephrasing.}
\label{fig:case_access}
\end{figure*}

Figure~\ref{fig:case_access} shows examples from the round replay in Section~\ref{sec:accessibility}. In each case the evidence is still preserved in memory, but it is returned in a later round than the raw turns. The badges give the round in which a unit is first retrieved and the maximum number of rounds for that store. We observe three recurring patterns:

\catsq~\textbf{Fragmenting}: one raw turn is split into several units, and a unit carrying a necessary part of the evidence ranks far behind the others.

\catsq~\textbf{Diluting}: the evidence is wrapped in unrelated or additional content, such as a long profile or an elaborated note, which weakens its match with the query.

\catsq~\textbf{Rephrasing}: the evidence is rewritten, sometimes with a slight shift in meaning, and no longer matches the query as well as the raw wording.

\subsection{Direct Examples for the Double-edged Memory Mechanisms}
\label{apdx:case_dual}

% 10-11 user: same memory system + same mechanism, one question helped (mem right, RAG (1rd) wrong) and one hurt
% (mem wrong, RAG (1rd) right); Qwen3.6-27B. Cases: Rethink_Cost/rethink_user_mem/analysis/mem_benefit/labels_dual/*.json
\newtcolorbox{dualbox}[1]{enhanced, colback=white, colframe=CaseFrame, boxrule=0.6pt, arc=4pt,
  left=5pt, right=5pt, top=3pt, bottom=4pt, width=\textwidth, before skip=0pt, after skip=0pt,
  title={#1}, fonttitle=\small\bfseries, coltitle=white, colbacktitle=CaseFrame, toptitle=1pt, bottomtitle=1pt,
  % 10-13 user: frame and title bar shaded left to right, benefit green -> tax red (the frame fill shows through the title)
  frame code={\path[left color=BenefitFrame, right color=TaxFrame, rounded corners=4pt] (frame.south west) rectangle (frame.north east);},
  title code={}}
\newcommand{\dualhead}[2]{{\small\bfseries\color{#1}#2}\par\vspace{2pt}}
% \dualpair{benefit side}{tax side}: two halves with a thin rule between them
\newcommand{\dualpair}[2]{\noindent
  \begin{minipage}[t]{0.485\linewidth}\vspace{0pt}#1\end{minipage}\hfill
  {\color{black!20}\vrule width 0.4pt}\hfill
  \begin{minipage}[t]{0.485\linewidth}\vspace{0pt}#2\end{minipage}}

The same structure or mechanism of a memory system can help on one question and hurt on another. Figures~\ref{fig:case_dual_1} and \ref{fig:case_dual_2} pair, for each mechanism, a question that the memory system answers correctly while RAG (1rd) fails (\textcolor{BenefitGreen}{memory benefit}) with a question where the same mechanism makes it fail while RAG (1rd) succeeds (\textcolor{TaxRed}{memory tax}).

\begin{figure*}[p]
\centering
\begin{dualbox}{Focused Atomic Extraction \textperiodcentered\ \textsc{Memory-R2} \textperiodcentered\ LME-s}
\dualpair{%
\dualhead{BenefitGreen}{Memory Benefit}
\benq{How many properties did I view before making an offer on the townhouse in the Brookside neighborhood?}{4}
\caseans{RAG (1rd)}{1}{\benbad}
\benchunk{I actually had a home inspection done on the 3-bedroom \key{townhouse} I'm buying in the \key{Brookside} neighborhood \ldots}
\caseans{\textsc{Memory-R2}}{4}{\caseok}
\memunit{}{User recently saw a beautiful 3-bedroom \key{bungalow} in the Oakwood neighborhood on January 22nd \ldots}
\memunit{}{User viewed a \key{1-bedroom condo} on February 10th but the noise from the highway was a deal-breaker}
\bennote{each viewing becomes its own short fact, so all of them fit in the context}
}{%
\dualhead{TaxRed}{Memory Tax}
\benq{How much discount will I get on my next purchase at FreshMart?}{\$5}
\caseans{RAG (1rd)}{\$5}{\caseok}
\benchunk{I've been trying to shop at FreshMart more often since, you know, 500 points gets me a \key{\$5 discount} on my next purchase.}
\caseans{\textsc{Memory-R2}}{\$1}{\benbad}
\memunit{}{2023/05/27 (Sat) 06:17 --- User: User has been trying to shop at FreshMart more often due to the \key{500 points discount offer}}
\benlost{splitting the turn into one-line facts drops the amount ``\$5''}
}
\end{dualbox}

\vspace{6pt}
\begin{dualbox}{Multi-hop Linking \textperiodcentered\ \textsc{PlugMem} \textperiodcentered\ LME-s}
\dualpair{%
\dualhead{BenefitGreen}{Memory Benefit}
\benq{What is the average age of me, my parents, and my grandparents?}{59.6}
\caseans{RAG (1rd)}{information not available}{\benbad}
\benchunk{\ldots\ \key{my mom is 55 and my dad is 58} \ldots}
\caseans{\textsc{PlugMem}}{59.6}{\caseok}
\memunit{}{(reasoning note) the user states: ``I'm \key{32} now'' \ldots\ ``User's mother is \key{55} years old and their father is \key{58} years old.'' \ldots\ ``My grandma is \key{75} and my grandpa is \key{78}''}
\bennote{the note gathers facts from three sessions and computes the answer}
}{%
\dualhead{TaxRed}{Memory Tax}
\benq{How many weeks ago did I attend the `Summer Nights' festival at Universal Studios Hollywood? (asked on 2023/08/05)}{3 weeks ago}
\caseans{RAG (1rd)}{3 weeks ago}{\caseok}
\benchunk{\ldots\ I \key{just got back} from an amazing day at Universal Studios Hollywood - we had so much fun at the ``Summer Nights'' festival \ldots}
\caseans{\textsc{PlugMem}}{6 weeks ago}{\benbad}
\memunit{}{(reasoning note) If the visit was 3 weeks before July 15, and today is 3 weeks after July 15, the total time elapsed since the visit is \key{$3 + 3 = 6$ weeks}.}
\benlost{only the note's wrong conclusion reaches the QA model, not the raw turn}
}
\end{dualbox}

\vspace{6pt}
\begin{dualbox}{Implicit Preference Exposure \textperiodcentered\ \textsc{OpenViking} \textperiodcentered\ PM-v2 / LME-s}
\dualpair{%
\dualhead{BenefitGreen}{Memory Benefit \textperiodcentered\ PM-v2}
\benq{I'm planning a year-long self-study to go from a curious reader to a solid, working knowledge of two challenging, math-heavy fields \ldots}{a plan for astrophysics and quantum computing}
\caseans{RAG (1rd)}{a generic plan for two math-heavy fields}{\benbad}
\benchunk{Selena reads a book with 400 pages. Harry reads a book with 20 fewer than half the number of pages Selena's book has.}
\caseans{\textsc{OpenViking}}{the astrophysics and quantum computing plan}{\caseok}
\memunit{}{preferences/.../late\_night\_reading\_habits.md: Gets drawn into deep, stimulating articles about \key{astrophysics} (e.g., black holes) and \key{quantum computing} (e.g., qubits) late at night.}
\bennote{a passing remark becomes an explicit preference line}
}{%
\dualhead{TaxRed}{Memory Tax \textperiodcentered\ LME-s}
\benq{How much did I spend on each coffee mug for my coworkers?}{\$12}
\caseans{RAG (1rd)}{\$12}{\caseok}
\benchunk{By the way, speaking of gifts, I once spent \key{\$60 on some coffee mugs} for my coworkers, and it was a bit of a splurge, but they loved them.}
\caseans{\textsc{OpenViking}}{\$10--20}{\benbad}
\memunit{}{preferences/user/gift\_giving\_budget.md: Wants to create a budgeting plan to avoid overspending on gifts \ldots\ \key{Spends \$10-20 per coworker gift}}
\benlost{the concrete purchase is generalised into a budget range taken from the assistant's suggestion}
}
\end{dualbox}
\caption{Direct examples of the dual nature of memory-1: the same mechanism of a memory system brings memory benefit on one question (left) and memory tax on another (right).}
\label{fig:case_dual_1}
\end{figure*}

\begin{figure*}[p]
\centering
\begin{dualbox}{Temporal Anchoring \textperiodcentered\ \textsc{SimpleMem} \textperiodcentered\ LME-s}
\dualpair{%
\dualhead{BenefitGreen}{Memory Benefit}
\benq{Which task did I complete first, fixing the fence or trimming the goats' hooves?}{fixing the fence}
\caseans{RAG (1rd)}{trimming the goats' hooves}{\benbad}
\benchunk{I just fixed that broken fence on the east side of my property \key{three weeks ago} \ldots}
\caseans{\textsc{SimpleMem}}{fixing the fence}{\caseok}
\memunit{}{The user fixed a broken fence on the east side of the user's property \key{three weeks prior to 2023-05-25}.}
\memunit{}{The user successfully trimmed the goats' hooves \key{two weeks before 2023-05-25} \ldots}
\bennote{both relative times are anchored to the same date, so the order is explicit}
}{%
\dualhead{TaxRed}{Memory Tax}
\benq{How long did I use my new binoculars before I saw the American goldfinches returning to the area?}{two weeks}
\caseans{RAG (1rd)}{2 weeks}{\caseok}
\benchunk{I did manage to sneak in some birding time \key{a week ago} when I took a walk around my neighborhood after dinner. \ldots\ I did notice that the American goldfinches \ldots}
\caseans{\textsc{SimpleMem}}{3 weeks}{\benbad}
\memunit{}{The user noted \key{on 2023-05-20T22:12:00} that American goldfinches are returning to the area}
\memunit{}{The user reported taking a walk around their neighborhood after dinner \key{one week prior to 2023-05-20T22:12:00}}
\benlost{the sighting is split from its walk and stamped with the session time}
}
\end{dualbox}

\vspace{6pt}
\begin{dualbox}{Explicit Forgetting \textperiodcentered\ \textsc{MAGMA} \textperiodcentered\ PM-v2}
\dualpair{%
\dualhead{BenefitGreen}{Memory Benefit}
\benq{Do you have any recommendations for fascinating and informative science-themed films or series that would make a long flight feel shorter and more engaging?}{a non-science series (\textit{Ted Lasso}), as the user asked to forget their interest in science documentaries}
\caseans{RAG (1rd)}{\textit{Cosmos}, ``since you enjoy science documentaries''}{\benbad}
\benchunk{Since you enjoy science documentaries, you might love `Cosmos: A Spacetime Odyssey' \ldots\ [user]: \key{Please forget} that I enjoy watching science documentaries.}
\caseans{\textsc{MAGMA}}{\textit{Ted Lasso}}{\caseok}
\memunit{}{The user \key{does not want} recommendations for science documentaries.}
\bennote{the request becomes a negative preference}
}{%
\dualhead{TaxRed}{Memory Tax}
\benq{I'm planning to redecorate my study with an old-world travel theme --- could you recommend some good spots in Europe or Asia to hunt for authentic vintage prints \ldots?}{a general suggestion, as the user asked to forget that they collect antique maps}
\caseans{RAG (1rd)}{flea markets in Lisbon, paper art stalls in Kyoto}{\caseok}
\benchunk{[user]: \key{Please forget} that I collect antique maps from Europe and Asia. [assistant]: Got it --- I'll forget that detail \ldots}
\caseans{\textsc{MAGMA}}{antique maps, ``would fit perfectly with your collection''}{\benbad}
\memunit{}{The user now wants this information to be forgotten or ignored by the assistant. \textbullet\ \key{User collects antique maps from Europe and Asia} (contextual fact being retracted) \textbullet\ collecting antique maps}
\benlost{the request is rewritten into facts that restate the retracted detail}
}
\end{dualbox}

\vspace{6pt}
\begin{dualbox}{Processing for Heterogeneous Inputs \textperiodcentered\ \textsc{MemOS} \textperiodcentered\ MINT}
\dualpair{%
\dualhead{BenefitGreen}{Memory Benefit}
\benq{At the commit immediately after the \texttt{tabulate} dependency was first added to the project's \texttt{requirements.txt}, what version of \texttt{spotipy} was pinned in that same file?}{2.12.0}
\caseans{RAG (1rd)}{information not available}{\benbad}
\benchunk{\texttt{[step 50] (2021-02-12) diff {-}{-}git a/requirements.txt \ldots\ -spotipy==2.12.0 +spotipy==2.16.1}}
\caseans{\textsc{MemOS}}{2.12.0}{\caseok}
\memunit{}{On January 31, 2021, the user \key{added the `tabulate' library} (version 0.8.7) to the project's requirements.txt file.}
\memunit{}{On February 12, 2021, the user updated the `spotipy' library version \ldots\ \key{from 2.12.0 to 2.16.1}.}
\bennote{each commit becomes a short dated sentence}
}{%
\dualhead{TaxRed}{Memory Tax}
\benq{In what year did the project first add the \texttt{compared\_methods.py} file?}{2016}
\caseans{RAG (1rd)}{2016}{\caseok}
\benchunk{\texttt{[step 3] (2016-11-22T03:39:35Z) \ldots\ diff {-}{-}git a/compared\_methods.py b/compared\_methods.py new file \ldots}}
\caseans{\textsc{MemOS}}{2024}{\benbad}
\memunit{}{\key{On January 1, 2024}, the user shared a git diff for the `trajectory2vec' project \ldots\ The commit involved deleting files like `Kmeans\_test.py' \ldots}
\benlost{the summary carries a placeholder date instead of the commit time}
}
\end{dualbox}
\caption{Direct examples of the dual nature of memory-2: the same mechanism of a memory system brings memory benefit on one question (left) and memory tax on another (right).}
\label{fig:case_dual_2}
\end{figure*}

\onecolumn
\section{Full Results}
\label{apdx:more_results}

\subsection{Full Results of LongMemEval-s}
\label{apdx:full_lme}

\begin{table}[H]
\centering
\caption{Full results on LongMemEval-s with \texttt{GPT-5.6 Luna (low)} including judge accuracy (\%) followed by cost metrics. Token cost is split into memory construction (Build), maintenance (Mant) and inference (Infer), and end-to-end time into offline building and online QA. LongMemEval-s consists of 500 corpora (for \texttt{GPT-5.6 Luna}, we use a subset of 100 as mentioned in Appendix~\ref{apdx:benchmark_details}), and we report the aggregate result over all corpora, i.e., the overall benchmark-level score.}
\label{tab:full_lme}
\vspace{-0.6em}
\small
\setlength{\tabcolsep}{6pt}
\resizebox{\textwidth}{!}{%
\begin{tabular}{l l| c c | c c c | c c | c c}
\toprule
\multirow{2}{*}{\textbf{Type}} & \multicolumn{1}{l}{\multirow{2}{*}{\textbf{Method}}} & \multirow{2}{*}{Acc} & \multirow{2}{*}{\shortstack{Retrieval\\Scope}} & \multicolumn{3}{c|}{Token Cost} & \multicolumn{2}{c|}{E2E Time} & \multirow{2}{*}{\shortstack{LLM\\Calls (K)}} & \multirow{2}{*}{\shortstack{Memory\\Storage (K)}} \\
\cmidrule(lr){5-7} \cmidrule(lr){8-9}
 & \multicolumn{1}{l}{} & & & Build (M) & Mant (M) & Infer (M) & \shortstack{Offline\\Building} & \shortstack{Online\\QA} & & \\
\midrule
\multicolumn{11}{c}{\cellcolor{black!6}\textbf{GPT-5.6 Luna (low)}} \\
\midrule
\multirow{2}{*}{Retrieval-only} & RAG (1rd) & 82.0 & 1.91\% & -- & -- & 0.27 & 0.11h & 3.4min & 0.10 & 122.5 \\
 & RAG (2rd) & 90.0 & 2.75\% & -- & -- & 0.65 & 0.11h & 7.7min & 0.20 & 122.5 \\
\midrule
\multirow{3}{*}{\clsFlat{}Flat store} & A-Mem & 84.0 & 2.28\% & 29.3 & 145.0 & 0.44 & 117.0h & 3.2min & 98.4 & 188.5 \\
 & MemOS & 78.0 & 2.20\% & 58.8 & -- & 0.12 & 63.8h & 5.8min & 25.2 & 44.7 \\
 & SimpleMem & 84.0 & 3.23\% & 38.5 & -- & 1.42 & 13.9h & 28.5min & 6.98 & 90.9 \\
\midrule
\multirow{3}{*}{\clsHier{}Hier/Lifecycle} & MemoryOS & 80.0 & 3.18\% & 57.6 & 13.4 & 0.86 & 126.5h & 135.6min & 104.1 & 153.4 \\
 & LightMem & 77.0 & 3.48\% & 5.22 & -- & 0.09 & 15.2h & 3.0min & 2.67 & 9.67 \\
 & OpenViking & 63.0 & 6.70\% & 71.7 & -- & 0.18 & 24.9h & 5.0min & 10.9 & 27.2 \\
\midrule
\multirow{3}{*}{\clsGraph{}Graph-based} & HippoRAG2 & 81.0 & 1.28\% & 79.5 & -- & 0.46 & 11.6h & 4.4min & 118.5 & 119.4 \\
 & MAGMA & 76.0 & 4.38\% & 79.9 & -- & 0.68 & 103.2h & 40.1min & 98.5 & 144.5 \\
 & PlugMem & 70.0 & 3.77\% & 39.3 & -- & 0.63 & 39.2h & 30.3min & 49.6 & 115.6 \\
\midrule
\multirow{2}{*}{\clsTrain{}Trained Policy} & Mem-T & 78.0 & 2.69\% & 51.7 & 180.4 & 4.90 & 76.0h & 22.7min & 114.2 & 489.1 \\
 & Memory-R2 & 79.0 & 2.51\% & 51.9 & 115.6 & 0.23 & 79.4h & 2.7min & 35.3 & 89.5 \\
\bottomrule
\end{tabular}%
}
\vspace{-1em}
\end{table}

\begin{figure}[H]
\centering
\begin{subfigure}[b]{0.49\textwidth}
    \centering
    \includegraphics[width=\linewidth]{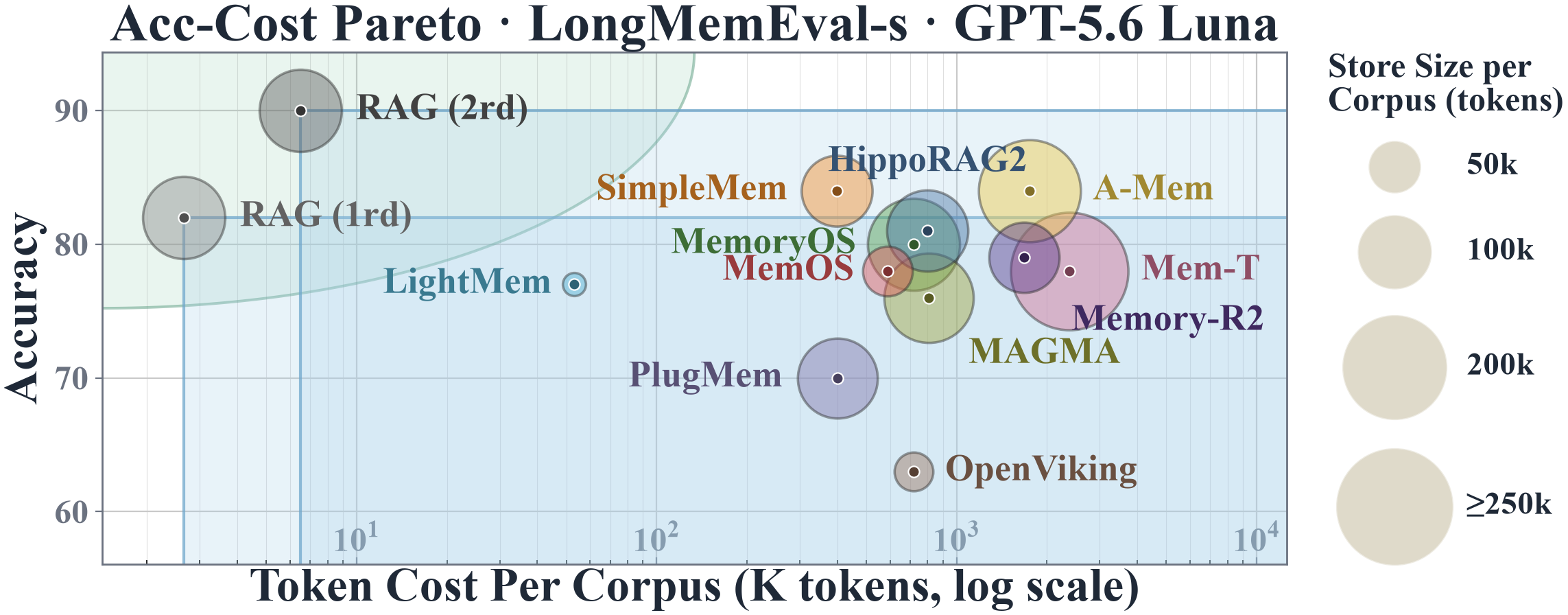}
\end{subfigure}\hfill
\begin{subfigure}[b]{0.49\textwidth}
    \centering
    \includegraphics[width=\linewidth]{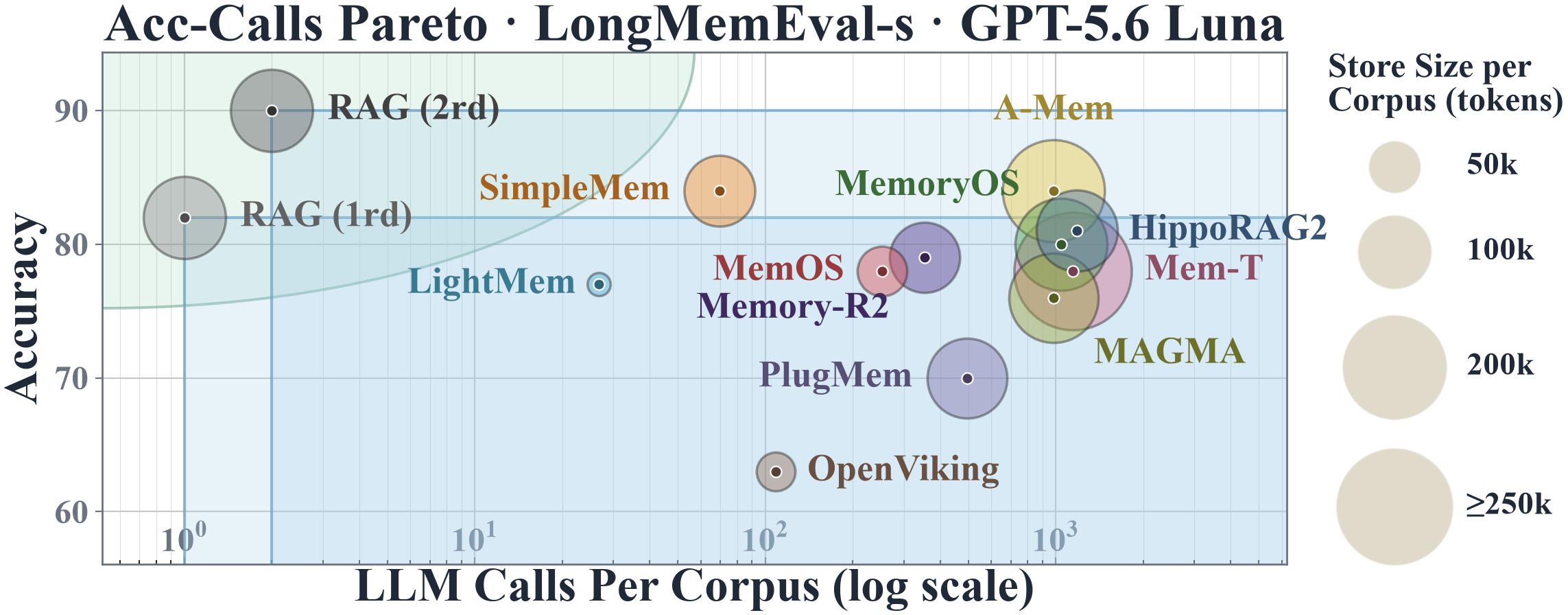}
\end{subfigure}
\par\vspace{-0.6em}
\caption{Pareto plots on LongMemEval-s with \texttt{GPT-5.6 Luna (low)}, accuracy vs.\ token cost (left) and accuracy vs.\ the number of LLM calls (right). Bubble size is the memory store size in tokens. LongMemEval-s consists of 500 corpora (for \texttt{GPT-5.6 Luna}, we use a subset of 100 as in Appendix~\ref{apdx:benchmark_details}), and report per-corpus averages.}
\label{fig:full_pareto_lme}
\end{figure}

\subsection{Full Results of PersonaMem-v2}
\label{apdx:full_persona}

\begin{table}[H]
\centering
\caption{Full results on PersonaMem-v2 with \texttt{Qwen3.6-27B} including judge accuracy (\%) followed by cost metrics. Token cost is split into memory construction (Build), maintenance (Mant) and inference (Infer), and end-to-end time into offline building and online QA. PersonaMem-v2 consists of 200 corpora (we use a subset of 40 as mentioned in Appendix~\ref{apdx:benchmark_details}), and we report the aggregate result over all corpora, i.e., the overall benchmark-level score.}
\label{tab:full_persona}
\vspace{-0.6em}
\small
\setlength{\tabcolsep}{6pt}
\resizebox{\textwidth}{!}{%
\begin{tabular}{l l| c c | c c c | c c | c c}
\toprule
\multirow{2}{*}{\textbf{Type}} & \multicolumn{1}{l}{\multirow{2}{*}{\textbf{Method}}} & \multirow{2}{*}{Acc} & \multirow{2}{*}{\shortstack{Retrieval\\Scope}} & \multicolumn{3}{c|}{Token Cost} & \multicolumn{2}{c|}{E2E Time} & \multirow{2}{*}{\shortstack{LLM\\Calls (K)}} & \multirow{2}{*}{\shortstack{Memory\\Storage (K)}} \\
\cmidrule(lr){5-7} \cmidrule(lr){8-9}
 & \multicolumn{1}{l}{} & & & Build (M) & Mant (M) & Infer (M) & \shortstack{Offline\\Building} & \shortstack{Online\\QA} & & \\
\midrule
\multicolumn{11}{c}{\cellcolor{black!6}\textbf{Qwen3.6-27B}} \\
\midrule
\multirow{2}{*}{Retrieval-only} & RAG (1rd) & 46.8 & 1.48\% & -- & -- & 2.77 & 0.07h & 58.3min & 0.93 & 166.2 \\
 & RAG (2rd) & 48.7 & 2.78\% & -- & -- & 7.68 & 0.07h & 103.7min & 1.87 & 166.2 \\
\midrule
\multirow{3}{*}{\clsFlat{}Flat store} & A-Mem & 48.4 & 1.68\% & 19.6 & 94.4 & 4.18 & 227.4h & 40.8min & 79.0 & 251.7 \\
 & MemOS & 50.2 & 2.29\% & 44.8 & -- & 1.31 & 35.3h & 69.9min & 20.7 & 43.9 \\
 & SimpleMem & 46.7 & 2.24\% & 19.8 & -- & 10.6 & 94.1h & 615.7min & 12.5 & 77.8 \\
\midrule
\multirow{3}{*}{\clsHier{}Hier/Lifecycle} & MemoryOS & 48.2 & 1.36\% & 32.3 & 13.1 & 4.48 & 145.7h & 36.2min & 84.6 & 169.7 \\
 & LightMem & 47.0 & 1.57\% & 29.8 & -- & 3.03 & 35.2h & 20.2min & 17.3 & 85.7 \\
 & OpenViking & 38.7 & 19.04\% & 40.2 & -- & 2.29 & 14.2h & 14.2min & 7.11 & 15.7 \\
\midrule
\multirow{3}{*}{\clsGraph{}Graph-based} & HippoRAG2 & 46.5 & 1.68\% & 24.7 & -- & 5.12 & 15.3h & 410.1min & 38.1 & 125.6 \\
 & MAGMA & 40.5 & 3.50\% & 49.2 & -- & 6.72 & 62.6h & 182.8min & 79.1 & 182.6 \\
 & PlugMem & 45.2 & 2.12\% & 27.9 & -- & 6.67 & 29.5h & 295.8min & 42.8 & 131.5 \\
\midrule
\multirow{2}{*}{\clsTrain{}Trained Policy} & Mem-T & 42.1 & 1.46\% & 35.4 & 161.9 & 61.0 & 59.0h & 6807.9min & 106.1 & 707.3 \\
 & Memory-R2 & 48.4 & 1.98\% & 29.0 & 64.4 & 2.61 & 85.6h & 21.3min & 20.4 & 114.5 \\
\bottomrule
\end{tabular}%
}
\vspace{-1em}
\end{table}

\begin{figure}[H]
\centering
\begin{subfigure}[b]{0.49\textwidth}
    \centering
    \includegraphics[width=\linewidth]{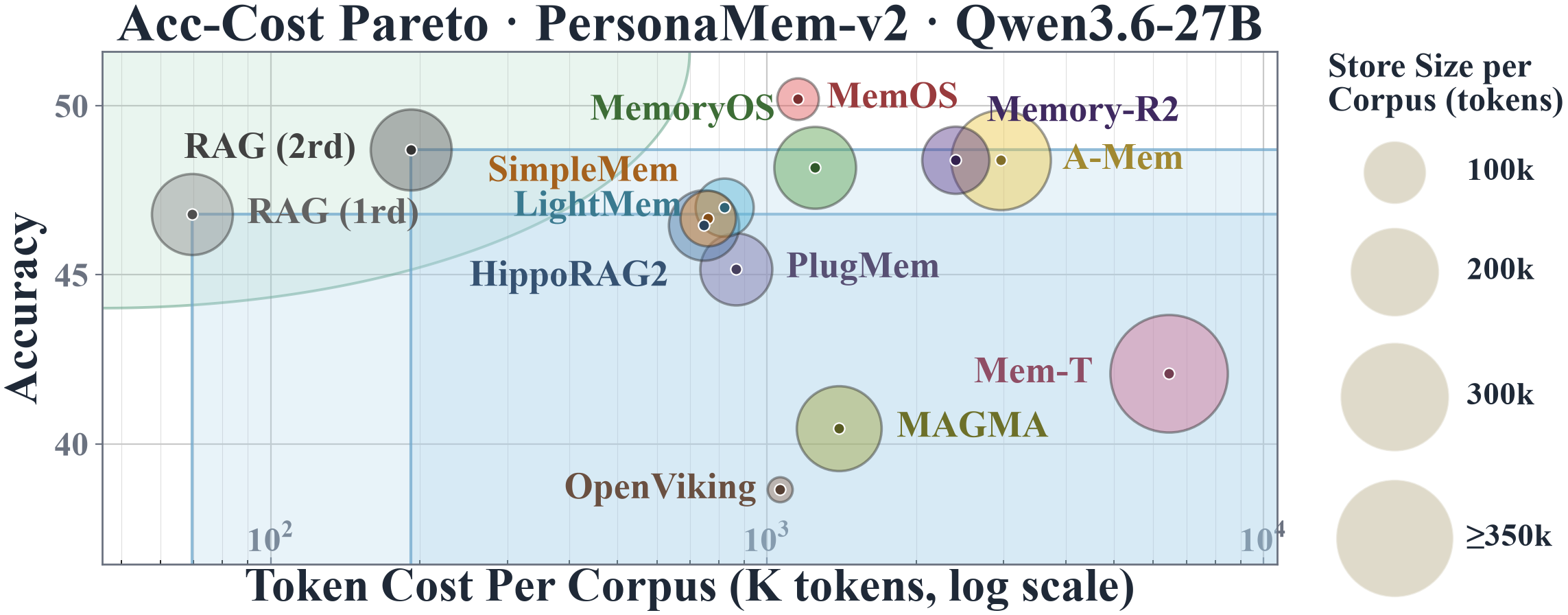}
\end{subfigure}\hfill
\begin{subfigure}[b]{0.49\textwidth}
    \centering
    \includegraphics[width=\linewidth]{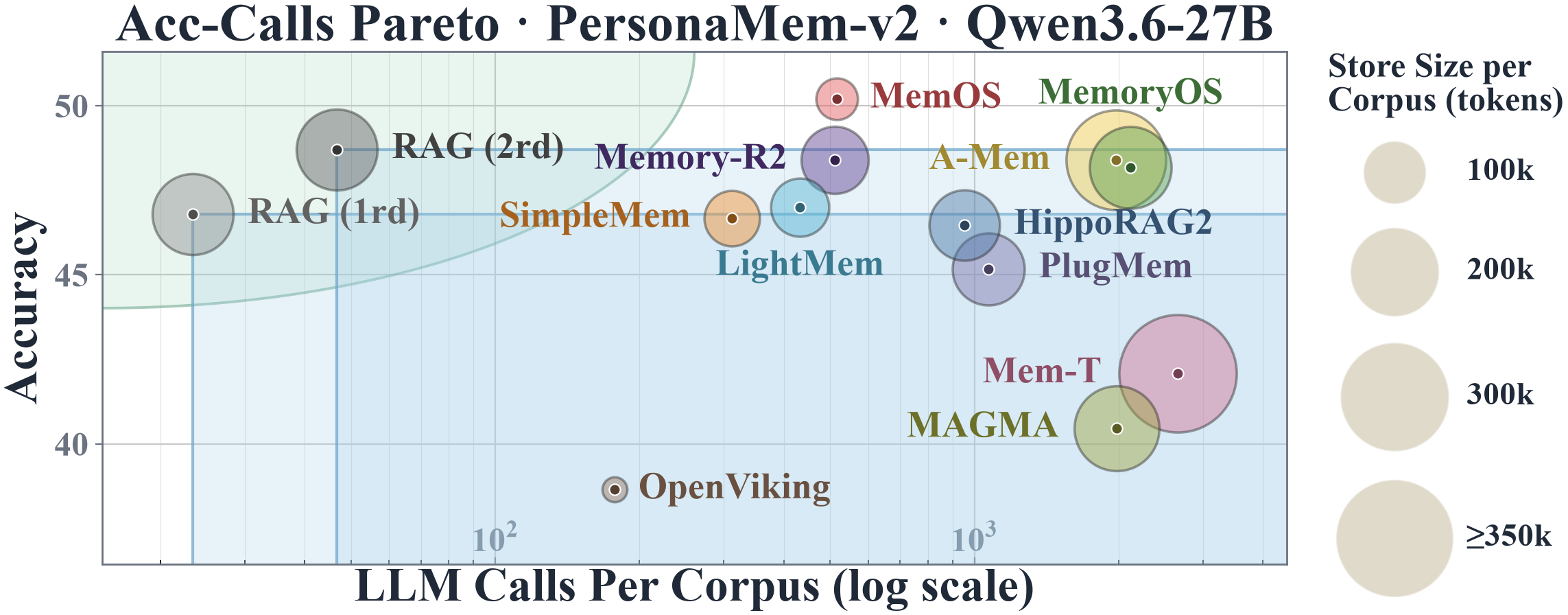}
\end{subfigure}
\par\vspace{-0.6em}
\caption{Pareto plots on PersonaMem-v2 with \texttt{Qwen3.6-27B}, accuracy vs.\ token cost (left) and accuracy vs.\ the number of LLM calls (right). Bubble size is the memory store size in tokens. PersonaMem-v2 consists of 200 corpora (we use a subset of 40 as mentioned in Appendix~\ref{apdx:benchmark_details}) and report per-corpus averages..}
\label{fig:full_pareto_persona}
\end{figure}

\subsection{Full Results of MINT-Eval}
\label{apdx:full_mint}

\begin{table}[H]
\centering
\caption{Full results on MINT-Eval with \texttt{Qwen3.5-4B} including judge accuracy (\%) followed by cost metrics. Token cost is split into memory construction (Build), maintenance (Mant) and inference (Infer), and end-to-end time into offline building and online QA. MINT-Eval consists of 595 corpora (we use a subset of 60 as mentioned in Appendix~\ref{apdx:benchmark_details}), and we report the aggregate result over all corpora, i.e., the overall benchmark-level score.}
\label{tab:full_mint}
\vspace{-0.6em}
\small
\setlength{\tabcolsep}{6pt}
\resizebox{\textwidth}{!}{%
\begin{tabular}{l l| c c | c c c | c c | c c}
\toprule
\multirow{2}{*}{\textbf{Type}} & \multicolumn{1}{l}{\multirow{2}{*}{\textbf{Method}}} & \multirow{2}{*}{Acc} & \multirow{2}{*}{\shortstack{Retrieval\\Scope}} & \multicolumn{3}{c|}{Token Cost} & \multicolumn{2}{c|}{E2E Time} & \multirow{2}{*}{\shortstack{LLM\\Calls (K)}} & \multirow{2}{*}{\shortstack{Memory\\Storage (K)}} \\
\cmidrule(lr){5-7} \cmidrule(lr){8-9}
 & \multicolumn{1}{l}{} & & & Build (M) & Mant (M) & Infer (M) & \shortstack{Offline\\Building} & \shortstack{Online\\QA} & & \\
\midrule
\multicolumn{11}{c}{\cellcolor{black!6}\textbf{Qwen3.5-4B}} \\
\midrule
\multirow{2}{*}{Retrieval-only} & RAG (1rd) & 25.8 & 6.73\% & -- & -- & 7.65 & 0.10h & 68.9min & 1.56 & 186.5 \\
 & RAG (2rd) & 26.1 & 11.93\% & -- & -- & 19.0 & 0.10h & 158.2min & 3.11 & 186.5 \\
\midrule
\multirow{3}{*}{\clsFlat{}Flat store} & A-Mem & 24.4 & 6.68\% & 25.0 & 137.4 & 7.42 & 44.4h & 50.9min & 85.9 & 244.1 \\
 & MemOS & 18.0 & 13.93\% & 54.4 & -- & 1.27 & 11.2h & 42.3min & 23.0 & 41.3 \\
 & SimpleMem & 18.6 & 4.96\% & 34.0 & -- & 27.7 & 23.8h & 191.6min & 18.0 & 68.9 \\
\midrule
\multirow{3}{*}{\clsHier{}Hier/Lifecycle} & MemoryOS & 19.0 & 31.87\% & 45.7 & 27.2 & 7.65 & 18.3h & 139.3min & 82.3 & 296.4 \\
 & LightMem & 20.8 & 25.50\% & 93.0 & -- & 2.03 & 53.5h & 18.7min & 67.8 & 82.8 \\
 & OpenViking & 10.2 & 24.95\% & 110.2 & -- & 12.3 & 17.5h & 23.6min & 14.5 & 39.6 \\
\midrule
\multirow{3}{*}{\clsGraph{}Graph-based} & HippoRAG2 & 24.6 & 6.43\% & 52.0 & -- & 7.62 & 11.8h & 385.2min & 62.5 & 150.6 \\
 & MAGMA & 23.7 & 23.31\% & 68.3 & -- & 12.7 & 33.1h & 206.1min & 85.6 & 200.4 \\
 & PlugMem & 21.5 & 5.93\% & 38.5 & -- & 9.17 & 11.7h & 217.6min & 46.6 & 204.8 \\
\midrule
\multirow{2}{*}{\clsTrain{}Trained Policy} & Mem-T & 27.0 & 9.51\% & 54.0 & 199.2 & 86.6 & 87.0h & 717.2min & 118.2 & 1202.9 \\
 & Memory-R2 & 21.3 & 30.01\% & 34.0 & 65.1 & 2.34 & 68.4h & 17.8min & 20.9 & 66.9 \\
\bottomrule
\end{tabular}%
}
\vspace{-1em}
\end{table}

\begin{figure}[H]
\centering
\begin{subfigure}[b]{0.49\textwidth}
    \centering
    \includegraphics[width=\linewidth]{Figures/4_mint_4b_score-token.png}
\end{subfigure}\hfill
\begin{subfigure}[b]{0.49\textwidth}
    \centering
    \includegraphics[width=\linewidth]{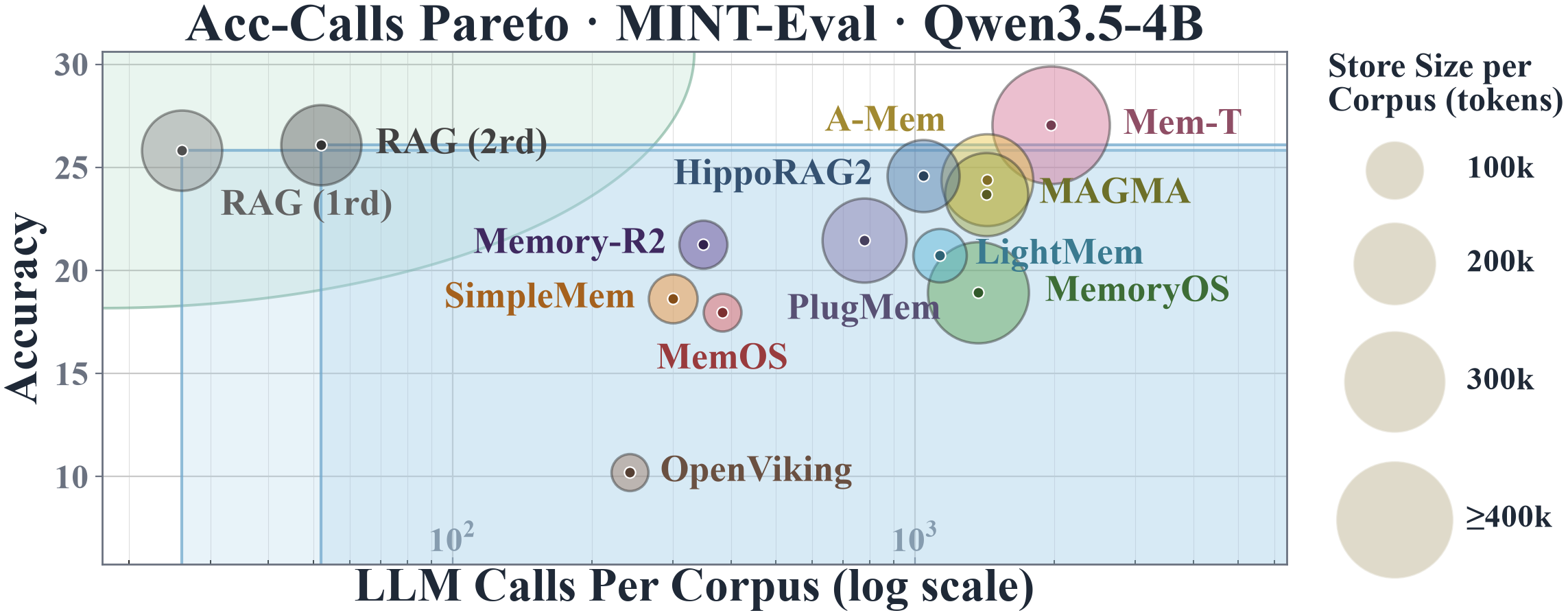}
\end{subfigure}
\par\vspace{-0.6em}
\caption{Pareto plots on MINT-Eval with \texttt{Qwen3.5-4B}, accuracy vs.\ token cost (left) and accuracy vs.\ the number of LLM calls (right). Bubble size is the memory store size in tokens. MINT-Eval consists of 595 corpora (we use a subset of 60 as mentioned in Appendix~\ref{apdx:benchmark_details}), and report per-corpus averages.}
\label{fig:full_pareto_mint}
\end{figure}

\twocolumn

\onecolumn
\section{Prompt Templates}
\label{apdx:prompt_templates}

\definecolor{QABoxFrame}{HTML}{6B7280}
\definecolor{QABoxBg}{HTML}{F4F5F6}
\definecolor{JudgeBoxFrame}{HTML}{6B7280}
\definecolor{JudgeBoxBg}{HTML}{F4F5F6}

\subsection{QA Prompts}
\label{apdx:qa_prompts}

\begin{tcolorbox}[
  enhanced, breakable, contbreak, colback=QABoxBg, colframe=QABoxFrame,
  boxrule=0.6pt, arc=4pt,
  left=10pt, right=10pt, top=8pt, bottom=8pt,
  width=\linewidth,
  title=\textbf{QA Prompt: General QA},
  fonttitle=\bfseries, coltitle=white, colbacktitle=QABoxFrame, halign title=center
]
\scriptsize\ttfamily
\textbf{[SYSTEM]}\\[2pt]
You are a helpful assistant with a long-term memory of your past
conversations with the user. Answer the user's question based on your
memories.\\[4pt]
Follow these rules:\\[2pt]
1. Be concise and specific. For dates use the format "15 July 2023"; for
durations answer in years/months/days.\\[2pt]
2. Time: reason from the timestamps attached to your memories and from the
current date given with the question. Resolve relative expressions ("last
year", "how long ago") against the current date, not against when the
conversation happened.\\[2pt]
3. Conflicts: if memories disagree because something changed over time, trust
the most recent memory, unless the question asks about an earlier point in
time.\\[2pt]
4. The needed evidence may be scattered across several conversations; combine
related memories before answering.\\[2pt]
5. Do not make up information. If the needed information does not appear in
your memories, say that the information is not available.\\[6pt]
\textbf{[USER]}\\[2pt]
\textless{}MEMORY\textgreater{} 
\textcolor{black!70}{[[MEMORY BLOCK]]}\\[4pt]
\textcolor{black!70}{(OPTIONAL USER PROFILE SECTION:)}\\
\textless{}USER PROFILE\textgreater{}\\
\textcolor{black!70}{[[USER PROFILE]]}\\[4pt]
Current date: \textcolor{black!70}{[[QUESTION DATE]]}\\
Question: \textcolor{black!70}{[[QUESTION]]}\\[4pt]
Answer concisely, providing only the answer content.
\end{tcolorbox}
\nopagebreak[4]%
\refstepcounter{table}\label{tab:longmemeval_qa_prompt}%
{\small Table~\thetable: QA prompt for general (open-ended) questions. The user-profile section
is omitted when no profile is available for this memory system.\par}
\bigskip

\begin{tcolorbox}[
  enhanced, breakable, contbreak, colback=QABoxBg, colframe=QABoxFrame,
  boxrule=0.6pt, arc=4pt,
  left=10pt, right=10pt, top=8pt, bottom=8pt,
  width=\linewidth,
  title=\textbf{QA Prompt: Open-ended Preference QA},
  fonttitle=\bfseries, coltitle=white, colbacktitle=QABoxFrame, halign title=center
]
\scriptsize\ttfamily
\textbf{[SYSTEM]}\\[2pt]
You are a helpful assistant with a long-term memory of your past
conversations with the user. Answer the user's question based on your
memories.\\[4pt]
Follow these rules:\\[2pt]
1. Be concise and specific. For dates use the format "15 July 2023"; for
durations answer in years/months/days.\\[2pt]
2. Time: reason from the timestamps attached to your memories and from the
current date given with the question. Resolve relative expressions ("last
year", "how long ago") against the current date, not against when the
conversation happened.\\[2pt]
3. Conflicts: if memories disagree because something changed over time, trust
the most recent memory, unless the question asks about an earlier point in
time.\\[2pt]
4. The needed evidence may be scattered across several conversations; combine
related memories before answering.\\[6pt]
\textbf{[USER]}\\[2pt]
\textless{}MEMORY\textgreater{} 
\textcolor{black!70}{[[MEMORY BLOCK]]}\\[4pt]
\textcolor{black!70}{(OPTIONAL USER PROFILE SECTION:)}\\
\textless{}USER PROFILE\textgreater{}\\
\textcolor{black!70}{[[USER PROFILE]]}\\[4pt]
Current date: \textcolor{black!70}{[[QUESTION DATE]]}\\
Question: \textcolor{black!70}{[[QUESTION]]}\\[4pt]
Read all sentences carefully and extract the information that may be relevant
to the question step by step.\\[2pt]
Your answer should fully consider the user's preferences.
\end{tcolorbox}
\nopagebreak[4]%
\refstepcounter{table}\label{tab:pref_qa_prompt}%
{\small Table~\thetable: QA prompt for fully open-ended preference questions (e.g., the single-session-preference
questions of LongMemEval), which have no single reference answer. Compared with Table~\ref{tab:longmemeval_qa_prompt},
the rule on unavailable information is dropped and the answer is asked to take the user's preferences into account.\par}
\bigskip

\begin{tcolorbox}[
  enhanced, breakable, contbreak, colback=QABoxBg, colframe=QABoxFrame,
  boxrule=0.6pt, arc=4pt,
  left=10pt, right=10pt, top=8pt, bottom=8pt,
  width=\linewidth,
  title=\textbf{QA Prompt: MCQ},
  fonttitle=\bfseries, coltitle=white, colbacktitle=QABoxFrame, halign title=center
]
\scriptsize\ttfamily
\textbf{[SYSTEM]}\\[2pt]
You are a helpful assistant that provides personalized responses based on the
user's preferences in conversation history.\\[2pt]
The following are relevant excerpts from your conversation history with this
user. Use them to personalize your response.\\[6pt]
\textbf{[USER]}\\[2pt]
\textless{}MEMORY\textgreater{} 
\textcolor{black!70}{[[MEMORY BLOCK]]}\\[4pt]
\textcolor{black!70}{(OPTIONAL USER PROFILE SECTION:)}\\
\textless{}USER PROFILE\textgreater{}\\
\textcolor{black!70}{[[USER PROFILE]]}\\[4pt]
\textless{}QUESTION\textgreater{}
\textcolor{black!70}{[[QUESTION]]}\\[4pt]
\textless{}CANDIDATE RESPONSES\textgreater{}\\
A. \textcolor{black!70}{[[CANDIDATE A]]}\\
B. \textcolor{black!70}{[[CANDIDATE B]]}\\
C. \textcolor{black!70}{[[CANDIDATE C]]}\\
D. \textcolor{black!70}{[[CANDIDATE D]]}\\[4pt]
Which candidate response best fits this user? Answer with one letter (A, B, C
or D) and nothing else. No explanation.
\end{tcolorbox}
\nopagebreak[4]%
\refstepcounter{table}\label{tab:personamem_qa_prompt}%
{\small Table~\thetable: QA prompt for multiple-choice questions. The user-profile section
is omitted when no profile is available.\par}

\subsection{LLM-as-Judge Prompts}
\label{apdx:judge_prompts}

\begin{tcolorbox}[
  enhanced, breakable, contbreak, colback=JudgeBoxBg, colframe=JudgeBoxFrame,
  boxrule=0.6pt, arc=4pt,
  left=10pt, right=10pt, top=8pt, bottom=8pt,
  width=\linewidth,
  title=\textbf{LLM-as-Judge Prompt: Short QA (General)},
  fonttitle=\bfseries, coltitle=white, colbacktitle=JudgeBoxFrame, halign title=center
]
\scriptsize\ttfamily
I will give you a question, a correct answer, and a response from a model.
Please check whether the response is correct or not.\\
If the meaning of response is equivalent to the correct answer or the response
contains all the intermediate steps to get the correct answer, you should also
answer CORRECT.\\[4pt]
Notes:\\[2pt]
- For time related questions, both the gold answer and the generated answer can
be a specific date, month, year, or a relative time reference (like "last
Tuesday" or "next month"). Try your best to convert relative time reference
to specific date/time and compare them in the same format - as long as the
generated answer refers to the same date or time period as the gold answer, it
should be counted as CORRECT.\\[2pt]
- For rounding issues, such as the reference answer being 2 months and the
generated answer being 2 months and 3 days; These cases should be counted as
CORRECT.\\[4pt]
Question: \textcolor{black!70}{[[QUESTION]]}\\
Gold answer: \textcolor{black!70}{[[GOLD ANSWER]]}\\
Model Response: \textcolor{black!70}{[[GENERATED ANSWER]]}\\[4pt]\\
Just return the label CORRECT or WRONG in a json format with the key as
"label".
\end{tcolorbox}
\nopagebreak[4]%
\refstepcounter{table}\label{tab:judge_short_qa_prompt}%
{\small Table~\thetable: General LLM-as-Judge prompt for short-answer
questions.\par}
\bigskip

\begin{tcolorbox}[
  enhanced, breakable, contbreak, colback=JudgeBoxBg, colframe=JudgeBoxFrame,
  boxrule=0.6pt, arc=4pt,
  left=10pt, right=10pt, top=8pt, bottom=8pt,
  width=\linewidth,
  title=\textbf{LLM-as-Judge Prompt: Open-ended Question},
  fonttitle=\bfseries, coltitle=white, colbacktitle=JudgeBoxFrame, halign title=center
]
\scriptsize\ttfamily
I will give you a question, a rubric for desired personalized response, and a
response from a model. Please output CORRECT if the response satisfies the
desired response. Otherwise, output WRONG.\\
The model does not need to reflect all the points in the rubric. The response
is correct as long as it recalls and utilizes user's personal information
correctly.\\[4pt]
Question: \textcolor{black!70}{[[QUESTION]]}\\
Rubric: \textcolor{black!70}{[[GOLD ANSWER]]}\\
Model Response: \textcolor{black!70}{[[GENERATED ANSWER]]}\\[4pt]
You may provide reasoning, but you MUST end your response with your final
answer following the format:\\
Just return the label CORRECT or WRONG in a json format with the key as
"label".
\end{tcolorbox}
\nopagebreak[4]%
\refstepcounter{table}\label{tab:judge_open_ended_prompt}%
{\small Table~\thetable: LLM-as-Judge prompt for LongMemEval
single-session-preference questions.\par}
\bigskip

\begin{tcolorbox}[
  enhanced, breakable, contbreak, colback=JudgeBoxBg, colframe=JudgeBoxFrame,
  boxrule=0.6pt, arc=4pt,
  left=10pt, right=10pt, top=8pt, bottom=8pt,
  width=\linewidth,
  title=\textbf{LLM-as-Judge Prompt: MCQ},
  fonttitle=\bfseries, coltitle=white, colbacktitle=JudgeBoxFrame, halign title=center
]
\scriptsize\ttfamily
I will give you a 4-way multiple-choice question, its options, the correct
option, and a response from a model whose reply could not be parsed into a
clean option letter.\\[4pt]
Your job is to decide which option the model's response actually commits to,
and whether that is the correct one.\\[4pt]
Notes:\\[2pt]
- The model often quotes or paraphrases an option's text instead of giving its
letter. Match on meaning, not on exact wording.\\[2pt]
- Output CORRECT only if the response commits to exactly one option AND that
option is the correct one.\\[2pt]
- Output WRONG if the response is empty, refuses to answer, hedges between
several options, restates the question, or commits to an option that is not the
correct one.\\[2pt]
- Do not reward a response for merely sounding sensible or helpful. It must
select the correct option.\\[4pt]
Question: \textcolor{black!70}{[[QUESTION]]}\\[4pt]
Options:\\
\textcolor{black!70}{[[OPTIONS BLOCK]]}\\[4pt]
Correct option: \textcolor{black!70}{[[GOLD LETTER]]}.
\textcolor{black!70}{[[GOLD ANSWER]]}\\[4pt]
Model Response: \textcolor{black!70}{[[GENERATED ANSWER]]}\\[4pt]
Just return the label CORRECT or WRONG in a json format with the key as
"label".
\end{tcolorbox}
\nopagebreak[4]%
\refstepcounter{table}\label{tab:judge_mcq_prompt}%
{\small Table~\thetable: LLM-as-Judge fallback prompt for PersonaMem-v2
multiple-choice responses that cannot be parsed into an option letter.\par}
\bigskip

\end{document}